\documentclass[english]{article}
\usepackage{geometry}

\usepackage[T1]{fontenc}
\usepackage[latin9]{inputenc}
\usepackage{bm}
\usepackage{amsmath,mathtools}
\usepackage{amssymb}
\usepackage[unicode=true,
 bookmarks=false,
 breaklinks=false,pdfborder={0 0 1},colorlinks=false]
 {hyperref}
\hypersetup{
colorlinks,citecolor=blue,filecolor=blue,linkcolor=blue,urlcolor=blue}

\makeatletter
\usepackage{amsthm}
\usepackage{comment}
\usepackage{natbib}
\usepackage{booktabs}

\usepackage{graphicx}

\usepackage[linesnumbered,ruled,vlined]{algorithm2e}

\SetCommentSty{mycommfont}
\usepackage{algorithmic}

\usepackage{float}
\usepackage{multirow}
 
\usepackage{dsfont}
\usepackage{tcolorbox}

\usepackage{color}
\definecolor{yxc}{RGB}{255,0,0}
\definecolor{yjc}{RGB}{125,0,0}
\definecolor{ytw}{RGB}{255,69,0}
\definecolor{gen}{RGB}{0,0,200}

\allowdisplaybreaks

\DeclareMathOperator{\ind}{\mathds{1}}  

\usepackage{color}
\definecolor{yxccolor}{RGB}{255,0,0}
\definecolor{ycjcolor}{RGB}{125,0,0}
\definecolor{ytwcolor}{RGB}{255,69,0}
\definecolor{gencolor}{RGB}{0,0,200}

\newcommand{\Att}{\mathsf{Attn}}
\newcommand{\att}{\mathsf{attn}}
\newcommand{\ff}{\mathsf{ff}}
\newcommand{\MLP}{\mathsf{FF}}
\newcommand{\mlp}{\mathsf{ff}}
\newcommand{\TF}{\mathsf{TF}}

\title{Transformers Meet In-Context Learning: \\ A Universal Approximation Theory\footnotetext{The first two authors contributed equally.}}

\author{%
Gen Li\thanks{Department of Statistics and Data Science, Chinese University of Hong Kong.}
\and
Yuchen Jiao\footnotemark[1] 
\and Yu Huang\thanks{Department of Statistics and Data Science, the Wharton School, University of Pennsylvania.} \and Yuting Wei\footnotemark[2]
 \and Yuxin Chen\footnotemark[2]
}
\date{June 2025;~~ Revised: August 2025}

\makeatother

\begin{document}

\theoremstyle{plain} \newtheorem{lemma}{\textbf{Lemma}}\newtheorem{proposition}{\textbf{Proposition}}\newtheorem{theorem}{\textbf{Theorem}}

\theoremstyle{assumption}\newtheorem{assumption}{\textbf{Assumption}}
\theoremstyle{remark}\newtheorem{remark}{\textbf{Remark}}

\maketitle

\begin{abstract}

	Large language models are capable of in-context learning, the ability to perform new tasks at test time using a handful of input-output examples, without parameter updates. We develop a universal approximation theory to elucidate how transformers enable in-context learning. For a general class of functions (each representing a distinct task), we demonstrate how to construct a transformer that, without any further weight updates, can predict based on a few noisy in-context examples with vanishingly small risk. Unlike prior work that frames transformers as approximators of optimization algorithms (e.g., gradient descent) for statistical learning tasks, we integrate Barron's universal function approximation theory with the algorithm approximator viewpoint. Our approach yields approximation guarantees that are not constrained by the effectiveness of the optimization algorithms being mimicked, extending far beyond convex problems like linear regression. The key is to show that (i) any target function can be nearly linearly represented, with small $\ell_1$-norm, over a set of universal features, and (ii) a transformer can be constructed to find the linear representation --- akin to solving Lasso --- at test time.


\end{abstract}


\setcounter{tocdepth}{2}
\tableofcontents

\section{Introduction}

\subsection{In-context learning}

The transformer architecture introduced by \citet{vaswani2017attention}, which leverages a multi-head attention mechanism to capture intricate dependencies between tokens in a sequence, has catalyzed remarkable breakthroughs in large language models and reshaped algorithm designs across diverse domains \citep{khan2022transformers,shamshad2023transformers,lin2022survey,gillioz2020overview}.  Built upon and powered by the transformer structure, recent pre-trained foundation models (e.g., the Generative Pre-trained Transformer (GPT) series)  have unlocked a host of emergent capabilities that were previously unattainable  \citep{bommasani2021opportunities}.

One striking example is the emergent capability of ``in-context learning'' (ICL) --- a concept coined by \cite{brown2020language} with the release of GPT-3 --- which has since become a cornerstone of modern foundation models \citep{dong2022survey}. 
In a nutshell, in-context learning refers to the ability to make reliable predictions for new tasks at test time,  without any update of the learned model. A contemporary large language model, pretrained in a universal, task-agnostic fashion, can readily perform a new task on the fly when given just a handful of input-output demonstrations.  As a concrete example, a new task might be described using some function $f(\cdot)$ (which was unknown {\em a priori} during pretraining), and be presented in a prompt containing $N$ (possibly noisy) input-output examples:  
\[
\bm{x}_1 \rightarrow f(\bm{x}_1) + \text{noise}, ~~\bm{x}_2 \rightarrow f(\bm{x}_2) + \text{noise}, 
~~\cdots,~~
\bm{x}_{N} \rightarrow f(\bm{x}_{N}) + \text{noise}, ~~
\bm{x}_{N+1} \rightarrow ~?
\]
with the model then asked to predict $f(\bm{x}_{N+1})$ given the input instance $\bm{x}_{N+1}$. 
Remarkably, a pretrained model with ICL capabilities can often make accurate predictions in this setting --- without access to the target function $f$, and without any fine-tuning --- by implicitly performing a form of statistical learning based solely on the in-context examples.

\subsection{Approximation theory for in-context learning?}

The intriguing, phenomenal capability of in-context learning has sparked substantial interest from the theoretical community, 
motivating a flurry of recent activity to illuminate its statistical principles and uncover new insights. Such theoretical pursuits have attempted to tackle various facets of ICL, spanning representation power, training dynamics, generalization performance, to name just a few \citep{garg2022can,von2023transformers,von2023uncovering,ahn2023transformers,bai2023transformers}. 
In the current paper, we contribute to this growing body of work by investigating the effectiveness of transformers as universal approximators that support ICL in statistical learning settings.

\paragraph{Prior work: transformers as algorithm approximators (for convex problems).}
Approximation theory has emerged as a powerful lens for demystifying the representation power of transformers for ICL. 
Towards this end, a predominant approach adopted in recent work is to interpret transformers as algorithm approximators, whereby  transformers are constructed to emulate the iterative dynamics of classical optimization algorithms during training, such as gradient descent \citep{von2023transformers}, preconditioned gradient descent \citep{ahn2023transformers}, 
and Newton's method  \citep{giannou2023looped,fu2024transformers}. 
The underlying rationale is that: if each iteration of these optimization algorithms can be realized via a few attention and feed-forward layers, then a multi-layer transformer could, in principle, be constructed to emulate the full iterative procedure of an optimization algorithm,  as a means to approximate solutions of the statistical learning task. 
%
Beyond emulating fixed optimization procedures, transformers can also be constructed to support in-context algorithm selection with the aid of a few more carefully chosen layers, enabling automatic selection of appropriate algorithms based on in-context demonstrations \citep{bai2023transformers,hataya2024automatic}. 


While this algorithm approximator perspective is versatile --- owing to the broad applicability of optimization algorithms like gradient descent --- its utility is fundamentally constrained by the convergence properties of the algorithms being mimicked. Noteworthily, except for \citet{giannou2023looped,hataya2024automatic}, existing analyses from this perspective have been restricted to linear regression (or learning linear functions). Indeed, optimization algorithms such as gradient and Newton's methods enjoy global convergence guarantees primarily in the context of convex loss minimization problems like linear regression, which explains why prior work along this line focused predominantly on convex statistical learning settings. 
When this approach is extended to tackle more general problems, the resulting approximation guarantees must account for the optimization error inherent to these algorithms being approximated, thereby limiting the efficacy of this technical approach for addressing broader nonconvex statistical learning problems.

\paragraph{Transformers as universal function approximators (for general problems)?} 

In this work, we aim to extend existing approximation theory to accommodate more general, possibly nonconvex, learning problems within the framework of in-context learning. In particular, we seek to investigate transformers' capabilities as direct function approximators for general function classes. 
While universal function approximation theory has been well established for neural network models --- dating back to the seminal work of \citet{Barron1993Universal} (see also \citet{hornik1994degree,bach2017breaking,juditsky2000functional,kurkova2002bounds,lee2017ability,ji2020neural,ma2022barron}) --- 
there has been limited progress in adapting Barron's function approximation framework to illuminate the representation power of transformers for ICL. It was largely unclear how transformers can learn universal representation of a general class of functions  while being fully adaptive to in-context examples at test time.

%
%

\subsection{An overview of our main contributions} 

In this paper, we make progress towards understanding the approximation capability of transformers for statistical learning tasks in the setting of in-context learning, with the aim of accommodating general function classes. 
More concretely, consider a general class $\mathcal{F}$ 
of functions mapping from $\mathbb{R}^d$ to $\mathbb{R}$, with each function representing a distinct task. Assuming that the gradient of each function from $\mathcal{F}$ has a certain bounded Fourier magnitude norm, 
we show how to construct a universal transformer, with $L$ layers and input dimension $O(d+n)$ for some large enough parameter $n$, such that: for every function $f\in\mathcal{F}$, 
this transformer can make reliable prediction given $N$ in-context examples generated based on $f$ (in a way that works universally across all tasks in $\mathcal{F}$).   More precisely, the mean squared prediction error after observing $N$ in-context examples is bounded by (up to some logarithmic factor):   
$$
\sqrt{\frac{1}{N}} + \frac{n}{L} + \frac{\log|\mathcal{N}_\varepsilon|}{n},
$$
with $|\mathcal{N}_\varepsilon|$ denoting the $\varepsilon$-covering number of the function class and input domain of interest.  Clearly, the prediction risk can be vanishingly small with judiciously chosen parameters of the transformer, where the dimension $n$ and the depth $L$ are all design parameters that we can choose in the transformer architecture. 
Notably, the term ``universal transformer'' here refers to a model whose parameters depend only on $\mathcal{F}$, without prior knowledge about the specific in-context statistical learning tasks to be performed.

Our universality approximation theory implies the plausibility of designing a transformer that can predict in-context any target function in a prescribed, general function class $\mathcal{F}$ at test time,  without any sort of additional retraining or fine-tuning. This theory extends far beyond convex statistical learning problems like linear regression. 
From a technical standpoint, our analysis consists of (i) identifying a set of universal, task-agnostic features to linearly represent any function from the target function class $\mathcal{F}$, with corresponding linear coefficients exhibiting small $\ell_1$ norm, and (ii) constructing transformer layers  to perform in-context computation of the optimal linear coefficients for the task on the fly. Given the small $\ell_1$ norm of the target linear coefficients, the second part of our analysis can be interpreted as constructing a transformer to solve the celebrated Lasso problem at test time.  At a high level, the generality of our approximation theory is achieved by integrating Barron's universal function approximation framework with the modern perspective of transformers as algorithmic approximators.  
These design ideas shed light on how transformers can leverage general-purpose representations for learning a complicated function class (far beyond linear regression) while adapting
dynamically to in-context examples.

\subsection{Related work} 

\paragraph{In-context learning.} The remarkable  ICL capability of large language models (LLMs) has inspired an explosion of recent research towards better understanding its emergence and inner workings from multiple different perspectives. Several recent studies attempted to interpret transformer-based ICL through a Bayesian lens~\citep{xie2022explanation,ahuja2023context,zhang2023and,hahn2023theory}, while another strand of work \citep{li2023transformers,kwon2025out,cole2024provable} analyzed the generalization and stability properties of transformers in the context of ICL. Particularly relevant to the current paper is a seminal line of recent work exploring the representation power of transformers. For instance, \citet{akyurek2023learning,bai2023transformers,von2023transformers} demonstrated that transformers can implement gradient descent (GD) to perform linear regression in-context, whereas \citet{guo2024transformers} extended this capability to more complex scenarios involving linear functions built atop learned representations. Note, however, that empirical analysis conducted by \citet{shen2023pretrained} revealed significant functional differences between ICL and standard GD in practical settings. Further bridging these perspectives, \citet{vladymyrov2024linear} showed that linear transformers can implement complex variants of GD for linear regression, while \citet{von2023uncovering,dai2022can} unveiled connections between ICL and meta-gradient-based optimization. Additionally, \citet{fu2024transformers,giannou2023looped} constructed transformers capable of executing higher-order algorithms such as the Newton method. Expanding this further, \citet{giannou2024transformersemulateincontextnewtons,furuya2024transformers,wang2024incontext} showed that transformers can perform general computational operations and learn diverse function classes in-context. More recently, \citet{cole2025context} investigated the representational capabilities of multi-layer linear transformers, uncovering their potential to approximate linear dynamical systems in-context.
From a complementary perspective through the lens of loss landscapes, \citet{ahn2023transformers,mahankali2024one,cheng2024transformers} showed that transformers can implement variants of preconditioned or functional GD in-context. Another important line of research investigated the optimization dynamics underlying transformers trained to perform ICL. For instance, \citet{zhang2024trained,kim2024transformers} analyzed training dynamics for linear-attention transformers, while \citet{huang2024context,li2024training,nichani2024transformers,yang2024context} studied softmax-attention transformers across a variety of ICL tasks, including linear regression~\citep{huang2024context}, binary classification~\citep{li2024training}, causal structure learning~\citep{nichani2024transformers}, representation-based learning~\citep{yang2024context}, and chain-of-thought (CoT) reasoning~\citep{huang2025cot}. Furthermore, \citet{chen2024training} explored the optimization dynamics of multi-head attention mechanisms tailored to linear regression settings.

\paragraph{Representation theory of transformers.}
Substantial theoretical efforts have been recently devoted to characterizing the representational power and capabilities of transformers and self-attention mechanisms across a variety of  computational settings and statistical tasks~\citep{perez2019turing,elhage2021mathematical,liu2023transformerslearnshortcutsautomata,likhosherstov2021expressivepowerselfattentionmatrices,wen2023transformersuninterpretablemyopicmethods,yao2023selfattentionnetworksprocessbounded,chen2024provably}. A prominent strand of recent research~\citep{sanford2023representationalstrengthslimitationstransformers,wen2024rnnstransformersyetkey,jelassi2024repeatmetransformersbetter} revealed notable advantages of transformers over alternative architectures such as RNNs. Despite these advances, several work~\citep{Hahn2020,sanford2024transformersparallelcomputationlogarithmic,peng2024limitationstransformerarchitecture,chen2024theoreticallimitationsmultilayertransformer} also identified inherent limitations of transformers, proving that transformers might fail at certain computational tasks (e.g., parity) and establishing complexity-theoretic lower bounds concerning their representational capabilities. Recently, motivated by the widespread success of the CoT techniques --- which explicitly leverage intermediate reasoning steps --- several work~\citep{li2024chainthoughtempowerstransformers,merrillexpressive,feng2023revealingmysterychainthought} began investigating the theoretical foundations and expressive power of the CoT paradigm.

















\subsection{Notation}
\label{subsec:notations}

Throughout this paper, bold uppercase letters represent matrices, while bold lowercase letters represent column vectors. For any vector $\bm{v}$, 
we use $\|{\bm v}\|_2$ to denote its $\ell_2$ norm, and $\|{\bm v}\|_1$ its $\ell_1$ norm.
For any matrix $\bm{A}$, we denote by $[{\bm A}]_{i,j}$ its $(i,j)$-th entry, and $\|{\bm A}\|$ its spectral norm.
The indicator function $\mathds{1}(\cdot)$ takes the value $1$ when the condition in the parentheses is satisfied and zero otherwise.
The sign function $\mathsf{sign}(x)$ returns $1$ if $x>0$, $-1$ if $x<0$, and $0$ if $x=0$.
For any scalar function $\sigma:\mathbb{R}\to\mathbb{R}$, the notation $\sigma({\bm x})$ for ${\bm x}\in\mathbb{R}^d$ denotes the elementwise application of $\sigma$ to each entry of ${\bm x}$.
Let $\mathcal{X} = \{N,L,n,\log|\mathcal{N}_\varepsilon|,C_{\mathcal{F}},\sigma\}$ (and sometimes with the additional inclusion of some precision parameter $\varepsilon_{\mathsf{pred}}$).
The notation $f(\mathcal{X}) = O(g(\mathcal{X}))$ or $f(\mathcal{X}) \lesssim g(\mathcal{X})$ (resp.~$f(\mathcal{X}) \gtrsim g(\mathcal{X})$) means that there exists a universal constant $C_0> 0$ such that $f(\mathcal{X}) \le C_0 g(\mathcal{X})$ (resp.~$f(\mathcal{X}) \ge C_0 g(\mathcal{X})$) for any choice of $\mathcal{X}$. 
The notation $f(\mathcal{X})\asymp g(\mathcal{X})$ means $f(\mathcal{X}) \lesssim g(\mathcal{X})$ and $f(\mathcal{X}) \gtrsim g(\mathcal{X})$ hold simultaneously. 
We define $\widetilde{O}(\cdot)$ in the same way as $O(\cdot)$ except that it hides logarithmic factors. 
We also use $\bm{0}$  to denote the all-zero vector. For any positive integer $m$, we denote $[m]\coloneqq  \{1,\dots,m\}$.  

\section{Problem formulation}
\label{sec:formulation}

\subsection{Setting: in-context learning} 
To set the stage, we formulate an in-context learning setting for statistical learning tasks, which comprises the following components: 
\begin{itemize}
    \item {\em Function class.} We denote by $\mathcal{F}$ a class of 
    real-valued functions, mapping from $\mathbb{R}^d$ to $\mathbb{R}$, that we aim to learn. Each function $f\in \mathcal{F}$ represents a distinct prediction task (i.e., the task of predicting the output $f(\bm{x})$ given an input $\bm{x}$). 

    \item {\em Input sequence.} The input sequence, typically provided in the prompt at test time, is composed of $N$ noisy input-output pairs --- namely, $N$ in-context examples --- along with a new input vector for prediction. To be precise, the input sequence takes the form of
    \begin{subequations}
    \label{eq:prompt-defn}
    \begin{equation}
        \big( \bm{x}_1, y_1, \bm{x}_2, y_2, \dots, \bm{x}_N, y_N, \bm{x}_{N+1} \big),
        \label{eq:prompt-defn-P}
    \end{equation}
    where for every $i$, 
  \begin{align}
\bm{x}_i &\overset{\text{i.i.d.}}{\sim} \mathcal{D}_{\mathcal{X}},\qquad 
z_i \overset{\text{i.i.d.}}{\sim} \mathcal{D}_{\mathcal{Z}},\qquad 
y_i = f(\bm{x}_i) + z_i \text{ for some function }f\in \mathcal{F}. 
\end{align}  
\end{subequations}
Here, 
$\{\bm{x}_i\} \subset \mathbb{R}^d$ (resp.~$\{z_i\} \subset \mathbb{R}$) are input  vectors (resp.~noise) sampled randomly from the distribution $\mathcal{D}_{\mathcal{X}}$ (resp.~$\mathcal{D}_{\mathcal{Z}}$), 
and the corresponding output vectors are produced by some function $f\in \mathcal{F}$ not revealed to the learner.  
    Throughout this paper, the noise  $\{z_i\}$ is assumed to be independent zero-mean sub-Gaussian with the sub-Gaussian norm upper bounded by $\sigma$ \citep{vershynin2018high}, that is,  
    \begin{align}
        \mathbb{E}[z_i] = 0 \qquad \text{and} \qquad
        \mathbb{E}[{\rm e}^{t z_i}]\le \exp\Big(\frac{\sigma^2t^2}{2}\Big)
        \text{ for every }t\in \mathbb{R}.
    \end{align}
    For simplicity, we assume throughout that all input vectors lie within a unit Euclidean ball: 
    \begin{align}
    \bm{x} \in \mathcal{B} \coloneqq \{\bm{u}\mid \|\bm{u}\|_2\leq 1\}\qquad \text{for any input vector }\bm{x},
    \end{align}
    but this assumption can be easily relaxed and generalized.

		It is worth emphasizing that $\{(\bm{x}_i,y_i)\}$ should be regarded {\em not} as training examples, but rather as in-context demonstrations, as they are typically provided within the prompt at test time instead of being used during the (pre)training phase.

\end{itemize}

\paragraph{Goal.} The aim is to design a transformer that, given an input sequence as in \eqref{eq:prompt-defn} at test time, outputs a prediction $\widehat{y}_{N+1}$ with vanishingly small $\ell_2$ risk, namely,  
$$
	\mathbb{E}\big[ \big( \widehat{y}_{N+1} - f(\bm{x}_{N+1}) \big)^2\big] \approx 0
$$ 
in some average sense. Here, the input sequence \eqref{eq:prompt-defn} can be produced by any (a priori unknown) function $f$ from a prescribed, general function class $\mathcal{F}$. 
    Particular emphasis is placed on {\em universal design}, where the objective is to find a single transformer (to be described next) that makes reliable predictions simultaneously for all $f\in \mathcal{F}$, without knowing which task $f$ to tackle in advance. 
    This universal design requirement aligns closely with the concept of ICL, as the goal is for the pre-trained transformer to predict in-context, without performing any prompt-specific parameter updates of the transformer.


\subsection{Transformer architecture} 
Next, let us present a precise description of the transformer architecture to be used for in-context learning.

\paragraph{Input matrix.} 
Throughout this paper, 
we would like to use a matrix 
\begin{equation}
    \bm{H} = \big[\bm{h}_1, \cdots, \bm{h}_{N+1} \big] \in \mathbb{R}^{D\times (N+1)}
	\label{eq:input-matrix-H}
\end{equation}
to encode the input sequence \eqref{eq:prompt-defn-P} comprising $N$ input-output examples along with an additional new input vector. 
Here, the input dimension $D$ is typically chosen to be larger than $d$ to allow for incorporation of several useful auxiliary features. 
A matrix of this form will serve as the input to each layer of the transformer. 
More concretely, in our construction (to be detailed momentarily), an input matrix $\bm{H}$ for each layer takes the following form:  
\begin{align}
\bm{H}=\left[\begin{array}{cccc}
\bm{x}_{1} & \cdots & \bm{x}_{N} & \bm{x}_{N+1}\\
1 & \cdots & 1 & 1\\
y_{1} & \cdots & y_{N} & 0\\
\vdots & \mathsf{auxiliary} & \mathsf{info} & \vdots\\
	\widehat{y}_{1}^{\mathsf{prev}} & \cdots & \widehat{y}_{N}^{\mathsf{prev}} & \widehat{y}_{N+1}^{\mathsf{prev}}
\end{array}\right],
\label{eq:input-matrix-H-detailed}
\end{align}
where each column entails the original noisy input-output pair $(\bm{x}_i,y_i)$ (except the last column where $y_{N+1}$ is replaced with 0), a constant 1, a few dimension containing auxiliary information or certain intermediate updates, and some (intermediate) prediction $\widehat{y}_i^{\mathsf{prev}}$ obtained previously that we can take advantage of. 
Both the auxiliary information components and the $\widehat{y}_i^{\mathsf{prev}}$'s can vary across different layers of the transformer. 
As we shall see momentarily, in our construction,  the $\widehat{y}_i^{\mathsf{prev}}$'s are initialized to zero at the very beginning, and are updated in subsequent layers of the transformer to yield improved predictions of the target function values.

\paragraph{Basic building blocks.} 
Next, we single out two basic building blocks of transformers. 
\begin{itemize}
    \item {\em Attention layer.} For any input matrix $\bm{H}$ of the form \eqref{eq:input-matrix-H}, the (self)-attention operator is defined as
    \begin{align}
    \att(\bm{H}; \bm{Q}, \bm{K}, \bm{V}) \coloneqq \frac{1}{N}\bm{V}\bm{H}\sigma_{\att}\big((\bm{Q}\bm{H})^{\top}\bm{K}\bm{H} \big),
    \end{align}
    where $\bm{Q},\bm{K},\bm{V}\in \mathbb{R}^{D\times D}$ represent the parameter matrices, commonly referred to as the query, key, and value matrices, respectively, and the activation function  $\sigma_{\att}(\cdot)$ is applied either columnwise or entrywise to the input.  
    A multi-head (self)-attention layer, which we denote by $\Att_{\bm{\Theta}}(\cdot)$ 
    as parameterized by $\bm{\Theta}=\{\bm{Q}_m, \bm{K}_m, \bm{V}_m\}_{1\leq m\leq M} \subset \mathbb{R}^{D\times D}$,  computes a superposition of the input and the outputs from $M$ attention operators (or attention heads). Namely, given an input matrix $\bm{H}$, the output of the attention layer with $M$ attention heads is defined as
    \begin{align}
        \Att_{\bm{\Theta}}(\bm{H}) \coloneqq 
        \bm{H} + \sum_{m = 1}^M \att(\bm{H}; \bm{Q}_m, \bm{K}_m, \bm{V}_m).
    \end{align}
    This attention mechanism plays a pivotal role in the transformer architecture \citep{vaswani2017attention}, allowing one to dynamically attend to different parts of the input data. 
    
    \item {\em Feed-forward layer (or multilayer perceptron (MLP) layer).} Given an input matrix $\bm{H}$, the feed-forward layer produces an output as follows:
    \begin{align}
        \label{eq:defn-FF-MLP-layer}
        \MLP_{\bm{\Theta}}(\bm{H} ) \coloneqq \bm{H} + \bm{U}\sigma_{\ff}(\bm{W}\bm{H}),
    \end{align}
    where $\bm{\Theta}=\{\bm{U},\bm{W}\}\subset \mathbb{R}^{D\times D}$ 
    bundles the parameter matrices 
    $\bm{U}$ and $\bm{W}$ together, and the activation function $\sigma_{\mlp}(\cdot)$ is applied entrywise to the input. 
    
\end{itemize}

\noindent 
Throughout this paper, the two activation functions described above are chosen to be the logistic function and the ReLU function: 
\begin{align}
\sigma_{\att}(x) = \frac{{\rm e}^x}{{\rm e}^{x}+1},\qquad \sigma_{\ff}(x) = x\ind(x>0),
\label{eq:sigma-attn-ff-choice}
\end{align}
each of which is applied entrywise to its respective input. 
Additionally, in our construction, each attention and feed-forward layer will be carefully chosen so that their output preserve the same form as that of the input \eqref{eq:input-matrix-H-detailed}, detailed momentarily.


\paragraph{Multi-layer transformers.} 
With the aforementioned building blocks in place, we can readily introduce the multi-layer transformer architecture. 
Given an input $\bm{H}^{(0)}\in \mathbb{R}^{D\times (N+1)}$ of the form \eqref{eq:input-matrix-H}, a transformer comprising $L$ attention layers --- each coupled with a feed-forward layer --- carries out the following computation:
\begin{subequations}
\label{eq:transformer-structure-defn}
\begin{align}
\bm{H}^{(l)} & =\MLP_{\bm{\Theta}_{\mlp}^{(l)}}\Big(\Att_{\bm{\Theta}_{\att}^{(l)}}\big(\bm{H}^{(l-1)}\big)\Big), \qquad l = 1, \dots, L,
\end{align}
with the final output given by
\begin{align}
 \TF_{\bm{\Theta}}\big(\bm{H}^{(0)}\big) & \coloneqq \bm{H}^{(L)}.
 \end{align}
 \end{subequations}
Here,  $\bm{\Theta}$ encapsulates all parameter matrices:
$$
    \bm{\Theta}= \big\{ \bm{\Theta}^{(l)}_{\att}, \bm{\Theta}^{(l)}_{\mlp} \big\}_{1\leq l\leq L} 
    \quad \text{with }~\bm{\Theta}^{(l)}_{\att}= \big\{\bm{Q}_m^{(l)}, \bm{K}_m^{(l)}, \bm{V}_m^{(l)} \big\}_{1\leq m\leq M} ,\, \bm{\Theta}^{(l)}_{\mlp}= \big\{\bm{U}^{(l)}, \bm{W}^{(l)} \big\}\subset \mathbb{R}^{D\times D}.
$$
In particular, the transformer's prediction for the $(N+1)$-th input can be read out from  the very last entry of $\TF_{\bm{\Theta}}(\bm{H}^{(0)})$, i.e.,
\begin{equation}
    \widehat{y}_{N+1} = \mathsf{ReadOut}\big(\TF_{\bm{\Theta}}\big(\bm{H}^{(0)}\big)\big) \coloneqq 
    \big[ \bm{H}^{(L)}\big]_{D,N+1}. 
    \label{eq:y-hat-N-plus-1-readout}
\end{equation}

\subsection{Key quantities}
\label{sec:Barron-pars}

Before embarking on our main theory,  let us take a moment to isolate a couple of key quantities that play a crucial role in our theoretical development.


\paragraph{Barron-style parameter.} 
For any absolutely integrable function\footnote{In fact, we only need to ensure that $f$ is absolutely integrable within the unit ball $\mathcal{B}$, given our assumption that $\bm{x}\in \mathcal{B}$. } $f: \mathbb{R}^{d} \rightarrow \mathbb{R}$, 
we denote by $F_f$ its Fourier transform, 
which allows one to express
\begin{subequations}
\begin{align}
F_f(\bm{\omega} )  &= \frac{1}{(2\pi)^d}\int_{\bm{x}} e^{-j \bm{\omega}^{\top}{\bm x}}f({\bm x}) \mathrm{d} \bm{x} \\
f({\bm x}) &= \int_{\bm{\omega}} e^{j \bm{\omega}^{\top}{\bm x}}F_f(\bm{\omega} ) \mathrm{d} \bm{\omega}
\end{align}
\end{subequations}
with $j=\sqrt{-1}$ the imaginary unit. 
It is also helpful to define, for each $\bm{\omega}$, the maximum magnitude of the Fourier transform over the function class $\mathcal{F}$ as:
\begin{align}
F^{\mathsf{sup}}(\bm{\omega}) \coloneqq \sup_{f \in \mathcal{F}} \big| F_f(\bm{\omega}) \big|. 
\end{align}
Inspired by the seminal work \citet{Barron1993Universal}, we introduce the key quantity:
\begin{align}
\label{eq:condition-Fourier}
C_{\mathcal{F}} \coloneqq \sup_{f\in \mathcal{F}}|f(\bm{0})| + \int_{\bm{\omega}} \|\bm{\omega}\|_2F^{\mathsf{sup}}(\bm{\omega})\mathrm{d}\bm{\omega} < \infty,
\end{align}
which we shall refer to as the Barron parameter for $\mathcal{F}$. 
Recognizing that $j\bm{\omega} F_f(\bm{\omega})$ is precisely the Fourier transform of $\nabla f(\bm{x})$, one can alternatively express $C_{\mathcal{F}}$ as
\begin{align}
\label{eq:condition-Fourier-gradient}
C_{\mathcal{F}} = \sup_{f\in \mathcal{F}}|f(\bm{0})| + \int_{\bm{\omega}} \sup_{f\in \mathcal{F}} \big\| F_{\nabla f}(\bm{\omega}) \big\|_2 
\mathrm{d}\bm{\omega} , 
\end{align}
which tracks the $\ell_1$ norm of the maximum Fourier magnitude of the function gradient $\nabla f$. 
Informally, this quantity bounds the first moment of the Fourier magnitude distribution over this function class $\mathcal{F}$.  
Compared with the quantity $C_f\coloneqq \int_{\bm{\omega}} \|\bm{\omega}\|_2|F_f(\bm{\omega})|\mathrm{d}\bm{\omega}$ 
originally introduced in \cite{Barron1993Universal} for a given function $f$, the main difference lies in the fact that $C_{\mathcal{F}}$ involves taking the supremum over the entire function class $\mathcal{F}$. 

It is worth noting that \citet[Section~IX]{Barron1993Universal} has studied $C_f$ for a number of functions commonly used in statistical learning (e.g., Gaussian functions, logistic functions, ridge functions, polynomials, linear combination of Barron functions), which sheds light on the evaluation of $C_{\mathcal{F}}$ for related function classes. 
%
To help illustrate the effectiveness of our theoretical framework, we provide below a few examples of function classes and bound the corresponding Barron-style parameter $C_{\mathcal{F}}$, which are clearly far from exhaustive. 
For completeness, we provide the calculation in Appendix \ref{app:calculation-CF}. 
\begin{itemize}
	\item {\em Class of linear functions.} For a class of linear functions 
		\begin{align}
			\mathcal{F}^{\mathsf{linear}} \coloneqq
			\big\{ f_{\bm{a},b} \mid \|{\bm a}\|_2\le C_a,|b|\le C_b \big\}
			\qquad \text{with } f_{\bm{a},b}(\bm{x}) = {\bm a}^{\top}{\bm x} + b, 
		\end{align}
		we have $C_{\mathcal{F}^{\mathsf{linear}}}\le C_a + C_b$.


	\item {\em Linear combination of function classes.} For any $i=1,\dots, M$, let $\mathcal{F}_i$ be a function class with Barron-style parameter $C_{\mathcal{F}_i}<\infty$.  
		Consider the following class of  functions:  
		\begin{align}
			\mathcal{F}^{\mathsf{comp}} \coloneqq \{ g_{\bm{a},b} \mid \|{\bm a}\|_1\le C_a,|b|\le C_b, f_i\in\mathcal{F}_i, \forall i\},
			\qquad \text{with } g_{\bm{a},b}(\bm{x}) = \langle {\bm a}, {\bm f}({\bm x}) \rangle + b, 
		\end{align}
		where ${\bm f}(\bm{x}) = [f_i(\bm{x})]_{1\leq i\leq M}$ is a vector with the $i$-th component being  $f_i(\bm{x})$. Its Barron-style parameter satisfies $C_{\mathcal{F}^{\mathsf{comp}}}\le 2C_a\max_{1\leq i\leq M}C_{\mathcal{F}_i} + C_b$.
\item {\em Two-layer neural network with logistic activation.}
Consider the class of functions parameterized by a two-layer neural network, whose activation functions are the logistic (sigmoid) function; more precisely, consider
\begin{align*}
\mathcal{F}^{\mathsf{neural}}
&\coloneqq \left\{f_{\rho}({\bm x}) \mid {\rho} : \mathbb{R}^d \to \mathbb{R}\text{, with } |{\rho}({\bm a})|\le {\rho}_{\max}({\bm a})\text{ for any }\bm{a}
\right\},
\end{align*}
%
where 
\begin{align*}
f_{\rho}({\bm x}) = \int_{\bm a}{\rho}({\bm a})\sigma({\bm a}^{\top}{\bm x})\mathrm{d} {\bm a},\quad \text{with }\sigma(x) = \frac{1}{1+{\rm e}^{-x}}
.
\end{align*}
%
%
The associated Barron-style parameter obeys $C_{\mathcal{F}^{\mathsf{neural}}}\le \frac{1}{4}\int_{\bm a}{\rho}_{\max}({\bm a})
(\|{\bm a}\|_2+2)\mathrm{d} {\bm a}$.
\end{itemize}



\paragraph{Covering number.} 
Recall that we have restricted our input space to reside within the unit Euclidean ball $\mathcal{B} = \{{\bm x}\in\mathbb{R}^d:\|{\bm x}\|_2\le 1\}$. 
A set, denoted by $\mathcal{N}_{\varepsilon}$, is said to be an $\varepsilon$-cover of $\mathcal{F}\times \mathcal{B}$ if, for every $(f, {\bm x}) \in \mathcal{F}\times \mathcal{B}$, 
there exists some $(\widehat{f}, \widehat{{\bm x}}) \in \mathcal{N}_{\varepsilon}$ such that
\begin{align}
\|{\bm x} - \widehat{{\bm x}}\|_2 \le \varepsilon
\quad\text{and}\quad
\big|f({\bm x}) - f({\bm 0}) - (\widehat{f}(\widehat{{\bm x}}) - \widehat{f}({\bm 0}))\big| \le C_{\mathcal{F}}\varepsilon.
\label{eq:defn-eps-cover}
\end{align}
In this paper, we shall take $\mathcal{N}_{\varepsilon}$ to be an $\varepsilon$-cover of $\mathcal{F}$ with the smallest cardinality, and refer to $|\mathcal{N}_{\varepsilon}|$ as the covering number of $\mathcal{F}$ with precision $\varepsilon$.

\section{Main results: transformers as universal in-context learners}
\label{sec:reults}

Equipped with the preliminaries and key quantities in Section~\ref{sec:formulation}, we are now ready to present our main theoretical findings, as summarized in the theorem below.
\begin{theorem}\label{thm:main}
Assume that $\mathcal{F}$ is a subset of absolutely integrable functions $f$ mapping from the Euclidean ball $\mathcal{B}\subset\mathbb{R}^d$ to $\mathbb{R}$, and that it satisfies $C_{\mathcal{F}}<\infty$, where  $C_{\mathcal{F}}$ is defined in \eqref{eq:condition-Fourier}.
Let $\mathcal{N}_\varepsilon$ be an $\varepsilon$-cover of $\mathcal{F}\times \mathcal{B}$ (see \eqref{eq:defn-eps-cover}).  
Then one can construct a transformer 
such that: 
\begin{itemize}
    \item[i)]  it has $L$ layers, $M=O(1)$ attention heads per layer, and input dimension $D=d+2n+7$ for some $n\gtrsim \log|\mathcal{N}_\varepsilon|$; 

    \item[ii)] for every $f\in \mathcal{F}$ and any random input sequence $\{({\bm x}_i,y_i)\}_{1\le i\le N} \cup \{\bm{x}_{N+1}\}$ generated according to \eqref{eq:prompt-defn}, with probability at least $1-O(N^{-10})$, this transformer's prediction $\widehat{y}_{N+1}$ (cf.~\eqref{eq:y-hat-N-plus-1-readout})  satisfies
\begin{align}
\mathbb{E}\big[\big(\widehat{y}_{N+1} - f( \bm{x}_{N+1}) \big)^2 \big]
\lesssim 
\left(\sqrt{\frac{\log N}{N}} + \frac{n}{L}\right)C_{\mathcal{F}}(C_{\mathcal{F}} + \sigma) + \frac{C_{\mathcal{F}}^2\log|\mathcal{N}_\varepsilon|}{n},
\label{eq:prediction-error-thm1}
\end{align}

\end{itemize}
Here, the precision of the $\varepsilon$-cover is taken to satisfy  $\varepsilon\lesssim \sqrt{\frac{\log N}{N}} + \frac{n}{L}$; the expectation in \eqref{eq:prediction-error-thm1} is taken over the randomness of ${\bm x}_{N+1}$; and the probability that the event \eqref{eq:prediction-error-thm1} occurs is governed by the randomness in the input sequence $\{({\bm x}_i, y_i)\}_{1\le i\le N}$. 

\end{theorem}

It is worth taking a moment to reflect on the interpretation and implications of this theorem. 

    
\paragraph{Universal in-context prediction.} 
    Theorem~\ref{thm:main} establishes the existence of a pretrained, multi-layer multi-head transformer --- configured with judiciously chosen parameters, depth, number of heads, etc.~--- that can make reliable predictions based on in-context demonstrations for statistical learning tasks. 
		This in-context prediction capability is universal, in the sense that a single transformer can simultaneously handle all functions $f$ in a fairly general function class of interest (going far beyond linear function classes as studied in much of the prior literature), without requiring any additional training or prompt-specific parameter updates.

    \paragraph{Parameter choices.} According to Theorem \ref{thm:main}, the in-context prediction risk depends on the number of layers $L$, the number of input-output examples $N$,  the transformer's input dimension $D$, and the  intrinsic data dimension $d$. Specifically, when both the Barron-style parameter $C_{\mathcal{F}}$ and the noise level $\sigma$ are no larger than $O(1)$, the mean squared prediction error cannot exceed the order $\widetilde{O}(1/\sqrt{N} + n/L+ \log|\mathcal{N}_\varepsilon|/n)$, where $2n \approx D-d$.
    This result implies that properly increasing the model complexity --- through the use of deeper architectures (i.e., the ones with larger $L$) and  higher input dimension $D$ --- and utilizing more in-context examples (i.e., larger $N$) --- could potentially enhance the in-context prediction accuracy of the transformer. More concretely, to achieve an $\varepsilon_{\mathsf{pred}}$-accurate prediction (with $\varepsilon_{\mathsf{pred}}$ the target mean squared prediction error), it suffices to employ a transformer with parameters satisfying (up to log factors)
    \begin{subequations}
    \label{eq:param-choice-construction}
    \begin{align}
    D-d&\asymp C_{\mathcal{F}}^2\varepsilon_{\mathsf{pred}}^{-1}\log|\mathcal{N}_\varepsilon|, \\
    N&\gtrsim C_{\mathcal{F}}^2(C_{\mathcal{F}}+\sigma)^2\varepsilon^{-2}_{\mathsf{pred}}, \\
    L&\gtrsim (D-d) C_{\mathcal{F}}(C_{\mathcal{F}}+\sigma)\varepsilon^{-1}_{\mathsf{pred}}\asymp C_{\mathcal{F}}^3(C_{\mathcal{F}}+\sigma)\varepsilon_{\mathsf{pred}}^{-2}\log|\mathcal{N}_\varepsilon|.
    \end{align}
    \end{subequations}
 Note that both the dimension $D$ and the depth $L$ are parameters of the architecture that can be selected during pretraining. 


\paragraph{Model complexity vs.~covering number of the function class.} 
Our approximation theory allows the function class of interest to be fairly general. Notably, both the input dimension (including that of auxiliary features) and the depth of the transformer we construct only need to 
scale logarithmically with the covering number of the target function class (see \eqref{eq:param-choice-construction}). In other words, in order to achieve sufficient representation power for ICL, the model complexity needs to grow with the complexity of the target function class --- but a logarithmic scaling with the covering number of $\mathcal{F}$ suffices. 


\paragraph{Comparison with prior algorithm approximation viewpoint.} 
Theorem~\ref{thm:main} unveils that transformers can serve as universal function approximators in the framework of ICL for general function classes.  
  In contrast, a substantial body of recent work --- e.g., \citet{von2023transformers,von2023uncovering,bai2023transformers,ahn2023transformers,giannou2023looped,xie2022explanation,cheng2024transformers} --- has primarily focused on interpreting transformers as algorithm approximators. As alluded to previously, the approximation theory derived from the algorithm approximation perspective is often constrained  by the effectiveness of the specific algorithms being approximated. For instance, algorithms like gradient descent and Newton's method are typically not guaranteed to perform well outside the realm of convex optimization. This limitation partly explains why much of the prior literature has concentrated on relatively simple convex problems like linear regression. 
		By contrast, our universal approximation theory is not tied to the performance of such (mesa)-optimization algorithms, and as a result,  can often deliver direct approximation guarantees for much broader in-context learning problems (including nonconvex problems and learning general function classes). 

\paragraph{A glimpse of our technical innovation.} 
		To develop our universal approximation theory, we pursue a novel approach that  
		seamlessly integrates Barron's universal function approximation theory with the modern algorithm approximation perspective. Our technical approach hinges upon two key insights: 
		\begin{itemize}

			\item[(i)] Any target function $f \in \mathcal{F}$ can be approximately represented as a linear combination of a set of universal features as constructed in our analysis (which depend solely on $\mathcal{F}$ but not any specific function $f$). Moreover, the corresponding linear coefficients provably exhibit low $\ell_1$-norm, suggesting some sort of compressibility or even sparsity with respect to the set of universal features (as in the language of compressed sensing \citep{candes2006near}).

		\item[(ii)] As the classical Lasso algorithm \citep{tibshirani1996regression} is ideal to identify solutions with small $\ell_1$-norms in linear regression, we then proceed to construct transformers that can approximate Lasso solutions. 
        Given the convexity of Lasso, the modern algorithm approximator perspective comes in handy, illuminating how transformers can be designed to solve Lasso by, for example, emulating iterations of the proximal gradient method. 

		\end{itemize}
We will elaborate on the above ideas in the ensuing sections.

 %

\section{Analysis}
\label{sec:analysis}

In this section, we present the key steps for establishing Theorem~\ref{thm:main}. 
Informally, our proof comprises the following key ingredients: 
\begin{itemize}
    \item Identify a collection of universal features such that every function (or task) in $\mathcal{F}$ can be (approximately) represented as a linear combination of these features. Also, show that the corresponding linear coefficients always have small $\ell_1$ norm. 

    \item For each function $f\in \mathcal{F}$, the corresponding linear coefficients can be found by means of a Lasso estimator, which is efficiently solvable via the proximal gradient method. 

    \item A transformer can then be designed to approximate the above proximal gradient iterations. 
\end{itemize}
Throughout the proof, 
we let $\phi(x)$ denote the following sigmoid function:
\begin{align}\label{eq:def-func-phi}
	\phi(z) = \left(z+\frac12\right)\ind\left\{z+\frac12 > 0\right\} - \left(z-\frac12\right)\ind\left\{z-\frac12 > 0\right\},
\end{align}
which satisfies
$\lim_{z\to-\infty} \phi(z) = 0$, $\lim_{z\to\infty} \phi(z) = 1$, and $\phi(0) = 1/2$.

\paragraph{Step 1: constructing universal features for the target function class.}
In this step, we construct a finite collection of features to approximately represent $f(\bm{x})-f(\bm{0})$ in a linear fashion, with the aid of the sigmoid function defined in \eqref{eq:def-func-phi}.
Our construction is formally presented in the following lemma (along with its analysis), whose proof can be found in Appendix~\ref{subsec:proof-lem-discrete}.
\begin{lemma} \label{lem:discrete}
Consider any $\tau > 4$ and any $n \ge c_0\log|\mathcal{N}_{\varepsilon}|$ for some large enough constant $c_0>0$. There exist a collection of functions $\phi^{\mathsf{feature}}_i: \mathbb{R}^d \rightarrow \mathbb{R}$ ($1\leq i\leq n$) such that: for every $f \in \mathcal{F}$ and ${\bm x} \in \mathcal{B}$, one has 
\begin{align}\label{eq:lem-discrete}
\bigg|f({\bm x}) - f(\bm{0}) - \frac{1}{n}\sum_{i = 1}^n\rho^{\star}_{f, i}\phi^{\mathsf{feature}}_i(\bm{x})\bigg|
\lesssim C_{\mathcal{F}}\bigg(\frac{1}{\tau} + \tau\varepsilon + \sqrt{\frac{\log|\mathcal{N}_{\varepsilon}|}{n}}\bigg) 
\end{align}
for some $f$-dependent coefficients  $\{\rho^{\star}_{f,i}\}_{1\leq i\leq n} \subset \mathbb{R}$ obeying
\begin{align}\label{eq:lem-discrete-2}
|f({\bm 0})| + \frac{1}{n}\sum_{i = 1}^n|\rho^{\star}_{f, i}| < 4C_{\mathcal{F}}.
\end{align}
Here, the functions  $\{\phi^{\mathsf{feature}}_i(\cdot)\}_{1\leq i\leq n}$ are given by
\begin{align}
\label{eq:defn-phi-feature-lemma}
\phi_{i}^{\mathsf{feature}}(\bm{x})=\phi\bigg(\tau\Big(\frac{1}{\|\bm{\omega}_{i}\|_{2}}\bm{\omega}_{i}^{\top}\bm{x}-t_{i}\Big)\bigg)
\end{align}
	for some $\{(t_i, {\bm \omega}_i)\}_{1 \le i \le n} \subset \mathbb{R}\times \mathbb{R}^{d}$ independent of any specific $f$, where $\phi(\cdot)$ is defined in \eqref{eq:def-func-phi}.   
\end{lemma}
\begin{remark} The specific choices of  $\{(t_i, {\bm \omega}_i)\}_{1 \le i \le n} $ can be found in the proof in Appendix~\ref{subsec:proof-lem-discrete}.  \end{remark}

In words, for any function $f\in \mathcal{F}$ and any $\bm{x}\in \mathcal{B}$, the quantity $f(\bm{x})-f(\bm{0})$ can be closely approximated by a linear combination of the features $\{\phi_i^{\mathsf{feature}}(\bm{x})\}$, where the $\ell_1$ norm of the linear coefficients is well-controlled. This reveals that $\{\phi_i^{\mathsf{feature}}(\cdot)\}$ can serve as general-purpose features capable of linearly representing arbitrary functions in the function class $\mathcal{F}$.
We remark here that the $f$-dependent coefficients $\{\rho_{f,i}^{\star}\}_{1\le i\le n}$ are not required to be positive.
For ${\bm \omega}_i$ and $t_i$, they are randomly sampled from a certain probability measure $\Lambda(\mathrm{d}t,\mathrm{d} {\bm \omega})$ as demonstrated in the proof of Lemma \ref{lem:discrete}, which depend solely on the function class $\mathcal{F}$ and can provably approximate each $f$ with high probability. 
In the rest of the proof, we shall take 
%
\begin{align}\label{eq:eps-dis}
\tau = 1/\sqrt{\varepsilon}
\qquad \text{and} \qquad
\varepsilon_{\mathsf{dis}} \coloneqq c_{\mathsf{dis}}C_{\mathcal{F}}\bigg(\sqrt{\varepsilon} + \sqrt{\frac{\log|\mathcal{N}_{\varepsilon}|}{n}}\bigg)
\end{align}
for some large enough constant $c_{\mathsf{dis}}>0$, which allow one to obtain (see Lemma~\ref{lem:discrete})
\begin{align}
\label{eq:f-approx-epsilon-dis}
\bigg|f({\bm x}) - f(\bm{0}) - \frac{1}{n}\sum_{i = 1}^n\rho^{\star}_{f, i}\phi^{\mathsf{feature}}_i(\bm{x})\bigg|
\le \varepsilon_{\mathsf{dis}}
\qquad 
\text{for every } f\in\mathcal{F} \text{ and  } \bm{x}\in \mathcal{B}. 
\end{align}

\paragraph{Step 2: learning linear coefficients in-context via Lasso.}
Armed with the general-purpose features $\{\phi_i^{\mathsf{feature}}(\cdot)\}$, we now proceed to show how the linear coefficients $\{\rho^{\star}_{f,i}\}$ in \eqref{eq:lem-discrete} can be approximately located in-context.

Recall from  Lemma~\ref{lem:discrete} that $\{\rho^{\star}_{f,i}\}$ satisfy some $\ell_1$-norm constraint (cf.~\eqref{eq:lem-discrete-2}). With this in mind, 
we attempt estimating $\{\rho^{\star}_{f,i}\}$ from the in-context demonstration $\{(\bm{x}_i,y_i)\}_{1\leq i\leq N}$ by means of the following regularized estimator (a.k.a.~the Lasso estimator): 
%
%
\begin{align}\label{eq:opt-regularized}
\mathop{\text{minimize}}\limits_{\bm{\rho}\in \mathbb{R}^{n+1}}~~\ell({\bm \rho}) \coloneqq  \frac{1}{N}\sum_{i = 1}^N (y_i - {\bm \phi}_i^{\top}{\bm \rho})^2 + \lambda \|{\bm \rho}\|_1,
\end{align}
where $\lambda$ denotes the regularized parameter to be suitably chosen, and  
${\bm \phi}_i \in \mathbb{R}^{n+1}$ is defined as 
\begin{align}
\label{eq:defn-phi-i-features}
\bm{\phi}_{i}=\big[\phi_{1}^{\mathsf{feature}}(\bm{x}_i),\phi_{2}^{\mathsf{feature}}(\bm{x}_i),\cdots,\phi_{n}^{\mathsf{feature}}(\bm{x}_i),1\big]^{\top}.
\end{align}
%
%

In general, it is difficult to obtain an exact solution of \eqref{eq:opt-regularized}, which motivates us to analyze approximate solutions instead. More specifically, we would like to analyze the prediction risk of any $\widehat{\bm{\rho}}$ obeying
\begin{align}\label{eq:def-hatrho-opt}
\ell(\widehat{{\bm \rho}}) - \ell({{\bm \rho}}^{\star})\le \varepsilon_{\mathsf{opt}}
\end{align}
for some accuracy level $\varepsilon_{\mathsf{opt}}$, 
where ${\bm \rho}^{\star}\in \mathbb{R}^{n+1}$ collects the $f$-dependent coefficients $\rho^{\star}_{f,i}$  in Lemma \ref{lem:discrete} in the following way: 
\begin{align}\label{eq:def-rho-star}
{\bm \rho}^{\star} = \left[\frac{\rho^{\star}_{f,1}}{n},\cdots,\frac{\rho^{\star}_{f,n}}{n},\rho^{\star}_{f,0}\right]^{\top},\quad \mathrm{where}\quad 
\rho^{\star}_{f,0}= f({\bm 0}).
\end{align}
Note that in \eqref{eq:def-hatrho-opt}, we are comparing $\ell(\widehat{\bm{\rho}})$ with $\ell(\bm{\rho}^{\star})$ rather than that of the minimizer of \eqref{eq:opt-regularized}, as it facilitates our analysis.  
The following lemma quantifies the prediction error and the $\ell_1$ norm of any $\widehat{\bm \rho}$ obeying \eqref{eq:def-hatrho-opt}, whose proof is postponed to Appendix \ref{subsec:proof-lem-estimator}.

\begin{lemma} \label{lem:estimator}
Consider any given $\lambda \ge c_{\lambda}\left( \sqrt{\frac{\log N}{N}}\sigma + C_{\mathcal{F}}^{-1}\varepsilon_{\mathsf{dis}}^2\right)$ for some sufficiently large constant $c_{\lambda}>0$, where $\varepsilon_{\mathsf{dis}}$ is defined in \eqref{eq:eps-dis}. 
For any $\widehat{{\bm \rho}}$ that is statistically independent from $\bm{x}_{N+1}$ and satisfies \eqref{eq:def-hatrho-opt} with $\varepsilon_{\mathsf{opt}}\geq 0$, we have 
\begin{align}\label{eq:lem-estimator-1}
\mathbb{E}\Big[\big({\bm \phi}_{N+1}^{\top}\widehat{{\bm \rho}} - f({\bm x}_{N+1})\big)^2\Big]
&\lesssim \sqrt{\frac{\log N}{N}}\big(C_{\mathcal{F}}^2 + \lambda^{-2}\varepsilon_{\mathsf{opt}}^2 + \sigma\varepsilon_{\mathsf{dis}}\big) + \varepsilon_{\mathsf{dis}}^2 + \lambda C_{\mathcal{F}} + \varepsilon_{\mathsf{opt}} \\
 \label{eq:lem-estimator-2}
\|\widehat{{\bm \rho}}\|_1 &\lesssim C_{\mathcal{F}} + \lambda^{-1}\varepsilon_{\mathsf{opt}}
\end{align}
with probability at least $1-O(N^{-10})$. Here, the expectation in \eqref{eq:lem-estimator-1} is taken over the randomness of $\bm{x}_{N+1}$.
\end{lemma}
\begin{remark}
In the statement of Lemma~\ref{lem:estimator}, $\widehat{\bm{\rho}}$ is allowed to be statistically dependent on $\{(\bm{x}_i,y_i)\}_{1\leq i\leq N}$ but not on $\bm{x}_{N+1}$.  
\end{remark}


\paragraph{Step 3: solving the Lasso \eqref{eq:opt-regularized} via the inexact proximal gradient method.}


In light of Lemma~\ref{lem:estimator}, it is desirable to make the optimization error $\varepsilon_{\mathsf{opt}}$ as small as possible. Here, we propose to run the (inexact) proximal gradient method in an attempt to solve \eqref{eq:opt-regularized}. 
More precisely,  starting from the initialization ${\bm \rho}_0^{\mathsf{proximal}} = {\bm 0}$, the update rule for each iteration $t= 0,\dots, T$ is given by
\begin{subequations}
\begin{align}
{\bm \rho}_{t+1}^{\mathsf{proximal}} &= \mathsf{prox}^{\mathsf{proximal}}_{\eta\lambda\|\cdot\|_1}\Big({\bm \rho}_t^{\mathsf{proximal}} + \frac{2\eta}{N}\sum_{i = 1}^N (y_i - {\bm \phi}_i^{\top}{\bm \rho}_t^{\mathsf{proximal}}){\bm \phi}_i\Big) + {\bm e}_{t+1} \label{eq:update-rule-PGD-prox}\\
&= \mathsf{ST}_{\eta\lambda}\Big({\bm \rho}_t^{\mathsf{proximal}} + \frac{2\eta}{N}\sum_{i = 1}^N (y_i - {\bm \phi}_i^{\top}{\bm \rho}_t^{\mathsf{proximal}}){\bm \phi}_i\Big) + {\bm e}_{t+1},\label{eq:update-rule-PGD}
\end{align}
\end{subequations}
where $\eta>0$ stands for the stepsize, and we have included an additive term $\bm{e}_{t+1}$ that allows for inexact updates.  
Here, the proximal operator $\mathsf{prox}_{\eta\lambda\|\cdot\|_1}(\cdot)$ and the soft thresholding operator $\mathsf{ST}_{\eta\lambda}(\cdot)$ are given respectively by
\begin{subequations}
\begin{align}
\mathsf{prox}_{\eta\lambda\|\cdot\|_1}^{\mathsf{proximal}}({\bm x}) &\coloneqq \arg\min_{\bm \rho}\left\{\frac12\|{\bm x}-{\bm \rho}\|_2^2 + \eta\lambda\|{\bm \rho}\|_1 \right\}, \\
\mathsf{ST}_{\eta\lambda}(z) &\coloneqq \mathsf{sign}(z)\max\{|z|-\eta\lambda,0\}. 
\end{align}
\end{subequations}
Note that $\mathsf{ST}_{\eta\lambda}(\cdot)$ is applied entrywise in \eqref{eq:update-rule-PGD}.  

We now develop convergence guarantees for the above proximal gradient method. 
It can be shown that: 
after $T=(L-1)/2$ iterations, the (inexact) proximal gradient method \eqref{eq:update-rule-PGD} produces an iterate ${{\bm \rho}}_{T}^{\mathsf{proximal}}$ enjoying the following performance guarantees; the proof is postponed to Appendix~\ref{subsec:proof-lem-convergence}.
\begin{lemma} \label{lem:convergence}
Take $T=(L-1)/2$. 
Assume that 
\begin{align}
{\bm \rho}_0^{\mathsf{proximal}} = {\bm 0}, 
~~ 
\|{\bm \rho}^{\star}\|_1\lesssim C_{\mathcal{F}},~~
\lambda\gtrsim \sqrt{\frac{\log N}{N}}(C_{\mathcal{F}} + \sigma),~~ 
\eta = \frac{1}{2n},~~  
\text{and} ~~\|{\bm e}_t\|_1 \le \varepsilon_{\mathsf{approx}}\lesssim C_{\mathcal{F}} ~\text{for all }t \le T.
\end{align}
Then with probability at least $1-O(N^{-10})$, the output of the algorithm  \eqref{eq:update-rule-PGD} at the $T$-th iteration satisfies
\begin{align} \label{eq:convergence-rho}
\ell({\bm \rho}_T^{\mathsf{proximal}}) \le \ell({\bm \rho}^{\star}) + \frac{c_1nC_{\mathcal{F}}^2}{L} + c_1(L+n)\varepsilon_{\mathsf{approx}}\bigg(C_{\mathcal{F}} + \sigma + \max_{1\le k\le T}\|{\bm \rho}_{k}^{\mathsf{proximal}}\|_1 + \lambda\bigg)
\end{align}
for some universal constant  $c_1>0$, and for every $t \le T$ we have
\begin{align}\label{eq:lem-convergence-bound-rho}
\|{\bm \rho}_t^{\mathsf{proximal}}\|_1 \lesssim
C_{\mathcal{F}}  + \frac{nC_{\mathcal{F}}^2}{t\lambda} + \sqrt{N}(t+n)\max_{1\le k\le t}\|{\bm e}_{k}\|_1 + \frac{t+n}{\lambda}\max_{1\le k\le t}\big\{\|{\bm e}_k\|_1\|{\bm \rho}_k^{\mathsf{proximal}}\|_1\big\}.
\end{align}
\end{lemma}
\begin{remark}
The careful reader might remark that the upper bounds in Lemma~\ref{lem:convergence} appear somewhat intricate, as they depend on the $\ell_1$ norm of the previous iterates. Fortunately, once $\{\|\bm{e}_k\|_1\}$ are determined, we can apply mathematical induction to derive more concise bounds --- an approach to be carried out in Step 4. 
\end{remark}

\paragraph{Step 4: constructing the transformer  to emulate proximal gradient iterations.}
To build a transformer with favorable in-context learning capabilities, our design seeks to approximate the above proximal gradient iterations, which we elucidate in this step.

Let us begin by describing the input structure for each layer of our constructed transformer. 
For the $l$-th hidden layer ($0\leq l\leq L$), 
the input matrix $\bm{H}^{(l)}$ (see \eqref{eq:transformer-structure-defn}) takes the following form: 
\begin{align}
\label{eq:construction-Hl}
\bm{H}^{(l)} = \left( \begin{array}{ccccc}
{\bm x}_1^{(l)} & {\bm x}_2^{(l)} & \ldots & {\bm x}_N^{(l)} & {\bm x}_{N+1}^{(l)} \\
y_1^{(l)} & y_2^{(l)} & \ldots & y_N^{(l)} & 0 \\
{w}_1^{(l)} & {w}_2^{(l)} & \ldots & {w}_N^{(l)} & {w}_{N+1}^{(l)} \\
{\bm \phi}_1^{(l)} & {\bm \phi}_2^{(l)} & \ldots & {\bm \phi}_N^{(l)} & {\bm \phi}_{N+1}^{(l)} \\
{\bm \rho}^{(l)} & {\bm \rho}^{(l)} & \ldots & {\bm \rho}^{(l)} & {\bm \rho}^{(l)} \\
\lambda^{(l)} & \lambda^{(l)} & \ldots & \lambda^{(l)} & \lambda^{(l)} \\
\widehat{y}^{(l)} & \widehat{y}^{(l)} & \ldots & \widehat{y}^{(l)} & \widehat{y}^{(l)}
\end{array} \right)
\in \mathbb{R}^{(d+2n+7) \times (N+1)}
\end{align}
where 
$${\bm x}_i^{(l)}\in\mathbb{R}^{d+1}, \qquad {\bm \phi}_i^{(l)}, {\bm \rho}^{(l)} \in \mathbb{R}^{n+1}, \qquad \text{and}\qquad w_i^{(l)}, \lambda^{(l)}, \widehat{y}^{(l)} \in \mathbb{R}
\qquad \text{for all }1\leq i\leq N+1.
$$
Note that the last three row blocks in \eqref{eq:construction-Hl} contain $N+1$ identical copies of 
$\bm{\rho}^{(l)}$, $\lambda^{(l)}$ and $\widehat{y}^{(l)}$.   
In particular, $\bm{H}^{(0)}$ admits a simpler form,  
for which we initialize as follows: 
\begin{subequations}
\begin{align}
    {\bm x}_i^{(0)} &= [{\bm x}_i^{\top},1]^{\top}, ~~ \bm{\phi}_i^{(0)}=\bm{0} \quad &&\text{for all }1\leq i\leq N+1,  \\
    y_i^{(0)}&=y_i, ~~ w_i^{(0)}= 1 
    \quad &&\text{for all }1\leq i\leq N, \label{eq:edf-initial-yw}\\
     y_{N+1}^{(0)}& = 0,~~w_{N+1}^{(0)}= 0, ~~
    \bm{\rho}^{(0)}=\bm{0}, ~~ \lambda^{(0)}= \widehat{y}^{(0)}=0.\label{eq:edf-initial-wrholamyhat}
\end{align}
\end{subequations}
%
As a result, the input matrix ${\bm H}^{(0)}$ can be simply expressed as
\begin{align}
\label{eq:H0-input}
\bm{H}^{(0)} = \left( \begin{array}{ccccc}
{\bm x}_1 & {\bm x}_2 & \ldots & {\bm x}_N & {\bm x}_{N+1} \\
1 & 1 & \ldots & 1 & 1
\\
y_1 & y_2 & \ldots & y_N & 0 \\
1 & 1 & \ldots & 1 & 0 \\
{\bm 0} & {\bm 0} & \ldots & {\bm 0} & {\bm 0} 
\end{array} \right)
\in \mathbb{R}^{(d+2n+7) \times (N+1)}.
\end{align}

\begin{figure}[t]
    \centering
\includegraphics[width=0.95\linewidth]{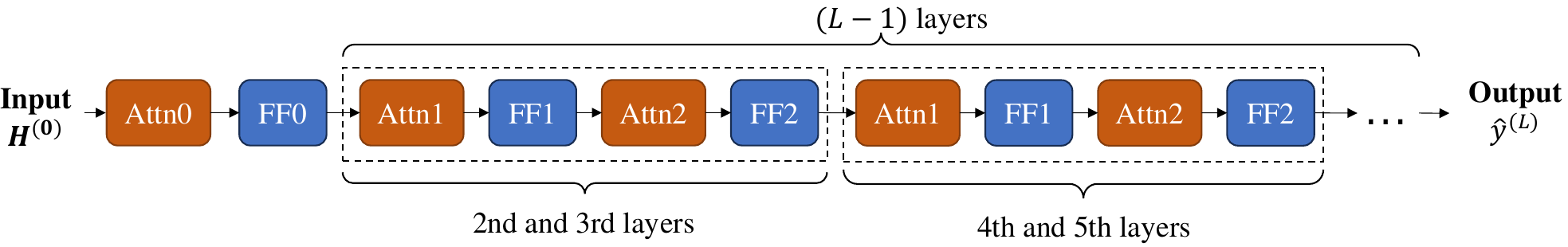}
    \caption{Structure of the desirable transformer.}
    \label{fig:transformer}
\end{figure}

Based on the above input structure of ${\bm H}^{(l)}$, we are positioned to present our construction, whose proof is postponed to Appendix \ref{subsec:proof-lem-opt}.
\begin{lemma}\label{lem:opt}
One can construct a transformer such that: 
\begin{itemize}
    \item[i)] it has $L$ layers, $M=O(1)$ attention heads per layer, and takes the matrix ${\bm H}^{(0)}$ (cf.~\eqref{eq:H0-input}) as input;
\item[ii)] the component ${\bm \rho}^{(L)}$ in the final output matrix ${\bm H}^{(L)}$ coincides with ${\bm \rho}_{(L-1)/2}^{\mathsf{proximal}}$ (cf.~\eqref{eq:update-rule-PGD}) for some $\bm{e}_t$, where we choose  
\begin{subequations}
\begin{align}
\lambda &\asymp \Big(\frac{\log N}{N}\Big)^{1/6} C_{\mathcal{F}}^{-1/3}\widehat{\varepsilon}^{2/3} + \sqrt{\frac{\log N}{N}}\Big(C_{\mathcal{F}} + \sigma\Big) + C_{\mathcal{F}}^{-1}\varepsilon_{\mathsf{dis}}^2,\label{eq:lambda}\\
\max_{1\le t\le T}\|{\bm e}_t\|_1&\lesssim \frac{C_{\mathcal{F}}}{(L+n)nN},
\end{align}
with 
\begin{align}\label{eq:def-hat-eps}
\widehat{\varepsilon} \coloneqq  \sqrt{\frac{\log N}{N}}C_{\mathcal{F}}(\sigma + C_{\mathcal{F}})
+\varepsilon_{\mathsf{dis}}^2 + \frac{nC_{\mathcal{F}}^2}{L};
\end{align}
\end{subequations}
\item[iii)]  for every $f\in\mathcal{F}$, the components ${\bm \rho}^{(L)}$ and $\widehat{y}^{(L)}$ in the final output matrix ${\bm H}^{(L)}$ (cf.~\eqref{eq:construction-Hl}) satisfy
\begin{subequations}
\begin{align}
\ell({\bm \rho}^{(L)}) - \ell({\bm \rho}^{\star}) &\lesssim \sqrt{\frac{\log N}{N}}C_{\mathcal{F}}(\sigma + C_{\mathcal{F}})
+\varepsilon_{\mathsf{dis}}^2 + \frac{nC_{\mathcal{F}}^2}{L}\label{eq:lem-opt}\\
\big({\bm \phi}_{N+1}^{\top}{\bm \rho}^{(L)} - \widehat{y}^{(L)}\big)^2&\lesssim  \sqrt{\frac{\log N}{N}}C_{\mathcal{F}}^2\label{eq:approx-haty}
\end{align}
\end{subequations}
with probability at least $1-O(N^{-10})$. 
\end{itemize}
\end{lemma}
In the proof of Lemma~\ref{lem:opt}, we construct a single multi-layer transformer whose structure is depicted in Figure~\ref{fig:transformer}.
It contains an attention layer \textbf{Attn0} and a feedforward layer \textbf{FF0}, which are designed for proper initialization, and they are followed by a sequence of repetitive blocks, each of which contains two attention layers and two feed-forward layers. 
Lemma~\ref{lem:opt} demonstrates such a simple structure allows one to emulate the iterations of the proximal gradient method (cf.~\eqref{eq:update-rule-PGD}) and achieves high optimization accuracy (see \eqref{eq:lem-opt}), while in the meantime controlling the fitting error between ${\bm \phi}_{N+1}^{\top}{\bm \rho}^{(L)}$ and $ \widehat{y}^{(L)}$ (see \eqref{eq:approx-haty}). And all this holds simultaneously for all $f$ in the function class of interest. 
More details of the designed transformer can be found in  
Appendix~\ref{subsec:proof-lem-opt}.

\paragraph{Step 5: putting everything together.} 
Equipped with the above results, we are ready to put all pieces together towards establishing our theory.

In view of \eqref{eq:lem-opt}, the output ${\bm \rho}^{(L)}$ of our constructed transformer in Lemma~\ref{lem:opt} 
satisfies 
\begin{align}\label{eq:eps-opt}
\ell({{\bm \rho}^{(L)}}) 
- \ell\left({\bm \rho}^{\star}\right)
&\le \varepsilon_{\mathsf{opt}},\notag\\
\text{by taking }\quad {\varepsilon}_{\mathsf{opt}} &\asymp \sqrt{\frac{\log N}{N}}C_{\mathcal{F}}(\sigma + C_{\mathcal{F}})
+\varepsilon_{\mathsf{dis}}^2 + \frac{nC_{\mathcal{F}}^2}{L}.
\end{align}
Invoking Lemma~\ref{lem:estimator} with $\widehat{\bm \rho} = {\bm \rho}^{(L)}$ reveals the following bound concerning the estimate ${\bm \rho}^{(L)}$:
\begin{align}\label{eq:MSE-1}
\mathbb{E}\big[\big({\bm \phi}_{N+1}^{\top}{{\bm \rho}}^{(L)} - f({\bm x}_{N+1})\big)^2\big]
&\overset{\text{(a)}}{\lesssim} \sqrt{\frac{\log N}{N}}\big(C_{\mathcal{F}}^2 + \lambda^{-2}\varepsilon_{\mathsf{opt}}^2 + \sigma C_\mathcal{F}\big) + \varepsilon_{\mathsf{dis}}^2 + \lambda C_{\mathcal{F}} + \varepsilon_{\mathsf{opt}}\notag\\
&\overset{\text{(b)}}{\lesssim} \sqrt{\frac{\log N}{N}}C_{\mathcal{F}}(C_{\mathcal{F}} + \sigma) + C_{\mathcal{F}}^2\varepsilon + \frac{C_{\mathcal{F}}^2\log|\mathcal{N}_\varepsilon|}{n}\notag\\
&\quad + \varepsilon_{\mathsf{opt}} + \lambda C_{\mathcal{F}} + \sqrt{\frac{\log N}{N}}\lambda^{-2}\varepsilon_{\mathsf{opt}}^2\notag\\
&\overset{\text{(c)}}{\lesssim} \sqrt{\frac{\log N}{N}}C_{\mathcal{F}}(C_{\mathcal{F}} + \sigma) + C_{\mathcal{F}}^2\varepsilon + \frac{C_{\mathcal{F}}^2\log|\mathcal{N}_\varepsilon|}{n}\notag\\
&\quad + \frac{nC_{\mathcal{F}}^2}{L} + \lambda C_{\mathcal{F}} + \sqrt{\frac{\log N}{N}}\lambda^{-2}\varepsilon_{\mathsf{opt}}^2
\end{align}
holds with probability at least $1-O(N^{-10})$, 
where in (a) we have used  \eqref{eq:lem-estimator-1} and a basic bound $\varepsilon_{\mathsf{dis}}\lesssim C_\mathcal{F}$ that holds under our assumptions, (b) arises from \eqref{eq:eps-dis}, and (c) results from \eqref{eq:eps-opt}.
Next, let us bound the last two terms in \eqref{eq:MSE-1}. 
\begin{itemize}
\item To begin with, 
with $\lambda$ as specified in \eqref{eq:lambda}, 
we obtain
\begin{align}
\lambda C_{\mathcal{F}}
&\overset{\text{}}{\lesssim} \Big(\frac{\log N}{N}\Big)^{1/6} C_{\mathcal{F}}^{2/3}\widehat{\varepsilon}^{2/3} + \sqrt{\frac{\log N}{N}}C_{\mathcal{F}}\big(C_{\mathcal{F}} + \sigma\big) + \varepsilon_{\mathsf{dis}}^2\notag\\
&\overset{\text{(a)}}{\lesssim} \sqrt{\frac{\log N}{N}}C_{\mathcal{F}}\big(C_{\mathcal{F}} + \sigma\big) + \varepsilon_{\mathsf{dis}}^2 + \widehat{\varepsilon}\label{eq:bound-lambda-C}\\
&\overset{\text{(b)}}{\lesssim} \sqrt{\frac{\log N}{N}}C_{\mathcal{F}}\big(C_{\mathcal{F}} + \sigma\big)  + C_{\mathcal{F}}^2\varepsilon + \frac{C_{\mathcal{F}}^2\log|\mathcal{N}_\varepsilon|}{n}+ \widehat{\varepsilon},\label{eq:proof-young}
\end{align}
 where
(a) invokes Young's inequality to derive
\begin{align}\label{eq:proof-young-1}
\Big(\frac{\log N}{N}\Big)^{1/6} C_{\mathcal{F}}^{2/3}\widehat{\varepsilon}^{2/3}
\le\frac13\left(\Big(\frac{\log N}{N}\Big)^{1/6} C_{\mathcal{F}}^{2/3}\right)^3 +  \frac23(\widehat{\varepsilon}^{2/3})^{3/2} \le \frac13\sqrt{\frac{\log N}{N}} C_{\mathcal{F}}^{2} + \frac23\widehat{\varepsilon},
\end{align}
and (b) applies \eqref{eq:eps-dis}.
\item 
Moreover, the last term in \eqref{eq:MSE-1} satisfies
\begin{align}\label{eq:proof-lambda-2epsopt}
\sqrt{\frac{\log N}{N}}\lambda^{-2}\varepsilon_{\mathsf{opt}}^2&\overset{\text{(a)}}{\lesssim} \Big(\frac{\log N}{N}\Big)^{1/2}\Big(\frac{\log N}{N}\Big)^{-1/3} C_{\mathcal{F}}^{2/3}\widehat{\varepsilon}^{-4/3}\varepsilon_{\mathsf{opt}}^2
\overset{\text{(b)}}{\lesssim} \Big(\frac{\log N}{N}\Big)^{1/6} C_{\mathcal{F}}^{2/3}\widehat{\varepsilon}^{2/3}\nonumber\\
&\overset{\text{(c)}}{\lesssim}\sqrt{\frac{\log N}{N}} C_{\mathcal{F}}^{2} + \widehat{\varepsilon}. 
\end{align}
Here, (a) results from the fact that $\lambda^2\gtrsim \Big(\frac{\log N}{N}\Big)^{1/3} C_{\mathcal{F}}^{-2/3}\widehat{\varepsilon}^{2/3}$, 
(b) is valid since $\widehat{\varepsilon} \asymp \varepsilon_{\mathsf{opt}}$ (see \eqref{eq:def-hat-eps} and \eqref{eq:eps-opt}), 
whereas (c) makes use of~\eqref{eq:proof-young-1}.
\end{itemize}
Substituting \eqref{eq:proof-young} and \eqref{eq:proof-lambda-2epsopt} into \eqref{eq:MSE-1}, and recalling the definition of $\widehat{\varepsilon}$ in \eqref{eq:def-hat-eps}, we arrive at
\begin{align}
\mathbb{E}\big[\big({\bm \phi}_{N+1}^{\top}{\bm \rho}^{(L)} - f({\bm x}_{N+1})\big)^2\big]
&\lesssim \sqrt{\frac{\log N}{N}}C_{\mathcal{F}}(\sigma + C_{\mathcal{F}})
+C_{\mathcal{F}}^2\varepsilon + \frac{C_{\mathcal{F}}^2\log|\mathcal{N}_{\varepsilon}|}{n} +  \frac{nC_{\mathcal{F}}^2}{L} \label{eq:approx-fxN+1}
\end{align}
with probability exceeding $1-O(N^{-10})$,
where we have used the fact that (see \eqref{eq:def-hat-eps} and \eqref{eq:eps-dis})
\begin{align*}
   \widehat{\varepsilon} \lesssim  \sqrt{\frac{\log N}{N}}C_{\mathcal{F}}(\sigma + C_{\mathcal{F}})
+ C_{\mathcal{F}}^2\varepsilon + \frac{C_{\mathcal{F}}^2\log|\mathcal{N}_\varepsilon|}{n} +\frac{nC_{\mathcal{F}}^2}{L}.
\end{align*}
Combining \eqref{eq:approx-fxN+1} with \eqref{eq:approx-haty} and recalling $\widehat{y}_{N+1} = \widehat{y}^{(L)}$ (cf.~\eqref{eq:y-hat-N-plus-1-readout}), we reach
\begin{align*}
\mathbb{E}\big[\big(\widehat{y}_{N+1} - f({\bm x}_{N+1})\big)^2\big]
&\le 2\mathbb{E}\big[\big(\widehat{y}_{N+1} - {\bm \phi}_{N+1}^{\top}{\bm \rho}^{(L)}\big)^2\big] + 2\mathbb{E}\big[\big({\bm \phi}_{N+1}^{\top}{\bm \rho}^{(L)} - f({\bm x}_{N+1})\big)^2\big]\notag\\
&\lesssim \sqrt{\frac{\log N}{N}}C_{\mathcal{F}}^2 + \sqrt{\frac{\log N}{N}}C_{\mathcal{F}}(\sigma + C_{\mathcal{F}})
+C_{\mathcal{F}}^2\varepsilon + \frac{C_{\mathcal{F}}^2\log|\mathcal{N}_{\varepsilon}|}{n} +  \frac{nC_{\mathcal{F}}^2}{L} \notag\\
&\asymp \sqrt{\frac{\log N}{N}}C_{\mathcal{F}}(\sigma + C_{\mathcal{F}})
+C_{\mathcal{F}}^2\varepsilon + \frac{C_{\mathcal{F}}^2\log|\mathcal{N}_{\varepsilon}|}{n} +  \frac{nC_{\mathcal{F}}^2}{L}.
\end{align*}
We can therefore conclude the proof of  Theorem~\ref{thm:main} by taking $\varepsilon \le \sqrt{\log N/N} + n/L$.


\section{Discussion}
\label{sec:discussion}

Through the lens of universal function approximation, this paper has investigated the in-context learning capabilities of transformers for statistical learning tasks,  
establishing approximation guarantees that extend far beyond the previously studied linear regression settings or the problems of learning linear functions. 
We have demonstrated that: for a fairly general function class $\mathcal{F}$ satisfying mild Fourier-type (or Barron-style) conditions, one can construct a universal multi-layer transformer achieving the following intriguing property:  for every task represented by some function $f\in\mathcal{F}$, 
the constructed transformer can readily utilize a few input-output examples to achieve vanishingly small in-context prediction risk, provided that the dimension and parameters of the transformers are suitably chosen. 
%
%
Our analysis imposes only fairly mild assumptions on $\mathcal{F}$, requiring neither linearity in the function class nor convexity in the learning problem, thereby offering improved theoretical understanding for the empirical success of transformer-based models in real-world statistical learning tasks.

Looking forward, we would like to establish tighter performance bounds on the prediction risk, particularly with respect to its dependence on the number of examples $N$ and the number of layers $L$. In addition, while our current analysis has focused on the approximation ability of a universal transformer, understanding how pretraining or finetuning affects generalization in in-context learning remains an open question, which we leave for future studies. Finally, the current paper is concerned with constructed approximation, and it would be of great interest to demystify end-to-end training dynamics for transformers, which are highly nonconvex and call for innovative technical ideas.


\section*{Acknowledgments}

G.~Li is supported in part by the Chinese University of Hong Kong Direct Grant for Research and the Hong Kong Research Grants Council ECS 2191363.
Y.~Wei is supported in part by the NSF 
CAREER award DMS-2143215 and the NSF grant CCF-2418156.  
Y.~Chen is supported in part by the Alfred P.~Sloan Research Fellowship,  the ONR grants N00014-22-1-2354 and N00014-25-1-2344, 
the Wharton AI \& Analytics Initiative's AI Research Fund, 
and the Amazon Research Award. 

\appendix

\section{Proof of key lemmas}

In this section, we provide complete proofs of the key lemmas introduced in Section~\ref{sec:analysis}.

\subsection{Proof of Lemma~\ref{lem:discrete}}
\label{subsec:proof-lem-discrete}

For notational convenience, we shall write the Fourier transform $F_f$ as
\begin{equation}
    F_f(\bm{\omega}) = |F_f(\bm{\omega})| e^{j \theta_f(\bm{\omega})}, 
\end{equation}
with $\theta_f(\bm{\omega})$ representing the angle.
By virtue of the Fourier transform of $f({\bm x})$, we have
\begin{align}
f({\bm x}) -f(\bm{0})
&=   \int_{\bm{\omega}} e^{j{\bm{\omega}}^{\top}{\bm x}} F_f({\bm{\omega}})\mathrm{d} {\bm{\omega}}
- \int_{\bm{\omega}} F_f({\bm{\omega}})\mathrm{d} {\bm{\omega}}
\notag\\
&=  \int_{\bm{\omega}} \big(e^{j{\bm{\omega}}^{\top}{\bm x}}e^{j\theta_f(\bm{\omega})} - e^{j\theta_f(\bm{\omega})}\big)|F_f({\bm{\omega}})|\mathrm{d} {\bm{\omega}}\notag\\
&\overset{\text{(a)}}{=} \int_{\bm{\omega}\ne \bm{0}}\Big(\cos\big({\bm{\omega}}^{\top}{\bm x} + \theta_f(\bm{\omega})\big) - \cos\big(\theta_f(\bm{\omega})\big)\Big)|F_f({\bm{\omega}})| \mathrm{d}\bm{\omega}, \label{eq:Fourier}
\end{align}
where (a) follows since $f({\bm x})$ is real-valued.
In view of \eqref{eq:Fourier}, we would like to construct a finite collection of features to approximately represent $f(\bm{x})-f(\bm{0})$, using the logistic function defined in \eqref{eq:def-func-phi}.

Towards this end, we first introduce a quantity related to the difference between $\phi(\tau x)$ and the unit step function $\ind(x > 0)$ as follows: 
\begin{align}
\delta_{\tau} \coloneqq \inf_{0 < \varepsilon \le 1/2} \Big\{2\varepsilon + \sup_{|x| \ge \varepsilon} \big|\phi(\tau x) - \ind(x > 0) \big|\Big\}.
\label{eq:defn-delta-tau}
\end{align}
The proof of Lemma \ref{lem:discrete} relies heavily upon the following result, whose proof is postponed to Appendix~\ref{sec:proof:lem:expression}.
\begin{lemma} \label{lem:expression}
For any $f \in \mathcal{F}$, any $\bm{x}$ obeying $\|\bm{x}\|_2\le 1$, and any $\tau > 2$, we have 
\begin{align}
\big|f({\bm x}) - f(\bm{0}) - f^{\mathsf{approx}}({\bm x}) \big| \le 3C_{\mathcal{F}}\delta_{\tau}, 
\label{eq:expression}
\end{align}
where for any $f\in \mathcal{F}$ we define
\begin{align}
f^{\mathsf{approx}}({\bm x})
\coloneqq \int_{{\bm \omega}\ne 0}\int_t \rho_f(t, {\bm \omega})\phi\big(\tau(\|{\bm \omega}\|_2^{-1}{\bm \omega}^{\top}{\bm x} - t)\big)\Lambda(\mathrm{d}t, \mathrm{d}{\bm \omega})
\label{eq:f-approx-x}
\end{align}
with $\phi(\cdot)$ defined in \eqref{eq:def-func-phi}. 
Here, $\Lambda(\mathrm{d}t, \mathrm{d}\bm{\omega})$ is a probability measure on $\mathbb{R}\times\mathbb{R}^d$ independent from $f$, while $\rho_f(t, \bm{\omega})$ denotes some weight function depending on $f, t, \bm{\omega}$ such that 
\begin{align}
\label{eq:rho-UB-lemma}
|\rho_f(t, \bm{\omega})| \le 3\Big(C_{\mathcal{F}} - \sup_{\tilde{f}\in\mathcal{F}}|\tilde{f}(\bm{0})|\Big)
\quad\text{and}\quad
\mathbb{E}_{(t, \bm{\omega}) \sim \Lambda}\big[|\rho_f(t, \bm{\omega})|\big] \le 3\Big(C_{\mathcal{F}} - \sup_{\tilde{f}\in\mathcal{F}}|\tilde{f}(\bm{0})|\Big).
\end{align}
\end{lemma}

Next, we would like to obtain a more succinct finite-sum approximation of the integration in \eqref{eq:f-approx-x} via random subsampling. 
Let us draw $n$ independent samples $(t_i, \bm{\omega}_i)$ ($1\leq i\leq n$) from the probability measure  $\Lambda(\mathrm{d}t, \mathrm{d}\bm{\omega})$.  
Applying the Hoeffding's inequality and the union bound over $\mathcal{N}_{\varepsilon}$ reveals that: with probability at least $3/4$, 
\begin{align*}
\bigg| \widehat{f}^{\mathsf{approx}}(\widehat{\bm x}) - \frac{1}{n}\sum_{i = 1}^n\rho_{\widehat{f}}(t_i, \bm{\omega}_i)\phi\big(\tau(\|\bm{\omega}_i\|_2^{-1}\bm{\omega}_i^{\top}\widehat{\bm x} - t_i)\big)\bigg|
\lesssim \widetilde{C}_{\mathcal{F}}\sqrt{\frac{\log|\mathcal{N}_{\varepsilon}|}{n}}
\end{align*}
holds simultaneously for all $(\widehat{f}, \widehat{\bm x}) \in \mathcal{N}_{\varepsilon}$, where $\widehat{f}^{\mathsf{approx}}(\widehat{\bm x})$ is defined in \eqref{eq:f-approx-x} with $f$ and $\bm{x}$ taken respectively to be $\widehat{f}$ and $\widehat{\bm{x}}$, and 
\begin{align}
\widetilde{C}_\mathcal{F} \coloneqq C_\mathcal{F} - \sup_{\tilde{f}\in\mathcal{F}}|\tilde{f}({\bm 0})|.
\end{align}
%
%
Note that the above application of the Hoeffding's inequality has used the following bounds for any fixed $\widehat{f}$ and $\widehat{\bm{x}}$:  
\begin{align*}
\mathbb{E}\big[\rho_{\widehat{f}}(t_i, \bm{\omega}_i)\phi\big(\tau(\|\bm{\omega}_i\|_2^{-1}\bm{\omega}_i^{\top}\widehat{\bm x} - t_i)\big)\big] &= \widehat{f}^{\mathsf{approx}}(\widehat{\bm x}), \\
\big|\rho_{\widehat{f}}(t_i, \bm{\omega}_i)\phi\big(\tau(\|\bm{\omega}_i\|_2^{-1}\bm{\omega}_i^{\top}\widehat{\bm x} - t_i)\big)\big| &\le 3\widetilde{C}_{\mathcal{F}}. 
\end{align*}
%

Similarly,  applying the Hoeffding's inequality and the union bound once again yields that: with probability at least $3/4$,
\begin{align*}
\bigg|\mathbb{E}_{(t, \bm{\omega}) \sim \Lambda}\big[\big|\rho_{\widehat{f}}(t, \bm{\omega})\big|\big] - \frac{1}{n}\sum_{i = 1}^n\big|\rho_{\widehat{f}}(t_i, \bm{\omega}_i)\big|\bigg|
\lesssim \widetilde{C}_{\mathcal{F}}\sqrt{\frac{\log|\mathcal{N}_{\varepsilon}|}{n}}
\end{align*}
holds simultaneously for all $(\widehat{f},\widehat{\bm x})\in\mathcal{N}_\varepsilon$,
where we have used the following bound:
\begin{align*}
\big|\rho_{\widehat{f}}(t_i, \bm{\omega}_i)\big| &\le 3\widetilde{C}_{\mathcal{F}}.
\end{align*}
%
Thus with probability at least $3/4$, one has
\begin{align*}
\frac{1}{n}\sum_{i = 1}^n\big|\rho_{\widehat{f}}(t_i, \bm{\omega}_i)\big| < \mathbb{E}_{(t, \bm{\omega}) \sim \Lambda}\big[\big|\rho_{\widehat{f}}(t, \bm{\omega})\big|\big] + \widetilde{C}_{\mathcal{F}}\sqrt{\frac{c_1\log|\mathcal{N}_{\varepsilon}|}{n}}
\overset{\text{(a)}}{\le} 3\widetilde{C}_{\mathcal{F}} + \widetilde{C}_{\mathcal{F}} = 4\widetilde{C}_{\mathcal{F}}
\end{align*}
simultaneously for all  $(\widehat{f}, \widehat{\bm x}) \in \mathcal{N}_{\varepsilon}$, 
where $c_1>0$ is some universal constant, and (a) is valid under the condition that $n\ge c_1\log |\mathcal{N}_{\varepsilon}|$.
This further shows that, with probability at least $3/4$, 
$$
|f({\bm 0})| + \frac{1}{n}\sum_{i = 1}^n\big|\rho_{\widehat{f}}(t_i, \bm{\omega}_i)\big| 
< |f({\bm 0})| + 4\widetilde{C}_\mathcal{F}
\le 4C_\mathcal{F}
$$
simultaneously for all $(\widehat{f},\widehat{\bm{x}})\in \mathcal{N}_{\varepsilon}$.

Having established several key properties for the epsilon-cover $\mathcal{N}_{\varepsilon}$, we now extend these properties to all $f\in \mathcal{F}$. 
For any $f \in \mathcal{F}$, there exists some $(\widehat{f}, \widehat{\bm x}) \in \mathcal{N}_{\varepsilon}$, such that for all $\|{\bm x}\|_2\le 1$,
\begin{align*}
&\bigg|f({\bm x})  - f(\bm{0}) -  \frac{1}{n}\sum_{i = 1}^n\rho_{\widehat{f}}(t_i, \bm{\omega}_i)\phi\big(\tau(\|\bm{\omega}_i\|_2^{-1}\bm{\omega}_i^{\top}{\bm x} - t_i)\big)\bigg|\notag\\
&\quad\le \big|f({\bm x}) - f({\bm 0}) - \widehat{f}(\widehat{\bm x}) + \widehat{f}({\bm 0}) \big| + \bigg|\widehat{f}(\widehat{\bm x}) - \widehat{f}(\bm{0}) -  \frac{1}{n}\sum_{i = 1}^n\rho_{\widehat{f}}(t_i, \bm{\omega}_i)\phi\big(\tau(\|\bm{\omega}_i\|_2^{-1}\bm{\omega}_i^{\top}\widehat{\bm x} - t_i)\big)\bigg|\notag\\
&\quad\quad + \bigg|\frac{1}{n}\sum_{i = 1}^n\left(\rho_{\widehat{f}}(t_i, \bm{\omega}_i)\phi\big(\tau(\|\bm{\omega}_i\|_2^{-1}\bm{\omega}_i^{\top}\widehat{\bm x} - t_i)\big)-\rho_{\widehat{f}}(t_i, \bm{\omega}_i)\phi\big(\tau(\|\bm{\omega}_i\|_2^{-1}\bm{\omega}_i^{\top}{\bm x} - t_i)\big)\right)\bigg|\notag\\
&\quad\overset{\text{(a)}}{\le} C_{\mathcal{F}}\Big(\varepsilon + 3\delta_{\tau} + \sqrt{\frac{c_1\log|\mathcal{N}_{\varepsilon}|}{n}} + 4\tau\varepsilon\Big),
\end{align*}
where $c_1$ is a universal constant, (a) uses the definition of the epsilon-cover \eqref{eq:defn-eps-cover}, and  
takes advantage of the fact that 
\begin{align*}
& \bigg|\frac{1}{n}\sum_{i = 1}^n\left(\rho_{\widehat{f}}(t_i, \bm{\omega}_i)\phi\big(\tau(\|\bm{\omega}_i\|_2^{-1}\bm{\omega}_i^{\top}\widehat{\bm x} - t_i)\big)-\rho_{\widehat{f}}( t_i, \bm{\omega}_i)\phi\big(\tau(\|\bm{\omega}_i\|_2^{-1}\bm{\omega}_i^{\top}{\bm x} - t_i)\big)\right)\bigg|\notag\\
&\quad= \frac{1}{n}\sum_{i = 1}^n\big|\rho_{\widehat{f}}(t_i, \bm{\omega}_i)\big|\Big|\phi\big(\tau(\|\bm{\omega}_i\|_2^{-1}\bm{\omega}_i^{\top}\widehat{\bm x} - t_i)\big)-\phi\big(\tau(\|\bm{\omega}_i\|_2^{-1}\bm{\omega}_i^{\top}{\bm x} - t_i)\big)\Big|\notag\\
&\quad\le \frac{1}{n}\sum_{i = 1}^n\big|\rho_{\widehat{f}}(t_i, \bm{\omega}_i)\big|\left|\tau\|\bm{\omega}_i\|_2^{-1}\bm{\omega}_i^{\top}(\widehat{\bm x}- {\bm x})\right|\notag\\
&\quad\le \frac{1}{n}\sum_{i = 1}^n\big|\rho_{\widehat{f}}(t_i, \bm{\omega}_i)\big|\tau\|\widehat{\bm x}- {\bm x}\|_2\le \frac{1}{n}\sum_{i = 1}^n\big|\rho_{\widehat{f}}(t_i, \bm{\omega}_i)\big|\tau \varepsilon\le 4C_{\mathcal{F}}\tau\varepsilon.
\end{align*}
Recalling that $\tau>4$, we arrive at
\begin{align}\label{eq:proof-lem-discrete-temp-1}
\bigg|f({\bm x}) - f(\bm{0}) -  \frac{1}{n}\sum_{i = 1}^n\rho_{\widehat{f}}(t_i, \bm{\omega}_i)\phi\big(\tau(\|\bm{\omega}_i\|_2^{-1}\bm{\omega}_i^{\top}{\bm x} - t_i)\big)\bigg|\lesssim C_{\mathcal{F}}\Big(\delta_{\tau} + \sqrt{\frac{\log|\mathcal{N}_{\varepsilon}|}{n}} + \tau\varepsilon\Big).
\end{align}

To finish up, it suffices to prove that $\delta_\tau\le 1/\tau$.
Recalling the definition of $\delta_\tau$ in \eqref{eq:defn-delta-tau} and the function $\phi(\tau x)$, we have $$|\phi(\tau x) - \ind(x>0)| = (1/2-\tau|x|)\ind\{\tau|x|\le 1/2\},$$ and as a result, 
$$
\sup_{|x|\ge \varepsilon'} \big|\phi(\tau x) - \ind(x>0) \big| = \sup_{|x|\ge \varepsilon'}(1/2-\tau|x|)\ind(\tau|x|\le 1/2) = \max\left\{\frac12 - \tau\varepsilon',0\right\}. 
$$
This reveals that
$$
\delta_\tau = \inf_{0\le \varepsilon'\le 1/2}\left\{2\varepsilon + \max\left\{\frac12 - \tau\varepsilon',0\right\}\right\}= \frac{1}{\tau} 
$$
for any $\tau \ge 2$.
Substituting it into \eqref{eq:proof-lem-discrete-temp-1} and recalling the definition of $\phi_i^{\mathsf{feature}}({\bm x})$ in \eqref{eq:defn-phi-feature-lemma}, we establish the claimed result \eqref{eq:lem-discrete} with $\rho_{f,i}^{\star} = \rho_{\widehat{f}}(t_i,{\bm \omega}_i)$.

\subsubsection{Proof of Lemma~\ref{lem:expression}}
\label{sec:proof:lem:expression}

%
%
%
%
It has been shown by~\citet[Lemma~5]{Barron1993Universal} that:  for any ${\bm \omega}\in\mathbb{R}^d$,  $\theta\in[0,2\pi)$ and any $\tau\ge 2$, there exists some function $\gamma_{{\bm \omega},\theta}(\cdot)$  such that
\begin{subequations}\label{eq:proof-lem-expression-g}
\begin{align}\label{eq:proof-lem-expression-temp-2}
\bigg|\frac{\cos(\bm{\omega}^{\top}{\bm x} + \theta) - \cos(\theta)}{\|\bm{\omega}\|_2} - \int_{t\in[-2, 1]} \gamma_{\bm{\omega}, \theta}(t)\phi\big(\tau(\|\bm{\omega}\|_2^{-1}\bm{\omega}^{\top}{\bm x} - t)\big)\mathrm{d}t\bigg| &\le 3\delta_{\tau}, \\
|\gamma_{\bm{\omega}, \theta}(t)| &\le 1. \label{eq:proof-lem-expression-int-gamma}
\end{align}
\end{subequations}
To make it self-contained, we will present the proof of \eqref{eq:proof-lem-expression-g} towards the end of this section. 
%
%
%
Based on the conclusion in \eqref{eq:proof-lem-expression-g}, we define an approximation of $f({\bm x}) - f({\bm 0})$ as
\begin{align}\label{eq:proof-lem-expression-barf}
\widetilde{f}^{\mathsf{approx}}({\bm x}) \coloneqq  \int_{\bm{\omega}\ne \bm{0}}\bigg( \int_{t\in[-2,1]} \phi\big(\tau(\|\bm{\omega}\|_2^{-1}\bm{\omega}^{\top}{\bm x} - t)\big)\gamma_{\bm{\omega}, \theta_f({\bm \omega})}(t)\mathrm{d}t \bigg)\|\bm{\omega}\|_2|F_f(\bm{\omega})|\mathrm{d}\bm{\omega}.
\end{align}
Combining \eqref{eq:proof-lem-expression-barf} and the Fourier transform of $f({\bm x}) - f(\bm{0})$ given in \eqref{eq:Fourier} yields 
\begin{align}
&\quad \bigg|f({\bm x}) - f(\bm{0}) - \widetilde{f}^{\mathsf{approx}}({\bm x})\bigg| \notag\\
&= \bigg| \int_{\bm{\omega}\ne \bm{0}}\Big(\frac{\cos\big({\bm{\omega}}^{\top}{\bm x} + \theta_f(\bm{\omega})\big) - \cos\big(\theta_f(\bm{\omega})\big)}{\|{\bm \omega}\|_2} - \int_{t\in[-2,1]} \phi\big(\tau(\|\bm{\omega}\|_2^{-1}\bm{\omega}^{\top}{\bm x} - t)\big)\gamma_{\bm{\omega}, \theta_f({\bm \omega})}(t)\mathrm{d}t\Big)\|{\bm \omega}\|_2|F_f({\bm{\omega}})| \mathrm{d}\bm{\omega} \bigg|\notag\\
&\overset{\text{(a)}}{\le} 3\delta_{\tau}\int_{\bm{\omega}} \|\bm{\omega}\|_2|F_f(\bm{\omega})|\mathrm{d}\bm{\omega}\overset{\text{(b)}}{\le} 3C_{\mathcal{F}}\delta_{\tau},
\label{eq:lem_expression}
\end{align}
where (a) inserts \eqref{eq:proof-lem-expression-temp-2}, and (b) applies the definition of $C_{\mathcal{F}}$ in \eqref{eq:condition-Fourier}.  

Next, let us define the following key quantity independent of $f$:
\begin{align*}
\Gamma_{\mathcal{F}} \coloneqq  3\int_{\bm{\omega}} \|{\bm \omega}\|_2F^{\mathsf{sup}}({\bm \omega})\mathrm{d}{\bm \omega},
\end{align*}
which can be bounded by
\begin{align}
\label{eq:UB-Gamma-F1}
\Gamma_{\mathcal{F}} \le 3\big(C_{\mathcal{F}} - \sup_{f\in\mathcal{F}}|f({\bm 0})|\big) < \infty
\end{align}
for any $f\in \mathcal{F}$. 
As a result,  
\begin{align}\label{eq:defn-Lambda}
\Lambda(\mathrm{d}t, \mathrm{d}{\bm \omega}) \coloneqq  \Gamma_{\mathcal{F}}^{-1}\ind(-2\le t\le 1)\,\|{\bm \omega}\|_2F^{\mathsf{sup}}({\bm \omega})\mathrm{d}t\mathrm{d}{\bm \omega}
\end{align} 
forms a valid probability measure on $\mathbb{R}\times\mathbb{R}^d$, given that $$\Lambda(\mathrm{d}t, \mathrm{d}{\bm \omega})\ge 0 \qquad \text{and}\qquad  
\int_{\bm \omega}\int_t\Lambda(\mathrm{d}t, \mathrm{d}{\bm \omega}) = 1.$$
Importantly, the probability measure  $\Lambda(\mathrm{d}t, \mathrm{d}{\bm \omega})$ allows one to show that the function
$\widetilde{f}^{\mathsf{approx}}({\bm x})$ (cf.~\eqref{eq:proof-lem-expression-barf}) coincides with the approximation $f^{\mathsf{approx}}({\bm x})$ defined in \eqref{eq:f-approx-x}:
\begin{align*}
\widetilde{f}^{\mathsf{approx}}({\bm x})
= \int_{{\bm \omega}\ne \bm{0}}\int_t \rho_f(t, {\bm \omega})\phi\big(\tau(\|{\bm \omega}\|_2^{-1}{\bm \omega}^{\top}{\bm x} - t)\big)\Lambda(\mathrm{d}t, \mathrm{d}{\bm \omega}) = f^{\mathsf{approx}}({\bm x}),
\end{align*}
where 
\begin{align}\label{eq:defn-rho-f}
\rho_f(t, {\bm \omega}) \coloneqq \frac{\gamma_{{\bm \omega}, \theta_f({\bm \omega})}(t)\|{\bm \omega}\|_2|F_f({\bm \omega})|\mathrm{d}t\mathrm{d}{\bm \omega}}{\Lambda(\mathrm{d}t, \mathrm{d}{\bm \omega})}.
\end{align}
Substitution into \eqref{eq:lem_expression} immediately establishes the advertised result  \eqref{eq:expression}.

To finish up, it suffices to note the following two facts: 
\begin{itemize}
\item
Combining \eqref{eq:defn-Lambda} and \eqref{eq:defn-rho-f}, we have
$$
|\rho_f(t, {\bm \omega})|\le 
\frac{|\gamma_{{\bm \omega}, \theta_f({\bm \omega})}(t)|\|{\bm \omega}\|_2|F_f({\bm \omega})|}{\Gamma_{\mathcal{F}}^{-1}\,\|{\bm \omega}\|_2F^{\mathsf{sup}}({\bm \omega})}
\overset{\text{(a)}}{\le} \Gamma_{\mathcal{F}}\overset{\text{(b)}}{\le} 3\Big(C_{\mathcal{F}} - \sup_{\tilde{f}\in\mathcal{F}}|\tilde{f}({\bm 0})| \Big),
$$
where (a) follows from \eqref{eq:proof-lem-expression-int-gamma} and (b) from \eqref{eq:UB-Gamma-F1}.
\item Substituting \eqref{eq:defn-rho-f} into the expectation of $|\rho_f(t, \bm{\omega})|$ gives
\begin{align*}
\mathbb{E}_{(t, \bm{\omega}) \sim \Lambda}\big[|\rho_f(t, \bm{\omega})|\big]  &=\int_{{\bm \omega}\ne \bm{0}}\int_t |\rho_f(t, {\bm \omega})|\Lambda(\mathrm{d}t, \mathrm{d}{\bm \omega})
= \int_{{\bm \omega}\ne \bm{0}}\int_{t\in[-2, 1]} \big|\gamma_{{\bm \omega}, \theta_f({\bm \omega})}(t)\big|\mathrm{d}t\|{\bm \omega}\|_2|
F_f({\bm \omega})|\mathrm{d}{\bm \omega}
\notag\\
&\overset{\text{(a)}}{\le} 3\int_{{\bm \omega}\ne \bm{0}}\|{\bm \omega}\|_2|
F_f({\bm \omega})|\mathrm{d}{\bm \omega}
\overset{\text{(b)}}{\le} 3\Big(C_{\mathcal{F}} - \sup_{\tilde{f}\in\mathcal{F}}|\tilde{f}({\bm 0})|\Big),
\end{align*}
where (a) uses \eqref{eq:proof-lem-expression-int-gamma} and (b) applies the definition of $C_{\mathcal{F}}$ in \eqref{eq:condition-Fourier}
\end{itemize}

\paragraph{Proof of relation~\eqref{eq:proof-lem-expression-g}.}
This proof is similar to the proof of Lemma 5 in \cite{Barron1993Universal}.
We begin by claiming the following result (to be proven shortly):
 for any function $g:[-1,1]\to \mathbb{R}$ satisfying $|g'(z)|\le 1$, $|g(-1)|\le 1$, and for any $\tau >0$,
 there exists a function $\gamma(\cdot)$ such that
 \begin{subequations}
 \label{eq:Barron-bound}
 \begin{align}
 \left|g(z) - \int_{t\in[-2, 1]}\gamma(t)\phi\big(\tau(z + t)\big)\mathrm{d}t\right|\le 3\delta_\tau,
 \end{align}
 where 
 \begin{align}
\int_{t\in[-2, 1]}|\gamma(t)|\mathrm{d} t\le 3.
 \end{align}
 \end{subequations}
 Here, $\delta_\tau$ and 
$\phi(\cdot)$ have been defined in \eqref{eq:defn-delta-tau} and \eqref{eq:def-func-phi}, respectively. 
Suppose for the moment that this claim is valid. 
To establish \eqref{eq:proof-lem-expression-g}, we find it convenient to introduce another auxiliary function 
$$g_{\bm{\omega},\theta}(z) \coloneqq  \frac{\cos(\|\bm{\omega}\|_2z + \theta) - \cos(\theta)}{\|\bm{\omega}\|_2}$$
  defined on $[-1,1]$, which clearly satisfies $$
  \big|g_{{\bm \omega},\theta}'(z)\big|\le 1\qquad \text{for all }z\in [-1,1],\qquad \text{and}\qquad |g_{{\bm \omega},\theta}(-1)|= |g_{{\bm \omega},\theta}(-1)-g_{{\bm \omega},\theta}(0)| \le 1.
  $$ 
Applying inequality \eqref{eq:Barron-bound} to 
$g_{{\bm \omega},\theta}(\cdot)$ and recognizing that $g_{{\bm \omega},\theta}\big(\frac{\bm{\omega}^{\top}\bm{x}}{\|\bm{\omega}\|_2}\big)=\frac{\cos(\bm{\omega}^{\top}{\bm x} + \theta) - \cos(\theta)}{\|\bm{\omega}\|_2}$ establish \eqref{eq:proof-lem-expression-g}.

To finish up, it remains to establish \eqref{eq:Barron-bound}. 
Define 
\begin{align}
    \gamma(t) = 
    \begin{cases}
        g'(t), & \text{if }t\in [-1,1], \\
        \mathsf{sign}\big(g(-1)\big),\qquad & \text{if }-1-|g(-1)|\le t < -1, \\
        0, & \text{else}. 
    \end{cases}
\end{align}
%
which clearly satisfies $|\gamma(t)| \le 1$.
Observe that for any $z\in[-1,1]$, it holds that
\begin{align*}
\int_{t\in[-2, 1]} \gamma(t)\ind(z-t>0)\mathrm{d} t = \int_{-2}^{z} \gamma(t)\mathrm{d} t = \int_{-1-|g(-1)|}^{-1}\mathsf{sign}\big(g(-1)\big)\mathrm{d} t + \int_{-1}^z g'(t)\mathrm{d} t = g(-1) + g(z) - g(-1) = g(z).
\end{align*}
Recalling the definition of $\delta_\tau$ in \eqref{eq:defn-delta-tau}, we arrive at
\begin{align*}
&\left|g(z) -  \int_{t\in[-2, 1]} \gamma(t)\phi\big(\tau(z-t)\big)\mathrm{d} t\right|
= \left|\int_{t\in[-2, 1]} \gamma(t)\big(\ind(z-t>0)-\phi\big(\tau(z-t)\big)\big)\mathrm{d} t\right|\notag\\
&\qquad \le\inf_{0<\varepsilon\le 1/2}\left\{ \int_{t:|z-t|\le \varepsilon}|\gamma(t)|\mathrm{d}t + \int_{t:|z-t|\ge \varepsilon} |\gamma(t)|\big|\ind(z-t>0)-\phi\big(\tau(z-t)\big)\big|\mathrm{d} t\right\}\notag\\
&\qquad \le \inf_{0<\varepsilon\le 1/2}\left\{2\varepsilon + 3\sup_{x:|x|\ge \varepsilon} \big|\phi(\tau x) - \ind(x>0) \big|\right\}\le 3\delta_\tau
\end{align*}
as claimed. 

\subsection{Proof of Lemma~\ref{lem:estimator}}
\label{subsec:proof-lem-estimator}

In view of \eqref{eq:lem-discrete-2} in Lemma~\ref{lem:discrete}, we have
$$\|{\bm \rho}^{\star}\|_1\lesssim C_{\mathcal{F}}.$$
A key step in this proof is to establish the following lemma, whose proof is postponed to Appendix~\ref{sec:proof:lem:empirical}.
\begin{lemma} \label{lem:empirical}
With probability at least $1-O(N^{-11})$, for any ${\bm \rho} \in \mathbb{R}^{n+1}$ (which can be statistically dependent on $\{(\bm{x}_i,y_i)\}_{1\leq i\leq N}$ but not on $\bm{x}_{N+1}$), we have
\begin{align}\label{eq:lem-empirical-1}
& \bigg|\frac{1}{N}\sum_{i = 1}^N \big[(y_i - {\bm \phi}_i^{\top}{\bm \rho})^2 - z_i^2\big] - \mathbb{E}\big[\big(f({\bm x}_{N+1}) - {\bm \phi}_{N+1}^{\top}{\bm \rho}\big)^2\big]\bigg| \notag\\
&\qquad \lesssim \sqrt{\frac{\log N}{N}}\big(\varepsilon_{\mathsf{dis}}^2 + \|{\bm \rho}-{\bm \rho}^{\star}\|_1^2 + \sigma(\varepsilon_{\mathsf{dis}} + \|{\bm \rho}-{\bm \rho}^{\star}\|_1)\big)
\end{align}
\begin{align}\label{eq:lem-empirical-2}
\text{and} 
\qquad 
\frac{1}{N}\sum_{i = 1}^N \big[(y_i - {\bm \phi}_i^{\top}{\bm \rho})^2 - (y_i - {\bm \phi}_i^{\top}{{\bm \rho}}^{\star})^2\big] \gtrsim  - \varepsilon_{\mathsf{dis}}^2-\sqrt{\frac{\log N}{N}}\sigma\|{\bm \rho}-{{\bm \rho}}^{\star}\|_1.
\end{align}
Here, the expectation in \eqref{eq:lem-empirical-1} is taken over the randomness of $\bm{x}_{N+1}$. 
\end{lemma}

Armed with this lemma, we can proceed to the proof of Lemma~\ref{lem:estimator}. 

\paragraph{Proof of \eqref{eq:lem-estimator-2}.}
Suppose for the moment that 
\begin{align}\label{eq:proof-lem-estimator-assu}
\|\widehat{{\bm \rho}}\|_1> 4\|{\bm \rho}^{\star}\|_1 +  C_{\mathcal{F}} + 4\lambda^{-1}\varepsilon_{\mathsf{opt}},
\end{align}
then it follows from Lemma~\ref{lem:empirical} that
\begin{align}
\ell(\widehat{{\bm \rho}}) - \ell({\bm \rho}^{\star}) &= \frac{1}{N}\sum_{i = 1}^N (y_i - {\bm \phi}_i^{\top}\widehat{{\bm \rho}})^2 + \lambda \|\widehat{{\bm \rho}}\|_1
- \frac{1}{N}\sum_{i = 1}^N (y_i - {\bm \phi}_i^{\top}{\bm \rho}^{\star})^2 - \lambda \|{\bm \rho}^{\star}\|_1 \notag \\
&\overset{\text{(a)}}{\ge} \lambda \|\widehat{{\bm \rho}}\|_1 - \lambda \|{\bm \rho}^{\star}\|_1 - 
C\varepsilon_{\mathsf{dis}}^2 - C\sqrt{\frac{\log N}{N}}\sigma \|\widehat{\bm \rho}-{\bm \rho}^{\star}\|_1 \notag\\
&\overset{\text{(b)}}{>} \lambda \|\widehat{\bm \rho}\|_1 - \frac{\lambda\|\widehat{\bm \rho}\|_1}{4} - 
C\varepsilon_{\mathsf{dis}}^2 - 2C\sqrt{\frac{\log N}{N}}\sigma \|\widehat{\bm \rho}\|_1
\label{eq:l-rhohat-rho-star-LB1}
\end{align}
for some universal constant $C>0$. Here, (a) results from \eqref{eq:lem-empirical-2}, whereas (b) invokes the properties (see \eqref{eq:proof-lem-estimator-assu}) that $\|\widehat{\bm \rho}\|_1 > 4\|{\bm \rho}^{\star}\|_1$ and $\|\widehat{\bm \rho} - {\bm \rho}^{\star}\|_1 \le \|\widehat{\bm \rho}\|_1 + \|{\bm \rho}^{\star}\|_1 < 2\|\widehat{\bm \rho}\|_1$.
In addition, under our assumption on $\lambda$, 
we have 
\begin{align}
\label{eq:lambda-LB-1357}
\lambda \ge 4C C_{\mathcal{F}}^{-1}\varepsilon_{\mathsf{dis}}^2 + 8C\sqrt{\log N/N}\sigma
\end{align}
as long as $c_{\lambda}$ is large enough, which then gives
\begin{align*}
C\varepsilon_{\mathsf{dis}}^2\le \frac{\lambda C_{\mathcal{F}}}{4} <  \frac{\lambda \|\widehat{\bm \rho}\|_1}{4}\qquad \text{and}\qquad 
2C\sqrt{\frac{\log N}{N}}\sigma \|\widehat{\bm \rho}\|_1 < \frac{\lambda \|\widehat{\bm \rho}\|_1}{4}.
\end{align*}
These combined with \eqref{eq:l-rhohat-rho-star-LB1} result in
\begin{align*}
\ell(\widehat{\bm \rho}) - \ell({\bm \rho}^{\star})
>\lambda \|\widehat{\bm \rho}\|_1 - \frac{\lambda\|\widehat{\bm \rho}\|_1}{4} - 
\frac{\lambda\|\widehat{\bm \rho}\|_1}{4} - \frac{\lambda\|\widehat{\bm \rho}\|_1}{4}
=
\frac{\lambda}{4} \|\widehat{\bm \rho}\|_1 > \varepsilon_{\mathsf{opt}}
\end{align*}
with the last inequality due to \eqref{eq:proof-lem-estimator-assu}. This, however, contradicts the $\varepsilon_{\mathsf{opt}}$-optimality of $\widehat{\bm \rho}$, which in turn justifies that the assumption \eqref{eq:proof-lem-estimator-assu} cannot possibly hold. As a result, we can conclude that
\begin{align}
\|\widehat{\bm \rho}_1\|\le 4\|{\bm \rho}^{\star}\|_1 +  C_{\mathcal{F}} + 4\lambda^{-1}\varepsilon_{\mathsf{opt}}\asymp C_{\mathcal{F}} + \lambda^{-1}\varepsilon_{\mathsf{opt}},
\label{eq:rho-hat-UB-123}
\end{align}
where the last relation is valid since $\|{\bm \rho}^{\star}\|_1\lesssim C_{\mathcal{F}}$ (see \eqref{eq:lem-discrete-2} and \eqref{eq:def-rho-star}).


\paragraph{Proof of \eqref{eq:lem-estimator-1}.}
%
Applying \eqref{eq:lem-empirical-1} in Lemma~\ref{lem:empirical} and making use of the fact that $\|\widehat{\bm \rho} - {\bm \rho}^{\star}\|_1 \le \|\widehat{\bm \rho}\|_1 + \| {\bm \rho}^{\star}\|_1\lesssim C_{\mathcal{F}} + \lambda^{-1}\varepsilon_{\mathsf{opt}}$ (see \eqref{eq:rho-hat-UB-123}), one can demonstrate that
\begin{align}
& \bigg| \mathbb{E}\big[\big({\bm \phi}_{N+1}^{\top}\widehat{\bm \rho} - f({\bm x}_{N+1})\big)^2\big] - \frac{1}{N}\sum_{i = 1}^N \big[(y_i - {\bm \phi}_i^{\top}\widehat{\bm \rho})^2 - z_i^2\big]  \bigg| \notag\\
&\qquad \lesssim \sqrt{\frac{\log N}{N}}\big(\varepsilon_{\mathsf{dis}}^2 + C_{\mathcal{F}}^2 + \lambda^{-2}\varepsilon_{\mathsf{opt}}^2 + \sigma(\varepsilon_{\mathsf{dis}} + C_{\mathcal{F}} + \lambda^{-1}\varepsilon_{\mathsf{opt}})\big).
\label{eq:proof-lem-estimator-temp-1}
\end{align}
Moreover, we make the observation that
\begin{align}
\frac{1}{N}\sum_{i = 1}^N (y_i - {\bm \phi}_i^{\top}\widehat{\bm \rho})^2 &\le \ell(\widehat{\bm \rho})\le \ell({\bm \rho}^{\star}) + \varepsilon_{\mathsf{opt}}\le  \frac{1}{N}\sum_{i = 1}^N (y_i - {\bm \phi}_i^{\top}{\bm \rho}^{\star})^2 + \lambda \|{\bm \rho}^{\star}\|_1 + \varepsilon_{\mathsf{opt}} \notag\\
&= \frac{1}{N}\sum_{i = 1}^N z_i^2 + \frac1N\sum_{i=1}^N\big(f({\bm x}_i)-{\bm \phi}_i^{\top}{\bm \rho}^{\star}\big)^2 + \frac2N\sum_{i=1}^Nz_i\big(f({\bm x}_i)-{\bm \phi}_i^{\top}{\bm \rho}^{\star}\big) + \lambda\|{\bm \rho}^{\star}\|_1 + \varepsilon_{\mathsf{opt}}\notag\\
&\le \frac{1}{N}\sum_{i = 1}^N z_i^2  + O\Big(\varepsilon_{\mathsf{dis}}^2 + \sqrt{\frac{\log N}{N}}\sigma\varepsilon_{\mathsf{dis}} + \lambda C_{\mathcal{F}}\Big)
+ \varepsilon_{\mathsf{opt}},
\label{eq:proof-lem-estimator-temp-2}
\end{align}
where the last inequality follows from the properties $(f({\bm x}_i)-\bm{\phi}_i^{\top}{\bm \rho}^{\star})^2\lesssim \varepsilon_{\mathsf{dis}}^2$ and $\|{\bm \rho}^{\star}\|_1\lesssim C_{\mathcal{F}}$ (see \eqref{eq:f-approx-epsilon-dis} and \eqref{eq:lem-discrete-2}), as well as the concentration bound for $\frac1N\sum_{i=1}^Nz_i\big(f({\bm x}_i)-{\bm \phi}_i^{\top}{\bm \rho}^{\star}\big)$ to be derived in \eqref{eq:proof-lem-empirical-boundE3}. 
Combine \eqref{eq:proof-lem-estimator-temp-1} and \eqref{eq:proof-lem-estimator-temp-2} to derive
\begin{align*}
\mathbb{E}\big[\big({\bm \phi}_{N+1}^{\top}\widehat{\bm \rho}-f({\bm x}_{N+1})\big)^2]
&\lesssim \sqrt{\frac{\log N}{N}}\big(\varepsilon_{\mathsf{dis}}^2 + C_{\mathcal{F}}^2 + \lambda^{-2}\varepsilon_{\mathsf{opt}}^2 + \sigma(\varepsilon_{\mathsf{dis}} + C_{\mathcal{F}} + \lambda^{-1}\varepsilon_{\mathsf{opt}})\big) + \varepsilon_{\mathsf{dis}}^2 + \lambda C_{\mathcal{F}} + \varepsilon_{\mathsf{opt}}\notag\\
&\asymp \sqrt{\frac{\log N}{N}}\big(C_{\mathcal{F}}^2 + \lambda^{-2}\varepsilon_{\mathsf{opt}}^2 + \sigma\varepsilon_{\mathsf{dis}}\big) + \varepsilon_{\mathsf{dis}}^2 + \lambda C_{\mathcal{F}} + \varepsilon_{\mathsf{opt}},
\end{align*}
where the last line follows since  $\sqrt{\frac{\log N}{N}}\sigma C_{\mathcal{F}}\lesssim \lambda C_{\mathcal{F}}$ and $\sqrt{\frac{\log N}{N}}\sigma\lambda^{-1}\varepsilon_{\mathsf{opt}}\lesssim \varepsilon_{\mathsf{opt}}$ (see \eqref{eq:lambda-LB-1357}).

\subsubsection{Proof of Lemma~\ref{lem:empirical}}
\label{sec:proof:lem:empirical}

We begin by establishing  \eqref{eq:lem-empirical-1}.
Given that $y_i=f(\bm{x}_i)+z_i$, 
we make note of the following decomposition: 
\begin{align}
& \bigg|\frac{1}{N}\sum_{i = 1}^N \big[(y_i - {\bm \phi}_i^{\top}{\bm \rho})^2 - z_i^2\big] - \mathbb{E}\big[\big(f({\bm x}_{N+1}) - {\bm \phi}_{N+1}^{\top}{\bm \rho}\big)^2\big]\bigg| \notag\\
&~~ = \bigg|\frac{1}{N}\sum_{i = 1}^N \Big\{\big(f({\bm x}_i) - {\bm \phi}_i^{\top}{\bm \rho}\big)^2 - \mathbb{E}\big[\big(f({\bm x}_{N+1}) - {\bm \phi}_{N+1}^{\top}{\bm \rho}\big)^2\big] + 2z_i\big(f({\bm x}_i) - {\bm \phi}_i^{\top}{\bm \rho}\big)\Big\}\bigg|\notag\\
&~~ \le \bigg|\frac{1}{N}\sum_{i = 1}^N \Big\{\big(f({\bm x}_i) - {\bm \phi}_i^{\top}{\bm \rho}\big)^2 - \big(f({\bm x}_i) - {\bm \phi}_i^{\top}{\bm \rho}^{\star}\big)^2\Big\}+\mathbb{E}\big[\big(f({\bm x}_{N+1}) - {\bm \phi}_{N+1}^{\top}{\bm \rho}^{\star}\big)^2\big]-\mathbb{E}\big[\big(f({\bm x}_{N+1}) - {\bm \phi}_{N+1}^{\top}{\bm \rho}\big)^2\big]\bigg|\notag\\
&~~\quad + \bigg|\frac{1}{N}\sum_{i = 1}^N \big(f({\bm x}_i) - {\bm \phi}_i^{\top}{\bm \rho}^{\star}\big)^2-
\mathbb{E}\big[\big(f({\bm x}_{N+1}) - {\bm \phi}_{N+1}^{\top}{\bm \rho}^{\star}\big)^2\big]\bigg|
+ 2\bigg|\sum_{i=1}^Nz_i\big(f({\bm x}_i) - {\bm \phi}_i^{\top}{\bm \rho}\big)\bigg|\eqqcolon \mathcal{E}_1 + \mathcal{E}_2 + \mathcal{E}_3,
\label{eq:proof-lem-empirical-1-decomp}
\end{align}
%
leaving us with three terms to cope with. 

\begin{itemize}
\item 
Regarding the term $\mathcal{E}_1$, we can apply a little algebra to derive
\begin{align}\label{eq:proof-lem-empirical-boundE1-pre}
\mathcal{E}_1 
&\leq \bigg|\frac2N\sum_{i=1}^N ({\bm \rho}^{\star} - {\bm \rho})^{\top}{\bm \phi}_i\big(f({\bm x}_i)-{\bm \phi}_i^{\top}{\bm \rho}^{\star}\big)-2\mathbb{E}\big[({\bm \rho}^{\star} - {\bm \rho})^{\top}{\bm \phi}_{N+1}\big(f({\bm x}_{N+1})-{\bm \phi}_{N+1}^{\top}{\bm \rho}^{\star}\big)\big]\bigg|\notag\\
&\quad + \bigg|\frac1N\sum_{i=1}^N \big(({\bm \rho}^{\star} - {\bm \rho})^{\top}{\bm \phi}_i\big)^2-\mathbb{E}\big[\big(({\bm \rho}^{\star} - {\bm \rho})^{\top}{\bm \phi}_{N+1}\big)^2\big]\bigg|\notag\\
&\le 2\left\|{\bm \rho}^{\star} - {\bm \rho}\right\|_1\bigg\|\frac1N\sum_{i=1}^N {\bm \phi}_i\big(f({\bm x}_i)-{\bm \phi}_i^{\top}{\bm \rho}^{\star}\big)-\mathbb{E}\big[{\bm \phi}_{N+1}\big(f({\bm x}_{N+1})-{\bm \phi}_{N+1}^{\top}{\bm \rho}^{\star}\big)\big]\bigg\|_{\infty}\notag\\
&\quad + \left\|{\bm \rho}^{\star} - {\bm \rho}\right\|_1^2\bigg\|\frac1N\sum_{i=1}^N {\bm \phi}_i{\bm \phi}_i^{\top}-\mathbb{E}\big[{\bm \phi}_{N+1}{\bm \phi}_{N+1}^{\top}\big]\bigg\|_{\infty},
\end{align}
where the first line arises from the elementary inequality $(a-b)^{2}-(a-c)^{2}=-(c-b)^{2}+2(c-b)(a-b)$ in conjunction with the triangle inequality. Here, for any matrix $\bm{A}$ we denote by  $\|\bm{A}\|_{\infty}=\max_{i,j}|A_{i,j}|$ the entrywise $\ell_{\infty}$ norm. 
%
Recognizing that $\|{\bm \phi}_i\|_{\infty}\le 1$ (see \eqref{eq:defn-phi-i-features} and \eqref{eq:defn-phi-feature-lemma}, as well as  $\phi(z)\in [-1,1]$ as in \eqref{eq:def-func-phi}), we obtain
\begin{align*}
\big\|{\bm \phi}_i \big(f({\bm x}_i)-{\bm \phi}_i^{\top}{\bm \rho}^{\star} \big)\big\|_{\infty}
&\le\|\bm{\phi}_i\|_{\infty} \sup_{{\bm x}\in \mathcal{B}} \big|f({\bm x})-\widetilde{\bm \phi}({\bm x})^{\top}{\bm \rho}^{\star} \big|\overset{\text{(a)}}{\lesssim} \varepsilon_{\mathsf{dis}},\\
\big\|{\bm \phi}_i{\bm \phi}_i^{\top}\big\|_{\infty}
&\le \|{\bm \phi}_i\|_{\infty}^2 \le 1
\end{align*}
with $\widetilde{\bm \phi}({\bm x}) = [\phi_1^{\mathsf{feature}}({\bm x}),\cdots, \phi_n^{\mathsf{feature}}({\bm x}),1]^{\top}$ and $\varepsilon_{\mathsf{dis}}$ defined in \eqref{eq:eps-dis}. Here,  (a) follows since 
\begin{align}\label{eq:proof-lem-empirical-temp-1}
    \sup_{{\bm x}\in \mathcal{B}} \big|f({\bm x})-\widetilde{\bm \phi}({\bm x})^{\top}{\bm \rho}^{\star}\big| \le \varepsilon_{\mathsf{dis}} + \big|f(\bm{0})-\rho^{\star}_{f,0} \big| = \varepsilon_{\mathsf{dis}},
\end{align}
which holds due to \eqref{eq:f-approx-epsilon-dis} as well as our choice $\rho^{\star}_{f,0}=f(\bm{0})$ (see \eqref{eq:def-rho-star}). 
Then applying Hoeffding's inequality shows that, with probability at least $1-O(N^{-12})$,
\begin{align*}
\bigg\|\frac1N\sum_{i=1}^N {\bm \phi}_i\big(f({\bm x}_i)-{\bm \phi}_i^{\top}{\bm \rho}^{\star}\big)-\mathbb{E}\big[{\bm \phi}_{N+1}\big(f({\bm x}_{N+1})-{\bm \phi}_{N+1}^{\top}{\bm \rho}^{\star}\big)\big]\bigg\|_{\infty}
&\lesssim \sqrt{\frac{\log N}{N}}\varepsilon_{\mathsf{dis}}
,\\
\bigg\|\frac1N\sum_{i=1}^N {\bm \phi}_i{\bm \phi}_i^{\top}-\mathbb{E}\big[{\bm \phi}_{N+1}{\bm \phi}_{N+1}^{\top}\big]\bigg\|_{\infty}
&\lesssim \sqrt{\frac{\log N}{N}}.
\end{align*}
Substitution into \eqref{eq:proof-lem-empirical-boundE1-pre} yields
\begin{align}\label{eq:proof-lem-empirical-boundE1}
\mathcal{E}_1\lesssim \sqrt{\frac{\log N}{N}}\|{\bm \rho}-{\bm \rho}^{\star}\|_1\big(\|{\bm \rho}-{\bm \rho}^{\star}\|_1 + \varepsilon_{\mathsf{dis}}\big)
\end{align}
with probability exceeding $1-O(N^{-12})$. 

\item 
With regards to $\mathcal{E}_2$,  taking the Hoeffding inequality in conjunction with \eqref{eq:proof-lem-empirical-temp-1} tell us that, with probability exceeding $1-O(N^{-12})$, 
\begin{align}\label{eq:proof-lem-empirical-boundE2}
\mathcal{E}_2\lesssim \sqrt{\frac{\log N}{N}}\varepsilon_{\mathsf{dis}}^2.
\end{align}

\item 
We now turn to $\mathcal{E}_3$. Recalling that $\|{\bm \phi}_i\|_{\infty}\le 1$ 
and making use of \eqref{eq:proof-lem-empirical-temp-1}, we apply the triangle inequality to obtain
\begin{align}\label{eq:proof-lem-empirical-boundE3}
\mathcal{E}_3 
&\le 2\bigg|\frac{1}{N}\sum_{i = 1}^N z_i\big(f({\bm x}_i) - {\bm \phi}_i^{\top}{\bm \rho}^{\star}\big)\bigg| + 2\bigg|\frac{1}{N}\sum_{i = 1}^N z_i{\bm \phi}_i^{\top}({\bm \rho}-{\bm \rho}^{\star})\bigg| \notag\\
&\le 2\bigg|\frac{1}{N}\sum_{i = 1}^N z_i\big(f({\bm x}_i) - {\bm \phi}_i^{\top}{\bm \rho}^{\star}\big)\bigg| + 2\bigg\|\frac{1}{N}\sum_{i = 1}^N z_i{\bm \phi}_i\bigg\|_{\infty}\|{\bm \rho}-{\bm \rho}^{\star}\|_1\notag\\
&\lesssim \sqrt{\frac{\log N}{N}}\sigma\big(\varepsilon_{\mathsf{dis}} + \|{\bm \rho}-{\bm \rho}^{\star}\|_1\big),
\end{align}
where we have used the sub-Gaussian assumption on $\{z_i\}$. 

\end{itemize}
Substituting \eqref{eq:proof-lem-empirical-boundE1}, \eqref{eq:proof-lem-empirical-boundE2} and \eqref{eq:proof-lem-empirical-boundE3} into \eqref{eq:proof-lem-empirical-1-decomp} reveals that, with probability at least $1-O(N^{-12})$, 
\begin{align*}
& \bigg|\frac{1}{N}\sum_{i = 1}^N \big[(y_i - {\bm \phi}_i^{\top}{\bm \rho})^2 - z_i^2\big] - \mathbb{E}\big[\big(f({\bm x}_{N+1}) - {\bm \phi}_{N+1}^{\top}{\bm \rho}\big)^2\big]\bigg|\notag\\
&\quad \lesssim \sqrt{\frac{\log N}{N}}\left(\|{\bm \rho}-{\bm \rho}^{\star}\|_1^2 + \|{\bm \rho}-{\bm \rho}^{\star}\|_1\varepsilon_{\mathsf{dis}} + \varepsilon_{\mathsf{dis}}^2 + \sigma\|{\bm \rho}-{\bm \rho}^{\star}\|_1 + \sigma\varepsilon_{\mathsf{dis}}\right)\notag\\
&\quad \asymp \sqrt{\frac{\log N}{N}}\left(\|{\bm \rho}-{\bm \rho}^{\star}\|_1^2 + \varepsilon_{\mathsf{dis}}^2 + \sigma\|{\bm \rho}-{\bm \rho}^{\star}\|_1 + \sigma\varepsilon_{\mathsf{dis}}\right) .
\end{align*}

Next, we turn to proving  \eqref{eq:lem-empirical-2}.
In view of the Hoeffding inequality, with probability at least $1-O(N^{-12})$ we have
\begin{align*}
\frac{1}{N}\sum_{i = 1}^N \big[(y_i - {\bm \phi}_i^{\top}{\bm \rho})^2 - (y_i - {\bm \phi}_i^{\top}{\bm \rho}^{\star})^2\big] 
&=\frac{1}{N}\sum_{i = 1}^N \big[\big(f({\bm x}_i) - {\bm \phi}_i^{\top}{\bm \rho}\big)^2 - \big(f({\bm x}_i) - {\bm \phi}_i^{\top}{\bm \rho}^{\star}\big)^2\big] + \frac{2}{N}\sum_{i = 1}^N z_i{\bm \phi}_i^{\top}({\bm \rho}^{\star}- {\bm \rho})\notag\\
&\ge- \frac1N\sum_{i=1}^N\big(f({\bm x}_i) - {\bm \phi}_i^{\top}{\bm \rho}^{\star}\big)^2
- \bigg\|\frac{2}{N}\sum_{i = 1}^N z_i{\bm \phi}_i\bigg\|_{\infty}\|{\bm \rho}^{\star}- {\bm \rho}\|_1\notag\\
&\gtrsim -\varepsilon_{\mathsf{dis}}^2 -\sqrt{\frac{\log N}{N}}\sigma\|{\bm \rho}^{\star}-{\bm \rho}\|_1. 
\end{align*}
where we have used the following facts (already proven previously): 
\begin{align*}
|f({\bm x}_i) - {\bm \phi}_i^{\top}{\bm \rho}^{\star}|\lesssim \varepsilon_{\mathsf{dis}},\qquad
\bigg\|\frac1N\sum_{i = 1}^N z_i{\bm \phi}_i\bigg\|_{\infty}\lesssim \sqrt{\frac{\log N}{N}}\sigma.
\end{align*}

\subsection{Proof of Lemma~\ref{lem:convergence}}
\label{subsec:proof-lem-convergence}

To begin with, we find it convenient to introduce an auxiliary sequence  ${\bm \rho}_{t+1}^{\star}$ obeying ${\bm \rho}_0^{\star} = {\bm 0}$ and 
\begin{align*}
{\bm \rho}_{t+1}^{\star} = \mathsf{ST}_{\eta\lambda}\left({\bm \rho}_t^{\mathsf{proximal}} + \frac{2\eta}{N}\sum_{i=1}^N\big(y_i-{\bm \phi}_i^{\top}{\bm \rho}_t^{\mathsf{proximal}}\big){\bm \phi}_i\right),
\end{align*}
where $\bm{\rho}_{t+1}^{\star}$ is obtained by running one {\em exact} proximal gradient iteration from $\bm{\rho}_t^{\mathsf{proximal}}$. 
Standard convergence analysis for the proximal gradient method (e.g., \citet{beck2017first}) reveals that
\begin{subequations}
\label{eq:proof-lem-convergence-PGD-12}
\begin{align}
\ell({\bm \rho}_{t+1}^{\star}) &\le \ell({\bm \rho}_{t}^{\mathsf{proximal}}),\label{eq:proof-lem-convergence-PGD-1}\\
\ell({\bm \rho}_{t+1}^{\star}) - \ell({\bm \rho}^{\star}) &\le n\big(\|{\bm \rho}_t^{\mathsf{proximal}} - {\bm \rho}^{\star}\|_2^2 - \|{\bm \rho}_{t+1}^{\star} - {\bm \rho}^{\star}\|_2^2\big),\label{eq:proof-lem-convergence-PGD-2}
\end{align}
\end{subequations}
where we recall our choice that $\eta = 1/(2n)$. 
For completeness, we shall provide the proof of \eqref{eq:proof-lem-convergence-PGD-12} towards the end of this subsection.

Recognizing that ${\bm \rho}_{t}^{\mathsf{proximal}} = {\bm \rho}_t^{\star} + {\bm e}_t$ for some additive term $\bm{e}_t$, we can invoke \eqref{eq:proof-lem-convergence-PGD-2} to show that
\begin{align}\label{eq:proof-lem-convergence-optdiff}
\ell({\bm \rho}_{t+1}^{\star}) - \ell({\bm \rho}^{\star}) 
&\le n\big(\|{\bm \rho}_t^{\mathsf{proximal}} - {\bm \rho}^{\star}\|_2^2 - \|{\bm \rho}_{t+1}^{\star} - {\bm \rho}^{\star}\|_2^2 \big) \notag\\
&= n\big(\|{\bm \rho}_t^{\star} - {\bm \rho}^{\star}\|_2^2 - \|{\bm \rho}_{t+1}^{\star} - {\bm \rho}^{\star}\|_2^2 + \|{\bm e}_t\|_2^2 
+ 2\bm{e}_t^{\top} (\bm{\rho}_t^{\star} - \bm{\rho}^{\star}) \big) \notag\\
&\le n\big(\|{\bm \rho}_t^{\star} - {\bm \rho}^{\star}\|_2^2 - \|{\bm \rho}_{t+1}^{\star} - {\bm \rho}^{\star}\|_2^2 + \|{\bm e}_t\|_1^2 + 2\|{\bm e}_t\|_1(O(C_{\mathcal{F}})+\|{\bm \rho}_t^{\star}\|_1)\big)\notag\\
&\le n\big(\|{\bm \rho}_t^{\star} - {\bm \rho}^{\star}\|_2^2 - \|{\bm \rho}_{t+1}^{\star} - {\bm \rho}^{\star}\|_2^2 +  \|{\bm e}_t\|_1^2 +2\|{\bm e}_t\|_1(O(C_{\mathcal{F}})+\|{\bm \rho}_t^{\mathsf{proximal}}\|_1 + \|{\bm e}_t\|_1)\big)\notag\\
&\le n\big(\|{\bm \rho}_t^{\star} - {\bm \rho}^{\star}\|_2^2 - \|{\bm \rho}_{t+1}^{\star} - {\bm \rho}^{\star}\|_2^2   +c_1\|{\bm e}_t\|_1(C_{\mathcal{F}}+\|{\bm \rho}_t^{\mathsf{proximal}}\|_1)\big)
\end{align}
for some universal constant $c_1>0$, where we have used  $\|{\bm \rho}^{\star}\|_1\lesssim C_{\mathcal{F}}$ (cf.~\eqref{eq:lem-discrete-2} and \eqref{eq:def-rho-star}) and the assumption $\|{\bm e}_t\|_1\lesssim C_{\mathcal{F}}$.
Define
$$
k_t = \arg\min_{1\le k\le t} \ell({\bm \rho}_k^{\star}).
$$
Summing \eqref{eq:proof-lem-convergence-optdiff} over iterations $0$ to $t$, we obtain a telescoping sum and can then deduce that
\begin{align}\label{eq:proof-lem-convergence-lrhostar}
\ell({\bm \rho}_{k_t}^{\star}) - \ell({\bm \rho}^{\star})  &= \min_{1\le k\le t}\ell({\bm \rho}_k^{\star}) - \ell({\bm \rho}^{\star})\notag\\ 
&\le 
\frac{1}{t}\sum_{k=1}^t \big( \ell({\bm \rho}_k^{\star}) - \ell({\bm \rho}^{\star}) \big) \le \frac{n\|{\bm \rho}_0^{\mathsf{proximal}} - {\bm \rho}^{\star}\|_2^2}{t} + c_1n\max_{1\le k< t} \big\{\|{\bm e}_k\|_1(C_{\mathcal{F}}+\|{\bm \rho}_k^{\mathsf{proximal}}\|_1) \big\}\notag\\
&{=} \frac{n\|{\bm \rho}^{\star}\|_1^2}{t} + c_1n\max_{1\le k< t}\big\{\|{\bm e}_k\|_1(C_{\mathcal{F}}+\|{\bm \rho}_k^{\mathsf{proximal}}\|_1)\big\},
\end{align}
where the last line follows since ${\bm \rho}_0^{\mathsf{proximal}} = {\bm 0}$.

In addition, it is seen that
\begin{align}
\ell({\bm \rho}_{t+1}^{\mathsf{proximal}}) - \ell({\bm \rho}_{t+1}^{\star}) 
&= \frac1N\sum_{i=1}^N\big(y_i - {\bm \phi}_i^{\top}{\bm \rho}_{t+1}^{\mathsf{proximal}}\big)^2 - \frac1N\sum_{i=1}^N\big(y_i - {\bm \phi}_i^{\top}{\bm \rho}_{t+1}^{\star}\big)^2 + \lambda\big\|{\bm \rho}_{t+1}^{\mathsf{proximal}}\big\|_1 - \lambda\|{\bm \rho}_{t+1}^{\star}\|_1\notag\\
&\le \frac2N\sum_{i=1}^N{\bm \phi}_i^{\top}\big({\bm \rho}_{t+1}^{\star} - {\bm \rho}_{t+1}^{\mathsf{proximal}}\big)\big(y_i - {\bm \phi}_i^{\top}{\bm \rho}_{t+1}^{\mathsf{proximal}}\big) + \lambda\big\|{\bm \rho}_{t+1}^{\mathsf{proximal}} - {\bm \rho}_{t+1}^{\star}\big\|_1\notag\\
&\le \big\|{\bm \rho}_{t+1}^{\star} - {\bm \rho}_{t+1}^{\mathsf{proximal}}\big\|_1\left(\left\|\frac2N\sum_{i=1}^N\big(y_i - {\bm \phi}_i^{\top}{\bm \rho}_{t+1}^{\mathsf{proximal}}\big){\bm \phi}_i\right\|_{\infty} + \lambda\right)\notag\\
&\overset{\text{(a)}}{\le} c_2\|{\bm e}_{t+1}\|_1 \big(C_{\mathcal{F}} + \sigma + \|{\bm \rho}_{t+1}^{\mathsf{proximal}}\|_1 + \lambda
\big) 
\label{eq:UB-l-rho-tplus1-tstar}
\end{align}
for some universal constant $c_2>0$, where the first inequality comes from the elementary inequality $(a-b)^{2}-(a-c)^{2}=-(c-b)^{2}+(c-b)(2a-2b)\leq 2(c-b)(a-b)$ as well as the triangle inequality. Here, (a) follows since
\begin{align*}
\frac1N\sum_{i=1}^N|y_i|&\le \frac1N\sum_{i=1}^N|f({\bm x}_i)| + \frac1N\sum_{i=1}^N|z_i|
\le \max_{1\le i\le N}|{\bm \phi}_i^{\top}{\bm \rho}^{\star}| + \varepsilon_{\mathsf{dis}} + \left|\frac1N\sum_{i=1}^N|z_i| - \mathbb{E}[|z|]\right| + \mathbb{E}[|z|] \notag\\
& \lesssim  \|\bm{\rho}^{\star}\|_1 \max_{1\leq i\leq N} \|\bm{\phi}_i\|_{\infty} + \varepsilon_{\mathsf{dis}} + \sigma 
\lesssim \|\bm{\rho}^{\star}\|_1 + \varepsilon_{\mathsf{dis}} 
 + C_{\mathcal{F}} + \sigma 
\asymp C_{\mathcal{F}} + \sigma ,
\end{align*}
a consequence of the sub-Gaussian assumption on $\{z_i\}$ and the facts that $\|\bm{\phi}_i\|_{\infty}\leq 1$, $\|\bm{\rho}^{\star}\|_1\lesssim C_{\mathcal{F}}$, and $\varepsilon_{\mathsf{dis}} = C_{\mathcal{F}}\left( \sqrt{\varepsilon} + (\frac{\log|N_\varepsilon|}{n})^{1/3}\right) \lesssim C_{\mathcal{F}}$ for the assumption that $\varepsilon\lesssim \sqrt{\log N/N} + n/L $ and $n\gtrsim \log|N_{\varepsilon}|$.  
Recalling that $\ell({\bm \rho}_{t+1}^{\star}) \le \ell({\bm \rho}_t^{\mathsf{proximal}})$ (see \eqref{eq:proof-lem-convergence-PGD-1}), we can invoke the bound \eqref{eq:UB-l-rho-tplus1-tstar} recursively to derive
\begin{align}
\ell({\bm \rho}_{t+1}^{\mathsf{proximal}})
&\le 
\ell({\bm \rho}_{t+1}^{\star}) + c_2\|{\bm e}_{t+1}\|_1\big(C_{\mathcal{F}} + \sigma + \|{\bm \rho}_{t+1}^{\mathsf{proximal}}\|_1 + \lambda\big)\notag\\
 &\le \ell({\bm \rho}_t^{\mathsf{proximal}}) + c_2\|{\bm e}_{t+1}\|_1\big(C_{\mathcal{F}} + \sigma + \|{\bm \rho}_{t+1}^{\mathsf{proximal}}\|_1 + \lambda\big)\label{eq:proof-lem-convergence-temp-4}\\
 &\le \ell({\bm \rho}_t^{\star}) + 
2c_2\max_{t\le k\le t+1}\big\{\|{\bm e}_{k}\|_1\big(C_{\mathcal{F}} + \sigma + \|{\bm \rho}_{k}^{\mathsf{proximal}}\|_1 + \lambda\big)\big\}\notag\\
&\le \ell({\bm \rho}_{k_t}^{\star}) + 
c_2(t+1-k_t)\max_{1\le k\le t+1}\big\{\|{\bm e}_{k}\|_1(C_{\mathcal{F}} + \sigma + \|{\bm \rho}_{k}^{\mathsf{proximal}}\|_1 + \lambda)\big\}.
\end{align}
It then follows from \eqref{eq:proof-lem-convergence-lrhostar} that
\begin{align*}
\ell({\bm \rho}_{t+1}^{\mathsf{proximal}}) - \ell({\bm \rho}^{\star})
&\le \frac{n\|{\bm \rho}^{\star}\|_1^2}{t} + c_1n\max_{1\le k\le t-1} \big\{\|{\bm e}_k\|_1(C_{\mathcal{F}}+\|{\bm \rho}_k^{\mathsf{proximal}}\|_1) \big\} \notag\\
&\quad + 
c_2(t+1)\max_{1\le k\le t+1}\big\{\|{\bm e}_{k}\|_1(C_{\mathcal{F}} + \sigma + \|{\bm \rho}_{k}^{\mathsf{proximal}}\|_1 + \lambda)\big\}\notag\\
&\le \frac{n\|{\bm \rho}^{\star}\|_1^2}{t} + c_3(t+n+1)\max_{1\le k\le t+1}\big\{\|{\bm e}_{k}\|_1(C_{\mathcal{F}} + \sigma + \|{\bm \rho}_{k}^{\mathsf{proximal}}\|_1 + \lambda)\big\},\notag
\end{align*}
where $c_3 = \max\{c_1,c_2\}$.
Recalling that $\|{\bm \rho}^{\star}\|_1\lesssim C_{\mathcal{F}}$ (cf.~\eqref{eq:lem-discrete-2} and \eqref{eq:def-rho-star}) as well as our parameter choice $T=\frac{L-1}{2}$,  we have thus completed the proof of \eqref{eq:convergence-rho}.

In addition, combining \eqref{eq:lem-estimator-2} in Lemma~\ref{lem:estimator} with the above result, we know that for $\lambda\gtrsim \sqrt{\frac{\log N}{N}}(C_{\mathcal{F}} + \sigma)$,
\begin{align*}
\|{\bm \rho}_t^{\mathsf{proximal}}\|_1
&\lesssim C_{\mathcal{F}} + \lambda^{-1}\left(\frac{n\|{\bm \rho}^{\star}\|_1^2}{t-1} + (t+n)\max_{1\le k\le t} \big\{ \|{\bm e}_{k}\|_1(C_{\mathcal{F}} + \sigma + \|{\bm \rho}_{k}^{\mathsf{proximal}}\|_1 + \lambda)\big\}\right) \notag\\
&\lesssim C_{\mathcal{F}} + \frac{n\|{\bm \rho}^{\star}\|_1^2}{t\lambda} + \sqrt{N}(t+n)\max_{1\le k\le t}\|{\bm e}_{k}\|_1 + \frac{t+n}{\lambda}\max_{1\le k\le t}\big\{\|{\bm e}_k\|_1\|{\bm \rho}_k^{\mathsf{proximal}}\|_1\big\},\quad t\ge 2.
\end{align*}
For $t=0$ and $t=1$, we have $\|{\bm \rho}_0^{\mathsf{proximal}} \|_1=0$, and with probability at least $1-O(N^{-20})$,
\begin{align*}
\|{\bm \rho}_1^{\mathsf{proximal}}\|_1
&\le \sum_{k=1}^n\max\left\{\left\|\frac{2\eta}{N}\sum_{i=1}^Ny_i{\bm \phi}_i\right\|_{\infty} +|{e}_{1,k}| - \lambda\eta ,0\right\}\notag\\
&\le \sum_{k=1}^n\max\left\{\frac{c}{n}\left(C_{\mathcal{F}} + \sqrt{\frac{\log N}{N}}\sigma\right) + |{e}_{1,k}| -\frac{\lambda}{2n}  ,0\right\}\lesssim C_{\mathcal{F}}+ \|{\bm e}_1\|_1,
\end{align*}
provided that $\lambda\ge 2c\sqrt{\frac{\log N}{N}}\sigma$. Here,  $e_{1,k}$ denotes the $k$-th element of ${\bm e}_1$.

\paragraph{Proof of \eqref{eq:proof-lem-convergence-PGD-1} and \eqref{eq:proof-lem-convergence-PGD-2}.}

Define
\begin{align*}
g({\bm \rho}) &= \frac1N \sum_{i=1}^N(y_i - {\bm \phi}_i^{\top}{\bm \rho})^2 , \\
\psi({\bm \rho}) &= g({\bm \rho}_t^{\mathsf{proximal}}) + ({\bm \rho} - {\bm \rho}_t^{\mathsf{proximal}})^{\top}\nabla g({\bm \rho}_t^{\mathsf{proximal}}) + n\|{\bm \rho} - {\bm \rho}_t^{\mathsf{proximal}}\|_2^2 + \lambda\|{\bm \rho}\|_1.
\end{align*}
Recall that $\eta = 1/(2n)$.
It is self-evident that ${\bm \rho}_{t+1}^{\star} = \arg\min_{\bm{\rho}} \psi({\bm \rho})$ and $\psi(\cdot)$ is $(2n)$-strongly convex.
Thus, for any $\widehat{\bm \rho}$, we have
\begin{align}\label{eq:proof-eq-PGD-temp-1}
\psi(\widehat{\bm \rho})\ge \psi({\bm \rho}_{t+1}^{\star}) + n\|{\bm \rho}_{t+1}^{\star} - \widehat{\bm \rho}\|_2^2.
\end{align}
In addition, observe that
\begin{align*}
& g({\bm \rho}_t^{\mathsf{proximal}}) + ({\bm \rho} - {\bm \rho}_t^{\mathsf{proximal}})^{\top}\nabla g({\bm \rho}_t^{\mathsf{proximal}}) + n\|{\bm \rho} - {\bm \rho}_t^{\mathsf{proximal}}\|_2^2 \notag\\
&\quad= g({\bm \rho}_t^{\mathsf{proximal}}) - \frac2N\sum_{i=1}^N({\bm \rho} - {\bm \rho}_t^{\mathsf{proximal}})^{\top}(y_i - {\bm \phi}_i^{\top}{\bm \rho}){\bm \phi}_i + n\|{\bm \rho} - {\bm \rho}_t^{\mathsf{proximal}}\|_2^2\notag\\
&\quad \ge g({\bm \rho}_t^{\mathsf{proximal}}) - \frac2N\sum_{i=1}^N({\bm \rho} - {\bm \rho}_t^{\mathsf{proximal}})^{\top}(y_i - {\bm \phi}_i^{\top}{\bm \rho}){\bm \phi}_i + \frac1N\sum_{i=1}^N\big({\bm \phi}_i^{\top}({\bm \rho} - {\bm \rho}_t^{\mathsf{proximal}})\big)^2=g({\bm \rho}),
\end{align*}
where the last inequality applies the fact that $\sum_{i=1}^N\|{\bm \phi}_i\|_2^2 \le Nn$.
Thus, we can conclude that
\begin{align}\label{eq:proof-eq-PGD-temp-2}
\psi({\bm \rho}_{t+1}^{\star})\ge g({\bm \rho}_{t+1}^{\star}) + \lambda \|{\bm \rho}_{t+1}^{\star}\|_1 = \ell ({\bm \rho}_{t+1}^{\star}).
\end{align}

Similarly, we can also demonstrate that
\begin{align*}
g({\bm \rho}_t^{\mathsf{proximal}}) + ({\bm \rho} - {\bm \rho}_t^{\mathsf{proximal}})^{\top}\nabla g({\bm \rho}_t^{\mathsf{proximal}}) 
&= g({\bm \rho}_t^{\mathsf{proximal}}) - \frac2N\sum_{i=1}^N({\bm \rho} - {\bm \rho}_t^{\mathsf{proximal}})^{\top}(y_i - {\bm \phi}_i^{\top}{\bm \rho}){\bm \phi}_i\notag\\
&= g({\bm \rho}) - \frac1N\sum_{i=1}^N\big({\bm \phi}_i^{\top}({\bm \rho} - {\bm \rho}_t^{\mathsf{proximal}})\big)^2 \le g({\bm \rho}),
\end{align*}
and as a consequence, 
\begin{align}\label{eq:proof-eq-PGD-temp-3}
\psi(\widehat{\bm \rho}) \le g(\widehat{\bm \rho}) + n\|\widehat{\bm \rho} - {\bm \rho}_t^{\mathsf{proximal}}\|_2^2 + \lambda\|\widehat{\bm \rho}\|_1 = \ell(\widehat{\bm \rho}) + n\|\widehat{\bm \rho} - {\bm \rho}_t^{\mathsf{proximal}}\|_2^2.
\end{align}

Substituting \eqref{eq:proof-eq-PGD-temp-2} and \eqref{eq:proof-eq-PGD-temp-3} into \eqref{eq:proof-eq-PGD-temp-1}, we see that for any $\widehat{\bm \rho}$,
\begin{align*}
\ell(\widehat{\bm \rho}) \ge \ell({\bm \rho}_{t+1}^{\star}) + n\big(\|\widehat{\bm \rho} - {\bm \rho}_{t+1}^{\star}\|_2^2 - \|\widehat{\bm \rho} - {\bm \rho}_t^{\mathsf{proximal}}\|_2^2\big).
\end{align*}
Taking $\widehat{\bm \rho} = {\bm \rho}_t^{\mathsf{proximal}}$ yields
\begin{align*}
\ell({\bm \rho}_t^{\mathsf{proximal}}) \ge \ell({\bm \rho}_{t+1}^{\star}) + n\|{\bm \rho}_t^{\mathsf{proximal}} - {\bm \rho}_{t+1}^{\star}\|_2^2 \ge \ell({\bm \rho}_{t+1}^{\star}), 
\end{align*}
which completes the proof of \eqref{eq:proof-lem-convergence-PGD-1}. In addition, taking $\widehat{\bm{\rho}}=\bm{\rho}^{\star}$ gives
\begin{align*}
\ell({\bm \rho}^{\star}) \ge \ell({\bm \rho}_{t+1}^{\star}) + n\big(\|{\bm \rho}^{\star} - {\bm \rho}_{t+1}^{\star}\|_2^2 -\|{\bm \rho}^{\star} - {\bm \rho}_t^{\mathsf{proximal}}\|_2^2\big),
\end{align*}
thus concluding the proof of \eqref{eq:proof-lem-convergence-PGD-2}.

\subsection{Proof of Lemma \ref{lem:opt}}
\label{subsec:proof-lem-opt}

In this proof, we first show how to construct the desirable transformer, followed by an analysis of this construction. 
In particular, we would like to show that the transformer as illustrated in Figure~\ref{fig:transformer}, with parameters specified below, satisfy properties i) and ii) in Lemma \ref{lem:opt}.
For ease of presentation, we denote
\begin{align}
\bm{H}^{(l)} & =\MLP_{\bm{\Theta}_{\mlp}^{(l)}}\big(\bm{H}^{(l-1/2)}\big), \quad \bm{H}^{(l-1/2)} = \Att_{\bm{\Theta}_{\att}^{(l)}}\big(\bm{H}^{(l-1)}\big),\qquad l = 1, \dots, L. 
\end{align}
Throughout this subsection, all components in ${\bm H}^{(l-1/2)}$ share the same superscript $l-1/2$ (e.g., ${\bm \phi}_j^{(l-1/2)}$, $\lambda_j^{(l-1/2)}$),  which help distinguish between different layers.


\subsubsection{High-level structure}
The overall structure of the transformer is shown in Figure~\ref{fig:transformer}.
\begin{itemize}
\item The architecture begins with an attention layer \textbf{Attn0} and a feedforward layer \textbf{FF0}, which serve to initialize certain variables (particularly the features identified in Lemma~\ref{lem:discrete}) based on the in-context inputs $\{{\bm x}_i\}$.

\item 
The remaining $(L-1)$ layers are divided into $(L-1)/2$ blocks with identical structure and parameters.
More concretely, each block consists of two attention layers and two feed-forward layers, $(\textbf{Attn1}, \textbf{FF1})$ and $(\textbf{Attn2}, \textbf{FF2})$,
which are designed to perform inexact proximal gradient iterations and update the corresponding  prediction. 
\end{itemize}

\noindent 
In the sequel, we shall first describe what update rule each layer is designed to implement, followed by detailed explanation about how they can be realized using the transformer architecture.

\subsubsection{Intended updates for each layer} 
Our transformer construction comprises multiple layers (as illustrated in Figure~\ref{fig:transformer}) designed to emulate the iterations of the inexact proximal gradient method \eqref{eq:update-rule-PGD}. 
Let us begin by describing the desired update to be performed at each layer, abstracting away the specifics of the transformer implementation.  
\begin{itemize}
\item \textbf{FF0}: This feed-forward layer intends to update the components ${\bm \phi}_j$ and $\lambda$ as follows: 
\begin{subequations}
\label{eq:update-rule-FF0-effective}
\begin{align}
{\bm \phi}_j^{(l)} &= {\bm \phi}_j^{(l-1/2)} + \big[\phi_1^{\mathsf{feature}}\big({\bm x}_j^{(l-1/2)}\big),\cdots,\phi_n^{\mathsf{feature}}\big({\bm x}_j^{(l-1/2)}\big), 1 \big]^{\top},\quad 1\le j\le N+1; \\
\lambda^{(l)} &=c_1\Big(\frac{\log N}{N}\Big)^{1/6} C_{\mathcal{F}}^{-1/3}\widehat{\varepsilon}^{2/3} + c_1\sqrt{\frac{\log N}{N}}\Big(C_{\mathcal{F}} + \sigma\Big) + c_1C_{\mathcal{F}}^{-1}\varepsilon_{\mathsf{dis}}^2\eqqcolon \overline{\lambda},
\label{eq:defn-lambda-bar-FF0}
\end{align}
\end{subequations}
where $\phi_i^{\mathsf{feature}}({\bm x})$ is defined in \eqref{eq:def-func-phi}, $\widehat{\varepsilon}$ is defined in Lemma \ref{lem:opt} (see \eqref{eq:def-hat-eps}), and $c_1>0$ is some large enough universal constant.

\item \textbf{Attn1}: In this attention layer, we attempt to implement the following updates: 
\begin{align}
{\bm \rho}^{(l-1/2)} 
&={\bm \rho}^{(l-1)}  + \frac{2\eta}{N}\sum_{i = 1}^{N} {\bm \phi}_i^{(l-1)}\Big\{y_i^{(l-1)} - \big({\bm \phi}_i^{(l-1)}\big)^{\top}{\bm \rho}^{(l-1)}\Big\} + {\bm e}^{(l-1)}
\label{eq:Attn1-true-update-e}
\end{align}
for some residual (or error) term $\bm{e}^{(l-1)}$, corresponding to an iteration of gradient descent in \eqref{eq:update-rule-PGD-prox} before the proximal operator is applied. 
This residual term $\bm{e}^{(l-1)}$ shall be bounded shortly. 

\item \textbf{FF1}: This feed-forward layer is designed to implement the following updates: 
\begin{subequations}
\label{eq:equiv-update-FF1}
\begin{align}
{\bm \rho}^{(l)} &=\mathsf{ST}_{\eta\lambda^{(l-1/2)}}\big({\bm \rho}^{(l-1/2)}\big), \\
\widehat{y}^{(l)} &=0,
\end{align}
\end{subequations}
which applies the proximal operator (i.e., soft-thresholding) to the output in \eqref{eq:Attn1-true-update-e}. As a result, this in conjunction with \eqref{eq:Attn1-true-update-e} in $\bf{Attn1}$ completes one (inexact) proximal gradient iteration \eqref{eq:update-rule-PGD-prox}.

\item \textbf{Attn2}:
This attention layer intends to update the prediction $\widehat{y}$ based on $\bm{\rho}^{(l)}$, namely,
\begin{align}\label{eq:update-haty}
\widehat{y}^{(l+1/2)} 
&=  \big({\bm \phi}_{N+1}^{(l)}\big)^{\top}{\bm \rho}^{(l)} + \widetilde{e}^{(l)},
\end{align}
where  $\widetilde{e}^{(l)}$ is some residual term that will be bounded momentarily. 

\item \textbf{Attn0, FF2}:
These layers do not update the hidden representation ${\bm H}^{(l)}$; instead, they are included to ensure consistency with the transformer architecture defined in \eqref{eq:transformer-structure-defn}. 
\end{itemize}

On a high level, this transformer implements the inexact proximal gradient method in \eqref{eq:update-rule-PGD}.
In particular, after passing the input through \textbf{Attn0} and \textbf{FF0}, we obtain
\begin{align*}
{\bm \phi}_j^{(1)} = {\bm \phi}_j,\quad \lambda^{(1)} = \overline{\lambda},\quad 1\le j\le N+1.
\end{align*}
The remaining layers then proceed as follows: 
\begin{itemize}
    \item All parameters except ${\bm \rho}^{(l)}$ and $\widehat{y}^{(l)}$ will stay fixed throughout the remaining layers, i.e.,  
\begin{align}
{\bm \phi}_j^{(l)} = {\bm \phi}_j,\quad \lambda^{(l)} = \lambda^{(l+1/2)} = \overline{\lambda},\quad y_j^{(l)} = y_j^{(0)} = y_j,\quad w_j^{(l)} = w_j^{(0)},\quad \forall 1\le l\le L,~  1\le j\le N+1.
\end{align}
where $w_j^{(0)} = 1$ for $1\le j\le N$ and $w_{N+1}^{(0)} = 0$, as previously defined in \eqref{eq:edf-initial-yw} and \eqref{eq:edf-initial-wrholamyhat}.
%

\item The components ${\bm \rho}^{(l)}$ are updated in a way that resembles \eqref{eq:update-rule-PGD}, namely, 
\begin{align}\label{eq:rela-rhol-rhot}
{\bm \rho}^{(2t+1)} = {\bm \rho}_t^{\mathsf{proximal}}\quad \text{for }t\ge 1  \qquad \text{and}\qquad {\bm \rho}^{(0)} = {\bm \rho}_0^{\mathsf{proximal}} = {\bm 0},\quad \lambda = \overline{\lambda}.
\end{align}

\item The components $\widehat{y}^{(l)}$ are computed to approximate the prediction of $f({\bm x}_{N+1})$, namely, 
$$
\widehat{y}^{(2t)} = 0,\quad \widehat{y}^{(2t+1)}\approx {\bm \phi}_{N+1}^{\top}{\bm \rho}_t^{\mathsf{proximal}}\quad \text{for }t\ge 1.
$$
\end{itemize}


\subsubsection{Parameter design in our transformer construction}
Next, we explain how the transformer architecture can be designed to implement the updates described above for each layer. 


\begin{itemize}
\item \textbf{FF0}:
Note that the function $\phi(z)$ (cf.~\eqref{eq:def-func-phi}) is intimately connected with the ReLU function $\sigma_{\mathsf{ff}}(\cdot)$ as:
\begin{align}
\phi(z) = \sigma_{\mathsf{ff}}(z+1/2) - \sigma_{\mathsf{ff}}(z-1/2).
\end{align}
This allows us to decompose \eqref{eq:def-func-phi} as
\begin{align}
\phi_i^{\mathsf{feature}}({\bm x}) 
&= \sigma_{\mathsf{ff}}\left(\tau_{\mathsf{ff}}\left(\frac{{\bm \omega}_i^{\top}{\bm x}}{\|{\bm \omega}_i\|_2} - t_i\right)+\frac12\right) - \sigma_{\mathsf{ff}}\left(\tau_{\mathsf{ff}}\left(\frac{{\bm \omega}_i^{\top}{\bm x}}{\|{\bm \omega}_i\|_2} - t_i\right)-\frac12\right)  ,\quad 1\le i\le n,\label{eq:decomp-phi-feature}
\end{align}
where we take
\begin{align}
\tau_{\mathsf{ff}} = 1/\sqrt{\varepsilon}.
\end{align}
Moreover, the last entry in ${\bm \phi}_j^{(l)}$ can be expressed by $1 = \sigma_{\mathsf{ff}}(1) - \sigma_{\mathsf{ff}}(-1)$. 
In order for this layer to carry out \eqref{eq:update-rule-FF0-effective}, we take the parameter matrices ${\bm W}^{(l)}\in\mathbb{R}^{D\times D}$ and ${\bm U}^{(l)}\in\mathbb{R}^{D\times D}$ (see \eqref{eq:defn-FF-MLP-layer}) to satisfy
\begin{subequations}
\begin{align}
{\bm W}_{1:2n+2,\,1:d+1}^{(l)} &= \left[
\begin{array}{cc}
\|{\bm \omega}_1\|_2^{-1}\tau_\mathsf{ff}{\bm \omega}_1^{\top} & -t_1\tau_\mathsf{ff} + \frac12\\
\vdots & \vdots \\
\|{\bm \omega}_n\|_2^{-1}\tau_\mathsf{ff}{\bm \omega}_n^{\top} & -t_n\tau_\mathsf{ff} + \frac12\\
{\bm 0}^{\top} & 1\\
\|{\bm \omega}_1\|_2^{-1}\tau_\mathsf{ff}{\bm \omega}_1^{\top} & -t_1\tau_\mathsf{ff} - \frac12\\
\vdots & \vdots \\
\|{\bm \omega}_n\|_2^{-1}\tau_\mathsf{ff}{\bm \omega}_n^{\top} & -t_n\tau_\mathsf{ff} - \frac12\\
{\bm 0}^{\top} & -1
\end{array}
\right], \qquad {\bm W}_{2n+3,:}^{(l)} = \overline{\lambda} {\bm u}_{d+1}^{\top},\\
{\bm U}^{(l)}_{d+4:d+n+4,\,1:2n+2} &= [{\bm I}_{n+1}, -{\bm I}_{n+1}],\qquad {\bm U}_{d+2n+6,:}^{(l)} = {\bm u}_{2n+3}^{\top},
\end{align}
\end{subequations}
with all remaining entries set to zero. Here, $\overline{\lambda}$ has been defined in \eqref{eq:defn-lambda-bar-FF0}, and 
${\bm W}_{i:j,\,r:h}$ denotes a submatrix of ${\bm W}$ consisting of rows $i$ through $j$ and columns $r$ through $h$,  
 $\bm{W}_{i,:}$ represents the $i$-th row of $\bm{W}$, 
 whereas ${\bm u}_i\in\mathbb{R}^{D}$ stands for the $i$-th standard basis vector.

\item \textbf{Attn1}: 
First, we discuss how the logistic function $\sigma_{\mathsf{attn}}(\cdot)$  defined in 
\eqref{eq:sigma-attn-ff-choice} can help us implement \eqref{eq:Attn1-true-update-e}.  Note that  $\sigma_{\mathsf{attn}}'(0) = 1/4$ and observe the Taylor expansion
$
\sigma_{\mathsf{attn}}(\tau^{-1}x) = \sigma_{\mathsf{attn}}(0) + \frac{x}{4
\tau} + O\big(\frac{x^2}{\tau^2}\big), 
$
which allow us to express
\begin{align}\label{eq:approx-sigma-attn}
x\approx 4\tau\big(\sigma_{\mathsf{attn}}(\tau^{-1}x) - \sigma_{\mathsf{attn}}(0)\big).
\end{align}
In light of this observation, we propose to carry out~\eqref{eq:Attn1-true-update-e} via the following updates that exploit the logistic function: 
\begin{align}
{\bm \rho}^{(l-1/2)} &={\bm \rho}^{(l-1)} + \frac{2}{N}\sum_{i=1}^{N+1}4\tau\Big(\sigma_{\mathsf{attn}}\big(\tau^{-1}\eta y_i^{(l-1)}\big) - \sigma_{\mathsf{attn}}(0)\Big){\bm \phi}_i^{(l-1)} \notag\\
&\quad - \frac{2}{N}\sum_{i=1}^{N+1}4\tau\left(\sigma_{\mathsf{attn}}\left(\tau^{-1}\eta \big({\bm \phi}_i^{(l-1)}\big)^{\top}{\bm \rho}^{(l-1)}\right) - \sigma_{\mathsf{attn}}(0)\right){\bm \phi}_i^{(l-1)} \notag\\
&\quad + \frac{2}{N}\sum_{i=1}^{N+1}4\tau\left(\sigma_{\mathsf{attn}}\big(\tau^{-1}\eta (1-{w}_i^{(l-1)})\widehat{y}^{(l-1)}\big) - \sigma_{\mathsf{attn}}(0)\right){\bm \phi}_i^{(l-1)},
\label{eq:Attn1-true-update-no-e}
\end{align}
where the error vector $\bm{e}^{(l-1)}$ can be straightforwardly determined by checking the difference between \eqref{eq:Attn1-true-update-e} and \eqref{eq:Attn1-true-update-no-e}. 
Next, to fully realize \eqref{eq:Attn1-true-update-no-e} via a attention layer, we use 4 attention heads in this layer and take the parameter matrices ${\bm V}_m^{(l)}$, ${\bm Q}_m^{(l)}$, and ${\bm K}_m^{(l)}\in\mathbb{R}^{D\times D}$ for $m = 1,2,3,4$ to be:
\begin{table}[H]
\begin{tabular}{ l l l }
${\bm V}_{1,(d+n+5:d+2n+5,\,d+4:d+n+4)}^{(l)} = 8\tau {\bm I}_{n+1}$ & ${\bm Q}_{1,(1,:)}^{(l)} = \tau^{-1}\eta {\bm u}_{d+2}^{\top}$ & ${\bm K}_{1,(1,:)}^{(l)} ={\bm u}_{d+1}^{\top}$\\
${\bm V}_2^{(l)} = -{\bm V}_1^{(l)}$ & $ {\bm Q}_{2,(1:n+1,\,d+4:d+n+4)}^{(l)} = \tau^{-1}\eta {\bm I}_{n+1}$ & $ {\bm K}_{2,(1:n+1,\,d+n+5:d+2n+5)}^{(l)} = {\bm I}_{n+1}$\\
${\bm V}_3^{(l)} = {\bm V}_1^{(l)}$ & $ {\bm Q}_{3,(1,:)}^{(l)} = \tau^{-1}\eta({\bm u}_{d+1} - {\bm u}_{d+3})^{\top}$ & ${\bm K}_{3,(1,:)} = {\bm u}_{d+2n+7}^{\top}$
\end{tabular}
\end{table}

with all remaining entries set to zero.
Here, ${\bm V}_{m,(i:j,\,r:h)}^{(l)}$ denotes a submatrix of ${\bm V}_m^{(l)}$ comprising rows $i$ through $j$ and columns  $r$ through $h$,  ${\bm V}_{m,(i,\,:)}^{(l)}$ denotes the $i$-th row of ${\bm V}_m^{(l)}$, and ${\bm u}_i\in\mathbb{R}^{D}$ represents the $i$-th standard basis vector.

Before proceeding, let us bound the residual vector ${\bm e}^{(l-1)}$ in \eqref{eq:Attn1-true-update-e}. 
Observing that $\sigma_{\mathsf{attn}}'(0) = 1/4$ and $|\sigma_{\mathsf{attn}}''(x)|=|{\rm e}^x - {\rm e}^{-x}|/({\rm e}^{x/2} + {\rm e}^{-x/2})^4 < 0.5$ for all $x$, we can show that
\begin{align}\label{eq:approx-sigma-att}
\Big|x - 4\tau\big[\sigma_{\mathsf{attn}}(\tau^{-1} x) - \sigma_{\mathsf{attn}}(0)\big]\Big| \le 0.5\times 4\tau \tau^{-2} |x|^2 = 2\tau^{-1}|x|^2,\qquad \forall x\in\mathbb{R}.
\end{align}
Taking this collectively with \eqref{eq:Attn1-true-update-no-e} and the choice that $1-w_i\neq 0$ only for $i=N+1$, we arrive at
\begin{align}\label{eq:proof-bound-e-pre}
\|{\bm e}^{(l)}\|_{\infty}
&\le \frac{4\eta^2}{N\tau}\sum_{i=1}^{N}\big(y_i^{(l)}\big)^2 + \frac{4\eta^2}{N\tau}\sum_{i=1}^{N+1}\left(\big({\bm \phi}_i^{(l)}\big)^{\top}{\bm \rho}^{(l)}\right)^2 \notag + \frac{4\eta^2\left(\widehat{y}^{(l)}\right)^2}{N\tau} + \frac{2\eta}{N}\left|\widehat{y}^{(l)} - \big({\bm \phi}_{N+1}^{(l)}\big)^{\top}{\bm \rho}^{(l)}\right|\notag\\
&\lesssim \frac{\eta^2}{\tau}\left(C_{\mathcal{F}}^2 + \sigma^2 + \|{\bm \rho}^{(l)}\|_1^2 + \big(\widehat{y}^{(l)}\big)^2\right)+ \frac{2\eta}{N}\left|\widehat{y}^{(l)} - \big({\bm \phi}_{N+1}^{(l)}\big)^{\top}{\bm \rho}^{(l)}\right|,
\end{align}
where we have used the facts that 
\begin{align*}
|y_i^{(l)}| &= |y_i|\le |f({\bm x}_i)| + |z_i|\le|{\bm \phi}_i^{\top}{\bm \rho}^{\star}| + \varepsilon_{\mathsf{dis}} + |z_1|\lesssim \|{\bm \rho}^{\star}\|_1 + C_{\mathcal{F}} + \sigma \asymp C_{\mathcal{F}} + \sigma, \\ 
|\big({\bm \phi}_{i}^{(l)}\big)^{\top}{\bm \rho}^{(l)}| &= |{\bm \phi}_{i}^{\top}{\bm \rho}^{(l)}|\le \|{\bm \phi}_{i}\|_{\infty}\|{\bm \rho}^{(l)}\|_1\le \|{\bm \rho}^{(l)}\|_1.
\end{align*}

\item \textbf{FF1}: 
In order to implement the soft-thresholding operator using the feed-forward layer, we first need to inspect the connection between the soft-thresholding operator and the ReLU function $\sigma_{\mathsf{ff}}(\cdot)$. Towards this end, observe that 
\begin{align}
\mathsf{ST}_{\eta\lambda}(z) &= z +\eta\lambda - (z+\eta\lambda)\ind(z+\eta\lambda>0) +  (z-\eta\lambda)\ind(z-\eta\lambda>0) \notag\\
&=z +\sigma_{\mathsf{ff}}(\eta\lambda) - \sigma_{\mathsf{ff}}(z+\eta\lambda) +  \sigma_{\mathsf{ff}}(z-\eta\lambda).
\end{align}
Consequently, we propose to design a feed-forward layer capable of implementing the following updates (in an attempt to carry out the proposed update \eqref{eq:equiv-update-FF1}):    
\begin{align*}
{\bm \rho}^{(l)} &= {\bm \rho}^{(l-1/2)} +\sigma_{\mathsf{ff}}(\eta\lambda^{(l-1/2)})- \sigma_{\mathsf{ff}}({\bm \rho}^{(l-1/2)} + \eta\lambda^{(l-1/2)}) + \sigma_{\mathsf{ff}}\big({\bm \rho}^{(l-1/2)} - \eta\lambda^{(l-1/2)}\big) \notag\\
&= \mathsf{ST}_{\eta\lambda^{(l-1/2)}}\big({\bm \rho}^{(l-1/2)}\big),\notag\\
\widehat{y}^{(l)} &= \widehat{y}^{(l-1/2)} -\sigma_{\mathsf{ff}}\big(\widehat{y}^{(l-1/2)}\big) + \sigma_{\mathsf{ff}}\big(-\widehat{y}^{(l-1/2)}\big)= 0.
\end{align*}
To do so, it suffices to set ${\bm W}^{(l)}\in\mathbb{R}^{D\times D}$, ${\bm U}^{(l)}\in\mathbb{R}^{D\times D}$ to be
\begin{subequations}
\begin{align}
{\bm W}^{(l)}_{1:2n+5,\,:} &= \left[
\begin{array}{cccc}
{\bm 0}_{(n+1)\times (d+n+4)} &{\bm I}_{n+1} & \eta{\bm 1} &{\bm 0}\\
{\bm 0}_{(n+1)\times (d+n+4)}& {\bm I}_{n+1}& -\eta{\bm 1}& {\bm 0}\\
&    \eta{\bm u}_{d+2n+6}^{\top} && \\
 &    {\bm u}_{d+2n+7}^{\top} && \\
 &    -{\bm u}_{d+2n+7}^{\top} &&
\end{array}
\right],\\
{\bm U}_{d+n+5:d+2n+5,\,1:2n+5}^{(l)} &= [-{\bm I}_{n+1},{\bm I}_{n+1},{\bm 1}_{n+1},{\bm 0},{\bm 0}],\qquad 
{\bm U}_{D,1:2n+5}^{(l)} = [{\bm 0}_{2n+3}^{\top},-1,1],
\end{align}
with all remaining entries set to zero.
\end{subequations}

\item \textbf{Attn2}: Recall that \eqref{eq:approx-sigma-attn} tells us that
\begin{align}
\label{eq:phi-rho-inner-approx-attn}
\big({\bm \phi}_{i}^{(l)}\big)^{\top}{\bm \rho}^{(l)}\approx 4\tau\Big\{\sigma_{\mathsf{attn}}\left(\tau^{-1}\big({\bm \phi}_{i}^{(l)}\big)^{\top}{\bm \rho}^{(l)}\right) - \sigma_{\mathsf{attn}}(0)\Big\}.
\end{align}
As a result, we would like to design 
this attention layer to actually implement 
\begin{align}
\widehat{y}^{(l+1/2)} &= \widehat{y}^{(l)} +4\tau \sum_{i = 1}^{N+1} \big(1-{w}_i^{(l)}\big)\left[\sigma_{\mathsf{attn}}\left(\tau^{-1}\big({\bm \phi}_{i}^{(l)}\big)^{\top}{\bm \rho}^{(l)}\right) - \sigma_{\mathsf{attn}}(0) \right], 
\label{eq:Attn2-actual-update}
\end{align}
and hence the residual term $\widetilde{e}^{(l)}$ in \eqref{eq:update-haty} can be easily determined by comparing \eqref{eq:update-haty} with \eqref{eq:Attn2-actual-update}. 

To realize \eqref{eq:Attn2-actual-update}, it suffices to use 2 attention heads, and 
set ${\bm V}_1^{(l+1)},{\bm V}_2^{(l+1)}\in\mathbb{R}^{D\times D}$, ${\bm Q}_1^{(l+1)},{\bm Q}_2^{(l+1)}\in\mathbb{R}^{D\times D}$, and ${\bm K}_1^{(l+1)},{\bm K}_2^{(l+1)}\in\mathbb{R}^{D\times D}$ to be:
\begin{align*}
{\bm V}_{1,(D,\,:)}^{(l+1)} &= 4\tau({\bm u}_{d+1} - {\bm u}_{d+3})^{\top},\quad {\bm Q}_{1,(1:n+1,\,d+4:d+n+4)}^{(l+1)} = \tau^{-1}{\bm I}_{n+1},\quad {\bm K}_{1,(1:n+1,\,d+n+5:d+2n+5)}^{(l+1)} = {\bm I}_{n+1},\notag\\
{\bm V}_2^{(l+1)} &= -{\bm V}_1^{(l+1)},\quad {\bm Q}_2^{(l+1)} = {\bm 0}, \quad {\bm K}_2^{(l+1)} = {\bm 0},
\end{align*}
with all remaining entries set to zero. Here, we recall that $\bm{u}_i$ denotes the $i$-th standard basis vector.

Before moving on, let us single out some bounds on $\widetilde{e}^{(l)}$. 
%
%
Recalling that $1-{{w}}_i^{(l)} = 0$ for all $1\le i\le N$ and making use of \eqref{eq:phi-rho-inner-approx-attn} and \eqref{eq:approx-sigma-att}, we can demonstrate that
\begin{align}
\label{eq:proof-bound-e-prime}
|\widetilde{e}^{(l)}|\le \frac{2\big(({\bm \phi}_{N+1}^{(l)})^{\top}{\bm \rho}^{(l)}\big)^2}{\tau}.
\end{align}
%

\item \textbf{Attn0, FF2}:
The parameter matrices in these layers are taken to be
\begin{align}
{\bm U}^{(l+1)} = {\bm W}^{(l+1)} = {\bm Q}^{(1)} = {\bm K}^{(1)} = {\bm V}^{(1)} = {\bm 0},
\end{align}
so as to ensure that
$$
{\bm H}^{(1/2)} = {\bm H}^{(0)}\qquad \text{and}\qquad {\bm H}^{(l+1)} = {\bm H}^{(l+1/2)}.
$$
\end{itemize}

\subsubsection{Verifying \eqref{eq:rela-rhol-rhot} and controlling the size of ${\bm e}_t$} 
We now verify \eqref{eq:rela-rhol-rhot} by induction, and establish upper bounds on ${\bm e}_t$ in \eqref{eq:update-rule-PGD}.

First, it is self-evident that ${\bm \rho}^{(1)} = {\bm \rho}_0^{\mathsf{proximal}} = {\bm 0}$.
Now, let us assume that ${\bm \rho}^{(2t-1)} = {\bm \rho}_{t-1}^{\mathsf{proximal}}$, and proceed to prove
${\bm \rho}^{(2t+1)} = {\bm \rho}_t^{\mathsf{proximal}}$.
In the $t$-th block (which contains $\textbf{Attn1},\textbf{FF1},\textbf{Attn2},\textbf{FF2}$), we have
\begin{align*}
{\bm \rho}^{(2t+1)} 
&= {\bm \rho}^{(2t)} =\mathsf{ST}_{\eta\overline{\lambda}}\big({\bm \rho}^{(2t-1/2)}\big) = \mathsf{ST}_{\eta\overline{\lambda}}\bigg({\bm \rho}^{(2t-1)}  + \frac{2\eta}{N}\sum_{i = 1}^{N} {\bm \phi}_i\big\{y_i - {\bm \phi}_i^{\top}{\bm \rho}^{(2t-1)}\big\} + {\bm e}^{(2t-1)}\bigg)\nonumber\\
&=\mathsf{ST}_{\eta\overline{\lambda}}\bigg({\bm \rho}_{t-1}^{\mathsf{proximal}}  + \frac{2\eta}{N}\sum_{i = 1}^{N} {\bm \phi}_i\big\{y_i - {\bm \phi}_i^{\top}{\bm \rho}_{t-1}^{\mathsf{proximal}}\big\}\bigg) + {\bm e}_{t} = {\bm \rho}_t^{\mathsf{proximal}},
\end{align*}
where 
$$
{\bm e}_{t} \coloneqq \mathsf{ST}_{\eta\overline{\lambda}}\bigg({\bm \rho}_{t-1}^{\mathsf{proximal}}  + \frac{2\eta}{N}\sum_{i = 1}^{N} {\bm \phi}_i\big\{y_i - {\bm \phi}_i^{\top}{\bm \rho}_{t-1}^{\mathsf{proximal}}\big\}+ {\bm e}^{(2t-1)}\bigg)  - \mathsf{ST}_{\eta\overline{\lambda}}\bigg({\bm \rho}_{t-1}^{\mathsf{proximal}}  + \frac{2\eta}{N}\sum_{i = 1}^{N} {\bm \phi}_i\big\{y_i - {\bm \phi}_i^{\top}{\bm \rho}_{t-1}^{\mathsf{proximal}}\big\}\bigg).
$$
Recalling that $\mathsf{ST}(\cdot)$ is a contraction operator and using \eqref{eq:proof-bound-e-pre}, we can show that
\begin{align}\label{eq:proof-bound-e-pre-2}
\|{\bm e}_{t}\|_\infty &\le \|{\bm e}^{(2t-1)}\|_\infty
\lesssim \frac{\eta^2}{\tau}\left(C_{\mathcal{F}}^2 + \sigma^2 + \|{\bm \rho}^{(2t-1)}\|_1^2 + \big(\widehat{y}^{(2t-1)}\big)^2\right)+ \frac{\eta}{N}\left|\widehat{y}^{(2t-1)} - {\bm \phi}_{N+1}^{\top}{\bm \rho}^{(2t-1)}\right|\notag\\
&\lesssim \frac{\eta^2}{\tau}\left(C_{\mathcal{F}}^2 + \sigma^2 + \|{\bm \rho}_{t-1}^{\mathsf{proximal}}\|_1^2 + \big(\widehat{y}^{(2t-1)}\big)^2\right)+ \frac{\eta}{N}\left|\widehat{y}^{(2t-1)} - {\bm \phi}_{N+1}^{\top}{\bm \rho}_{t-1}^{\mathsf{proximal}}\right|,
\end{align}
which is valid since $ {\bm \rho}^{(2t-1)} = {\bm \rho}_{t-1}^{\mathsf{proximal}}$.
Also, it follows from \eqref{eq:update-haty} that
\begin{align*}
\widehat{y}^{(2t-1)} 
 = \widehat{y}^{(2t-3/2)} 
&=  {\bm \phi}_{N+1}^{\top}{\bm \rho}^{(2t-2)} + \widetilde{e}^{(2t-2)}
=  {\bm \phi}_{N+1}^{\top}{\bm \rho}_{t-1}^{\mathsf{proximal}} + \widetilde{e}^{(2t-2)},
\end{align*}
where we have used ${\bm \rho}^{(2t-2)} = {\bm \rho}^{(2t-1)} = {\bm \rho}_{t-1}^{\mathsf{proximal}}$.
Moreover, it is seen from \eqref{eq:proof-bound-e-prime} that
\begin{align*}
|\widetilde{e}^{(2t-2)}|\le \frac{2\left({\bm \phi}_{N+1}^{\top}{\bm \rho}^{(2t-2)}\right)^2}{\tau} = \frac{2\left({\bm \phi}_{N+1}^{\top}{\bm \rho}_{t-1}^{\mathsf{proximal}}\right)^2}{\tau}.
\end{align*}
Substituting these into \eqref{eq:proof-bound-e-pre-2} and using $\|\bm{\phi}_{N+1}\|_{\infty}\leq 1$ then yield
\begin{align}\label{eq:proof-bound-e}
\|{\bm e}_{t}\|_{\infty}
&\lesssim \frac{\eta^2}{\tau}\left(C_{\mathcal{F}}^2 + \sigma^2 + \|{\bm \rho}_{t-1}^{\mathsf{proximal}}\|_1^2 + ({\bm \phi}_{N+1}^{\top}{\bm \rho}_{t-1}^{\mathsf{proximal}})^2 + (\widetilde{e}^{(2t-2)})^2\right)+ \frac{\eta}{N}|\widetilde{e}^{(2t-2)}|\notag\\
&\lesssim \frac{\eta^2}{\tau}\left(C_{\mathcal{F}}^2 +\sigma^2 + \|{\bm \rho}_t^{\mathsf{proximal}}\|_1^2 + ({\bm \phi}_{N+1}^{\top}{\bm \rho}_{t-1}^{\mathsf{proximal}})^2 + \frac{({\bm \phi}_{N+1}^{\top}{\bm \rho}_{t-1}^{\mathsf{proximal}})^4}{\tau^2}\right)+ \frac{\eta({\bm \phi}_{N+1}^{\top}{\bm \rho}_{t-1}^{\mathsf{proximal}})^2}{N\tau}\notag\\
&\lesssim \frac{\eta^2}{\tau}\left(C_{\mathcal{F}}^2 + \sigma^2 + \|{\bm \rho}_{t-1}^{\mathsf{proximal}}\|_1^2 + \frac{\|{\bm \rho}_{t-1}^{\mathsf{proximal}}\|_1^4}{\tau^2}\right)+ \frac{\eta\|{\bm \rho}_{t-1}^{\mathsf{proximal}}\|_1^2}{N\tau}.
\end{align}

\subsubsection{Proof of property iii) in Lemma \ref{lem:opt}} 
Equipped with the above transformer parameters, we are now positioned to establish property iii) in Lemma \ref{lem:opt}.
To this end,
we first establish the following lemma, whose proof is postponed to Appendix~\ref{sec:proof-lem:bound-rho}.
\begin{lemma}\label{lem:bound-rho}
 Suppose that $\lambda\ge C\sqrt{\log N/N} (C_{\mathcal{F}} + \sigma) + C_{\mathcal{F}}^{-1}\varepsilon_{\mathsf{dis}}^2$, and take $\tau$ to be sufficiently large such that
$$
\tau \ge CNn^2(L+n)\left(N+ n\right)C_{\mathcal{F}},
$$
with $C>0$ some large enough constant.
Then for all $t\ge 0$, it holds that
\begin{align*}
\|{\bm e}_t\|_1&\lesssim \frac{C_{\mathcal{F}}}{(L+n)nN}\qquad\text{and}\qquad 
\big\|{\bm \rho}_t^{\mathsf{proximal}}\big\|_1\lesssim n\sqrt{N}C_{\mathcal{F}}.
\end{align*}
\end{lemma}

When $L = 2T+1$ is an odd number, it holds that ${\bm \rho}^{(L)} = {\bm \rho}_T^{\mathsf{proximal}}$ and $\widehat{y}^{(L)} = \widehat{y}^{(L-1/2)}$. 
Combining Lemma \ref{lem:bound-rho} with \eqref{eq:proof-bound-e-prime} then reveals that: 
by taking
$$
\tau \ge CNn^2(L+n)\left(N+ n\right)C_{\mathcal{F}} \ge C n^2N^{5/4}C_{\mathcal{F}}$$
for some large enough constant $C>0$, we have
\begin{align*}
\big|{\bm \phi}_{N+1}^{\top}{\bm \rho}^{(L)} - \widehat{y}^{(L)}\big| 
&= 
\big|{\bm \phi}_{N+1}^{\top}{\bm \rho}^{(L-1)} - \widehat{y}^{(L-1/2)}\big|  =|\widetilde{e}^{(L-1)}| \le \frac{2({\bm \phi}_{N+1}^{\top}{\bm \rho}_{\frac{L-1}{2}}^{\mathsf{proximal}})^2}{\tau} \\
& 
\le  \frac{2\|{\bm \phi}_{N+1}\|_{\infty}^2 \big\|{\bm \rho}_{\frac{L-1}{2}}^{\mathsf{proximal}} \big\|_1^2}{\tau}
\le \left(\frac{\log N}{N}\right)^{1/4}C_{\mathcal{F}}.
\end{align*}

Additionally, taking Lemma \ref{lem:bound-rho} together with  \eqref{eq:convergence-rho} leads to
\begin{align} \label{eq:convergence}
\ell({\bm \rho}_T^{\mathsf{proximal}}) -\ell({\bm \rho}^{\star})
&\lesssim \frac{nC_{\mathcal{F}}^2}{L} + (L+n)\frac{C_{\mathcal{F}}}{(L+n)nN}(C_{\mathcal{F}} + \sigma + n\sqrt{N}C_{\mathcal{F}} + \lambda)\notag\\
&\lesssim \frac{nC_{\mathcal{F}}^2}{L}+
\frac{1}{\sqrt{N}}C_{\mathcal{F}}(C_{\mathcal{F}} + \sigma) + \lambda C_{\mathcal{F}}.
\end{align}
Recalling the bound on $\lambda C_\mathcal{F}$ in \eqref{eq:bound-lambda-C} and the definition of $\widehat{\varepsilon}$ in \eqref{eq:def-hat-eps}, we have
\begin{align*}
\lambda C_\mathcal{F}\lesssim \sqrt{\frac{\log N}{N}}C_{\mathcal{F}}\Big(C_{\mathcal{F}} + \sigma\Big) + \varepsilon_{\mathsf{dis}}^2 + \widehat{\varepsilon} \lesssim \sqrt{\frac{\log N}{N}}C_{\mathcal{F}}\Big(C_{\mathcal{F}} + \sigma\Big) + \varepsilon_{\mathsf{dis}}^2 + \frac{n C_\mathcal{F}^2}{L},
\end{align*}
which combined with \eqref{eq:convergence} completes the proof.

\subsubsection{Proof of Lemma \ref{lem:bound-rho}}
\label{sec:proof-lem:bound-rho}
We intend to prove the following inequalities by induction:
\begin{align}
\|{\bm e}_t\|_1&\le \frac{c_1C_{\mathcal{F}}}{(L+n)Nn},\label{eq:lem-bound-e}\\
\|{\bm \rho}_t^{\mathsf{proximal}}\|_1&\le c_2n\sqrt{N}C_{\mathcal{F}},\label{eq:lem-bound-rho}
\end{align}
for some sufficiently small (resp.~large) constant $c_1>0$ (resp.~$c_2>0$).

For the base case, we have $$\|{\bm e}_0\|_1=\|{\bm \rho}_0^{\mathsf{proximal}} - {\bm \rho}_0^{\star}\|_1 = 0,$$ and thus \eqref{eq:lem-bound-e} holds for $t=0$.
Next, we intend to prove that \eqref{eq:lem-bound-rho} holds for $t=k$ under the assumption that \eqref{eq:lem-bound-e} holds for $t\le k$ and \eqref{eq:lem-bound-rho} holds for $t\le k-1$.
According to \eqref{eq:lem-convergence-bound-rho}, it holds that
\begin{align*}
\|{\bm \rho}_{k}^{\mathsf{proximal}}\|_1 &\le
c_5C_{\mathcal{F}}  + \frac{c_5nC_{\mathcal{F}}^2}{k\lambda} + c_5\sqrt{N}(k+n)\max_{1\le i\le k}\|{\bm e}_{i}\|_1 + \frac{c_5(k+n)}{\lambda}\max_{1\le i\le k}\big\{\|{\bm e}_i\|_1\|{\bm \rho}_i^{\mathsf{proximal}}\|_1\big\}\notag\\
&\overset{\text{(a)}}{\le} c_5C_{\mathcal{F}} + 
\frac{c_5n\sqrt{N}C_{\mathcal{F}}}{c_{\lambda}} + c_5c_1 C_{\mathcal{F}}
+ \frac{c_5c_1c_2C_{\mathcal{F}}}{c_{\lambda}}+ \frac{c_5c_1}{c_{\lambda}\sqrt{N}n}\|{\bm \rho}_k^{\mathsf{proximal}}\|_1,
\end{align*}
where (a) results from the fact that $\lambda\ge c_{\lambda}C_{\mathcal{F}}/\sqrt{N}$.
For some sufficiently small constant $c_1>0$ obeying $c_1c_5/c_{\lambda} \le 1/4$, we have
\begin{align*}
\|{\bm \rho}_{k}^{\mathsf{proximal}}\|_1 
&\le \frac{4c_5C_{\mathcal{F}}}{3}\left(1+\frac{n\sqrt{N}}{c_{\lambda}} + c_1\right) + \frac{c_2C_{\mathcal{F}}}{3}\le c_2n\sqrt{N}C_{\mathcal{F}},
\end{align*}
provided that $c_2\ge 2c_5\left(1+c_{\lambda}^{-1} + c_1\right)$.

Next, we intend to establish that \eqref{eq:lem-bound-e} holds for $t=k+1$.
By virtue of \eqref{eq:proof-bound-e}, one has
\begin{align*}
\|{\bm e}_{k+1}\|_1
&\le \frac{c_3n\eta^2}{\tau}\left(C_{\mathcal{F}}^2 + \sigma^2 + \|{\bm \rho}_k^{\mathsf{proximal}}\|_1^2 + \frac{\|{\bm \rho}_k^{\mathsf{proximal}}\|_1^4}{\tau^2}\right)+ \frac{c_3n\eta\|{\bm \rho}_k^{\mathsf{proximal}}\|_1^2}{N\tau}\notag\\
&\le \frac{c_3}{4n\tau}\left(1+2c_2^2n^2N+ 2c_2^2n^3\right)C_{\mathcal{F}}^2 + \frac{c_3\sigma^2}{4n\tau},
\end{align*}
for $\tau \ge c_2n\sqrt{N}C_{\mathcal{F}}$.
Without loss of generality, we assume that $\sqrt{\log N/N}\sigma \le \sqrt{c_4}C_{\mathcal{F}}$,
since otherwise the upper bound in Theorem \ref{thm:main} is larger than $C_{\mathcal{F}}^2$ and holds trivially by outputting $\widehat{\bm \rho} = {\bm 0}$.
It is then seen that
\begin{align*}
\|{\bm e}_{k+1}\|_1
&\le \frac{c_3}{4n\tau}\left(1+2c_2^2n^2N+ 2c_2^2n^3\right)C_{\mathcal{F}}^2  + \frac{c_3c_4NC_{\mathcal{F}}^2}{4n\tau}\notag\\
&=\frac{C_{\mathcal{F}}^2}{\tau}\left(\frac{c_3}{4n}(1+2c_2^2n^2N+ 2c_2^2n^3 + c_4N)\right)\notag\\
&\le \frac{c_1C_{\mathcal{F}}}{(L+n)Nn},
\end{align*}
with the proviso that 
$$
\tau \ge \frac{Nn(L+n)C_{\mathcal{F}}}{c_1}\left(\frac{c_3}{4n}(1+2c_2^2n^2N+ 2n^3c_2^2 + c_4N)\right) + c_2n\sqrt{N}C_{\mathcal{F}}.
$$
This completes the proof.

\section{Barron-style parameters for several examples of function classes}
\label{app:calculation-CF}

We now upper bound the Barron-style parameters for several function classes mentioned in Section \ref{sec:Barron-pars}.

\paragraph{Class of linear functions.}
Consider any  linear function $f_{\bm{a},b}({\bm x}) \coloneqq {\bm a}^{\top}{\bm x} + b$, which clearly satisfies $f_{\bm{a},b}({\bm 0})=b$.   
As can be readily seen,  the gradient of $f_{\bm{a},b}({\bm x})$ w.r.t.~$\bm{x}$ and its Fourier transform are given by
$$
\nabla f_{\bm{a},b}({\bm x}) = {\bm a},\qquad F_{\nabla f_{\bm{a},b}}({\bm \omega}) = {\bm a}\delta({\bm \omega}),
$$
where $\delta(\cdot)$ denotes the Dirac function.
Thus, it follows from property \eqref{eq:condition-Fourier-gradient} that
$$
C_{\mathcal{F}^{\mathsf{linear}}} = \sup_{|b|\le C_b}|b| + \sup_{\|{\bm a}\|_2\le C_a}\|{\bm a}\|_2\int  \delta({\bm \omega})\mathrm{d}{\bm \omega} = C_a + C_b.
$$

\paragraph{Linear composition of function classes.}
For function $g_{\bm{a},b}(\bm{x}) = \langle {\bm a}, {\bm f}({\bm x}) \rangle + b$, its gradient with respect to ${\bm x}$ and its Fourier transform can be computed as
$$
\nabla g_{\bm{a},b}({\bm x}) = \sum_{i=1}^M a_i\nabla f_i({\bm x}),\qquad F_{\nabla g_{\bm{a},b}}({\bm \omega}) = \sum_{i=1}^M a_iF_{\nabla f_i}({\bm \omega}). 
$$
This linearity property combined with property \eqref{eq:condition-Fourier-gradient} then gives
$$
C_{\mathcal{F}^{\mathsf{comp}}}\le C_b + \sup_{\|{\bm a}\|_1\le C_a} \|{\bm a}\|_1\|{\bm f}({\bm 0})\|_{\infty} +  \sup_{\|{\bm a}\|_1\le C_a} \|{\bm a}\|_1\int \max_{1\leq i\leq M}\|F_{\nabla f_i}({\bm \omega})\|_2 \mathrm{d} {\bm\omega}\le 2C_a\max_{1\leq i\leq M} C_{\mathcal{F}_i} + C_b.
$$

\paragraph{Class of two-layer neural networks.}
We first claim that
\begin{align}\label{eq:gradient-Fourier-frho}
F_{\nabla f_{\rho}}({\bm \omega}) &=\int \rho({\bm a})\prod_{s=2}^d\delta(\widetilde{\omega}_s)\frac{\overline{\bm a}}{2\sinh\big(\pi\frac{\widetilde{\omega}_1}{\|{\bm a}\|_2}\big)}\frac{\widetilde{\omega}_1}{\|{\bm a}\|_2} \mathrm{d} {\bm a},
\end{align}
where $\sinh(x) = {\rm e}^x/2 - {\rm e}^{-x}/2$. 
Here,  $\overline{\bm a} \coloneqq {\bm a}/\|{\bm a}\|_2$ represents the direction of ${\bm a}$, $\widetilde{\omega}_s$ denotes the $s$-th entry of $\widetilde{\bm \omega} \coloneqq {\bm Q}{\bm \omega}$, and ${\bm Q}^{\top} \coloneqq [\overline{\bm a},\widetilde{\bm Q}]\in\mathbb{R}^{d\times d}$ is an orthogonal matrix whose 2nd-through-$d$th rows
$\widetilde{\bm Q}$ satisfy $\widetilde{\bm Q}^{\top}\widetilde{\bm Q} = {\bm I}_{d-1}$ and $\widetilde{\bm Q}^{\top}{\bm a} = {\bm 0}$.
The proof of \eqref{eq:gradient-Fourier-frho} is postponed to the end of this subsection. 

Recalling that $|\rho({\bm a})|\le \rho_{\mathsf{max}}({\bm a})$, we can invoke the claim \eqref{eq:gradient-Fourier-frho} to reach
\begin{align*}
\sup_{\rho}\|F_{\nabla f_{\rho}}({\bm \omega})\|_2 &\le \int \rho_{\mathsf{max}}({\bm a})\prod_{s=2}^d\delta(\widetilde{\omega}_s)\frac{1}{2\big|\sinh\big(\pi\frac{\widetilde{\omega}_1}{\|{\bm a}\|_2}\big)\big|}\frac{|\widetilde{\omega}_1|}{\|{\bm a}\|_2} \mathrm{d} {\bm a}.
\end{align*}
By computing the above integral, we obtain
\begin{align*}
\int \sup_{\rho}\|F_{\nabla f_{\rho}}({\bm \omega})\|_2 \mathrm{d}{\bm \omega} &\le \int \rho_{\mathsf{max}}({\bm a})\left(\prod_{s=2}^d\int\delta(\widetilde{\omega}_s)\mathrm{d} \widetilde{\omega}_s\right)\int \frac{1}{2|\sinh\big(\pi\frac{\widetilde{\omega}_1}{\|{\bm a}\|_2}\big)|}\frac{|\widetilde{\omega}_1|}{\|{\bm a}\|_2} \mathrm{d} \widetilde{\omega}_1 \mathrm{d} {\bm a}\notag\\
&= \int \rho_{\mathsf{max}}({\bm a})\int_0^{\infty} \frac{1}{\sinh\big(\pi\frac{\widetilde{\omega}_1}{\|{\bm a}\|_2}\big)}\frac{\widetilde{\omega}_1}{\|{\bm a}\|_2} \mathrm{d} \widetilde{\omega}_1 \mathrm{d} {\bm a}\notag\\
&= \int \rho_{\mathsf{max}}({\bm a})\int_0^{\infty} \frac{\|{\bm a}\|_2\omega}{\sinh\big(\pi\omega\big)}\mathrm{d} {\omega}\mathrm{d} {\bm a} \overset{\text{(i)}}{=} \frac14\int \rho_{\mathsf{max}}({\bm a})\|{\bm a}\|_2\mathrm{d} {\bm a},
\end{align*}
where (i) results from the fact that $$\int_{0}^{\infty} \frac{\omega}{\sinh\left(\pi\omega\right)} \mathrm{d} {\omega} = 2\sum_{n=0}^{\infty}\int_{0}^{\infty}\omega{\rm e}^{-(2n+1)\pi \omega}\mathrm{d}\omega = 2\sum_{n=0}^{\infty}\frac{1}{(2n+1)^2\pi^2} = 1/4.$$
Furthermore, it is readily seen that
$$
\sup_{\rho} |f_{\rho}({\bm 0})| = \frac12\bigg|\sup_{\rho}\int \rho({\bm a}) \mathrm{d}{\bm a}\bigg| \le \frac12\int \rho_{\mathsf{max}}({\bm a}) \mathrm{d}{\bm a}.
$$
Putting the above results together completes the proof.

\paragraph{Proof of Claim~\eqref{eq:gradient-Fourier-frho}.}
Basic calculus tells us that the gradient of $\sigma({\bm a}^{\top}{\bm x})$ w.r.t.~${\bm x}$ is
\begin{align}
\nabla \sigma({\bm a}^{\top}{\bm x}) = \frac{\exp(-{\bm a}^{\top}{\bm x})}{\left(1+\exp(-{\bm a}^{\top}{\bm x})\right)^2} {\bm a}.
\label{eq:grad-f-logistic}
\end{align}
Recall that the Fourier transform of scalar function ${\rm e}^{-x}(1+{\rm e}^{-x})^{-2}$ can be calculated as
\begin{align}
\label{eq:FT-sigmoid}
\frac{1}{2\pi}\int \frac{{\rm e}^{-x}}{(1+{\rm e}^{-x})^2}{\rm e}^{-j\omega x}\mathrm{d} x
= \frac{\omega}{2\sinh(\pi\omega)}= \frac{\omega}{{\rm e}^{\pi\omega} - {\rm e}^{-\pi\omega}}
\end{align}
for any $\omega \in \mathbb{R}$, 
where $\sinh(x) = {\rm e}^x/2 - {\rm e}^{-x}/2$.

Define $\widetilde{\bm x} \coloneqq {\bm Q}{\bm x}$, where ${\bm Q}^{\top}$ is the orthogonal matrix defined previously.
Denoting by $\widetilde{x}_s$ the $s$-th entry of $\widetilde{\bm{x}}$, we can then compute the Fourier transform of $\nabla  \sigma({\bm a}^{\top}{\bm x})$ (cf.~\eqref{eq:grad-f-logistic}) as 
\begin{align*}
F_{\nabla \sigma({\bm a}^{\top}\cdot)}({\bm \omega}) &=\frac{{\bm a}}{(2\pi)^d}\int \frac{{\rm e}^{-j{\bm \omega}^{\top}{\bm x}}\exp(-{\bm a}^{\top}{\bm x})}{(1+\exp(-{\bm a}^{\top}{\bm x}))^2}\mathrm{d}{\bm x} 
=\frac{{\bm a}}{2\pi}\int \frac{{\rm e}^{-j\widetilde{\omega}_1\widetilde{x}_1}\exp(-\|{\bm a}\|_2\widetilde{x}_1)}{(1+\exp(-\|{\bm a}\|_2\widetilde{x}_1))^2}\mathrm{d} \widetilde{x}_1 \frac{1}{(2\pi)^{d-1}}\prod_{s=2}^d\int {\rm e}^{-j\widetilde{w}_s\widetilde{x}_s}\mathrm{d} \widetilde{x}_s\notag\\
&=\prod_{s=2}^d\delta(\widetilde{\omega}_s)\frac{\overline{\bm a}}{2\sinh\big(\pi\frac{\overline{\bm a}^{\top}{\bm \omega}}{\|{\bm a}\|_2}\big)}\frac{\overline{\bm a}^{\top}{\bm \omega}}{\|{\bm a}\|_2},
\end{align*}
where $\delta(\cdot)$ represents the Dirac function as before and we have used \eqref{eq:FT-sigmoid}. 
Taking this together with the definition of $f_{\rho}({\bm x})$ establishes the claim ~\eqref{eq:gradient-Fourier-frho}.

\bibliographystyle{apalike}
\bibliography{refs}

\end{document}